\documentclass[10pt,twocolumn,letterpaper]{article}

\usepackage[pagenumbers]{cvpr} 

\usepackage{cuted} 

\usepackage{currfile} 

\usepackage{caption} 

\usepackage{microtype}

\usepackage{array}

\definecolor{cvprblue}{rgb}{0.21,0.49,0.74}
\usepackage[pagebackref,breaklinks,colorlinks,allcolors=cvprblue]{hyperref}

\def\confName{CVPR}
\def\confYear{2027}

\title{Reading Geometric Anchors: Viewpoint Generalization in Imitation Learning for Robotic Manipulation with RGB-Only Deployment}

\author{Chengyu Tao\\
College of Mechanical and Vehicle Engineering\\
Hunan University, Changsha, China\\
{\tt\small taochengyu@hnu.edu.cn}
\and
Chunxi Huang\thanks{Corresponding author.}\\
School of Design\\
Hunan University, Changsha, China\\
{\tt\small tracyhuang@hnu.edu.cn}}

\begin{document}
\maketitle
\begin{abstract}
Moving the camera changes where objects appear in an image, challenging RGB-based imitation learning.
A common strategy pairs synthesized views with the same demonstrated action, but ignores supporting physical correspondence across views. By predicting depth with camera calibrations, 3D policies may handle it yet suffer the issue of varied observed point sets. To address these limitations, we propose \textbf{ReGeoAnch}, a geometry-anchored imitation learning framework for viewpoint generalization. Specifically, we build global anchors in the robot base frame and local anchors in the gripper frame, providing shared physical references across viewpoints. Furthermore,
\textbf{Decoupled Appearance-Guided Geometric Readout} uses Point-to-Anchor Aggregation (PAA) to obtain comparable and stable geometry around anchors beyond pointwise correspondence. A controllable and decoupled information flow, Anchor-to-Policy Readout (APR), leverages complementary visual and geometric information and supports viewpoint generalization: view-dependent appearances are queried by view-invariant robot state to select task-relevant anchors, with geometric values alone to form the final readout.   
We further introduce \textbf{Cross-View Readout Consistency} to align the directions of policy readouts from paired source and synthesized target views.
Experiments on 12 manipulation tasks across MetaWorld and ManiSkill3 demonstrate improved viewpoint generalization over baseline policies.
\end{abstract}

\section{Introduction}
\label{sec:intro}

A manipulation skill should remain effective when the camera viewpoint changes, even though the required actions in robot coordinates remain unchanged~\cite{jiang2026camera}. In practice, an RGB-D camera can be used temporarily to collect demonstrations, while deployment may rely on an existing RGB camera mounted at a different position to fit the workspace, avoiding additional depth hardware at each installation~\cite{pang2024depthhelps,jiang2026camera,monolift}. We therefore study viewpoint generalization from single-view RGB-D demonstrations to a target RGB camera whose calibration is known during training, without target-view demonstrations.

\begin{figure}[!t]
    \centering
    \includegraphics[pagebox=cropbox,width=\columnwidth]{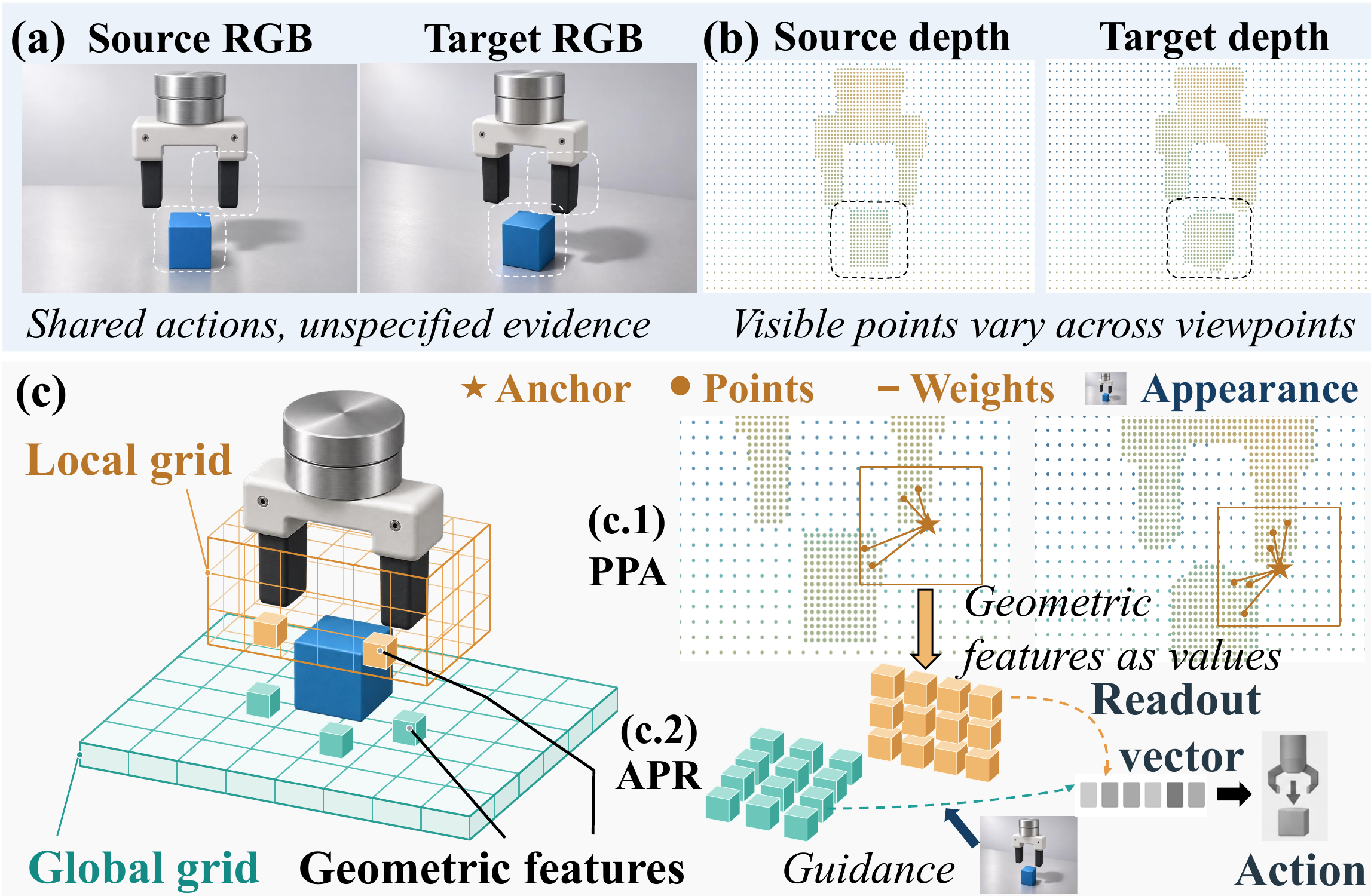}
\caption{(a) Novel-view synthesis provides paired views with shared action labels, but these labels do not specify which physical regions should support prediction.
(b) Point clouds capture different visible surfaces across views, while geometry alone may not identify task-relevant regions.
(c) ReGeoAnch uses global and gripper-local anchors as shared spatial references.
(c.1) \textit{Point-to-Anchor Aggregation} pools visible points within each anchor's neighborhood, providing more stable cross-view geometry to mitigate point-set variation.
(c.2) \textit{Anchor-to-Policy Readout} decouples the fusion with appearance to guide anchor selection, while combining geometric values for action prediction.}
    \label{fig:motivation}
    \vspace{-4pt}
\end{figure}

One approach to viewpoint generalization in RGB-only deployment is through target-view synthesis~\cite{vista,robonvs} and viewpoint-robust representation learning~\cite{pang2025reviwo,lee2025class}. Synthetic views share the same action labels but do not explicitly constrain how corresponding physical regions support prediction across views, as illustrated in \cref{fig:motivation}(a). 

When only the camera moves, scene geometry remains unchanged in robot coordinates. Therefore, a potential approach is to use geometry-based policies~\cite{ze2024dp3,groot,peract,liu2023framemining} with monocular depth estimated from RGB images~\cite{piccinelli2024unidepth,hu2024metric3dv2}. However, predicting a point cloud does not by itself yield a consistent policy representation. A single view samples only visible surfaces, so camera motion changes the observed points (\cref{fig:motivation}(b)). Therefore, point cloud-based representations produced by existing policies~\cite{ze2024dp3,groot} based on PointNet~\cite{qi2017pointnet} or PointNet++~\cite{qi2017pointnetplusplus} can vary across viewpoints. Pure 3D policies also ignore complementary RGB cues that help identify task-relevant regions.

To address the above limitations, we propose \textit{ReGeoAnch}, a new paradigm for viewpoint generalization in RGB-only deployment via Reading Geometric Anchors. ReGeoAnch follows the geometry-based pipeline, starting by predicting a pseudo point cloud. To provide a common spatial reference across viewpoints, we define the global grid in the robot base and local grid in the gripper frame, providing workspace context and precise gripper-relative geometry, respectively.

Based on these anchors, we propose a two-stage, \textit{Decoupled Appearance-Guided Geometric Readout} mechanism, as illustrated in \cref{fig:motivation}(c), incorporating multimodal fusion and simultaneously mitigating the viewpoint sensitivity. (\romannumeral1) \textit{Point-to-Anchor Aggregation} (PAA) adaptively aggregates observed points around each anchor into geometric features, providing comparable physical evidence without requiring pointwise correspondence. Therefore, PAA encourages more stable cross-view geometry within shared physical regions. (\romannumeral2) \textit{Anchor-to-Policy Readout} (APR) develops a decoupled information flow that controls the view-dependent visual appearance to select task-relevant anchors  for the subsequent action policy. Specifically, APR is based on a decoupled attention mechanism, where the query, i.e., the view-invariant robot state, matches with anchor appearance and other related information, combining the values derived from anchor geometry alone. Consequently, APR restricts how view-sensitive appearance and relatively robust geometric features contribute to the final readout, providing a controllable and interpretable solution to viewpoint generalization. The decoupled nature of APR is different from existing RGB-D fusion methods~\cite{monolift,gp3,act3d,ke2025diffuser,fpvnet}. These methods fuse visual and geometric information directly for representation learning, without sufficiently organizing the distinct roles of modalities that benefit cross-view robustness.

Finally, we introduce \textit{Cross-View Readout Consistency} that aligns final readouts from source and synthetic target views. It allows local observations to differ across viewpoints while encouraging consistent task-relevant conditioning representations for the controller.

Our contributions are summarized as follows:
\begin{itemize}
    \item We introduce ReGeoAnch, a geometry-anchored imitation learning framework for viewpoint generalization from single-view RGB-D demonstrations to calibrated RGB-only deployment.

    \item We combine a lightweight calibration-conditioned depth decoder with shared global and local anchors and a two-stage geometric readout. PAA aggregates predicted geometry at anchors, while APR decouples appearance-guided selection from geometric value content.

    \item We develop a cross-view readout consistency strategy to encourage stable policy conditioning across viewpoints, combined with an efficient view synthesis strategy.
\end{itemize}

\section{Related Work}
\label{sec:related}

\paragraph{Viewpoint generalization in manipulation.}
VISTA~\cite{vista} and RoboNVS~\cite{robonvs} augment demonstrations with synthesized images and videos, respectively.
Camera-conditioned policies~\cite{jiang2026camera} encode viewing rays, while RAVEN~\cite{klee2026raven} uses ray geometry for attention equivariant to joint transformations of the cameras and scene.
Representation learning provides another route: CLASS~\cite{lee2025class} aligns visual features according to the similarity of subsequent action sequences measured by dynamic time warping, and ReViWo~\cite{pang2025reviwo} separates view-dependent and view-invariant representations through cross-view reconstruction and contrastive learning.
GEM3D~\cite{gem3d} learns amodal 3D representations through point-cloud reconstruction and Gaussian-splatting-based novel-view supervision, then distills them into a single-view RGB-D policy.
ReGeoAnch also uses synthesized views, but focuses on aligning geometric readouts from PAA and APR with physical cross-view correspondences. 

\vspace{-5pt}

\paragraph{3D policies.}
DP3~\cite{ze2024dp3} max-pools point features into a scene vector via PointNet, while GROOT~\cite{groot} masks local tokens of object point clouds in PointNet++ to increase viewpoint robustness. To represent hierarchical structures, Frame Mining~\cite{liu2023framemining} exploits complementary robot coordinate frames, while VIHE~\cite{wang2024vihe} and RVT-2~\cite{goyal2024rvt2} render finer virtual views with Transformers. Rather than directly processing a single observed point cloud that may vary across viewpoints, ReGeoAnch uses PAA to represent comparable physical regions at shared anchors across views, further guided by RGB appearance with complementary information.

\vspace{-5pt}

\paragraph{Geometry-aware policies.}
For RGB-only deployment, MonoLift~\cite{monolift} distills an RGB--pseudo-depth teacher into a monocular RGB student, removing depth estimation at deployment, while GP3~\cite{gp3} uses geometric supervision to adapt a pretrained spatial encoder for multi-view RGB observations. 
For RGB-D representations, PerAct~\cite{peract} directly encodes RGB channels with 3D coordinates within voxels. Both Act3D~\cite{act3d} and 3D Diffuser Actor~\cite{ke2025diffuser} lift visual features into 3D space. The resulting 3D visual features are further processed via relative attention as keys and values to predict subsequent poses. FPV-Net~\cite{fpvnet} uses RGB-conditioned adaptive layer normalization to scale and shift point-cloud and language embeddings.

In the above methods, RGB appearance couples with geometry that contributes directly to the features used for action prediction, either through visual features and lifted 3D ones, or visual conditioning of geometric representations. To reduce sensitivity to view-dependent cues, ReGeoAnch uses a two-stage architecture with PAA and APR. This design decouples the roles of visual appearance and geometry. Together with cross-view readout alignment, ReGeoAnch encourages consistent policy conditioning across viewpoints.
\section{Method}
\label{sec:method}

\begin{figure*}[!t]
    \centering
    \includegraphics[pagebox=cropbox,width=\textwidth]{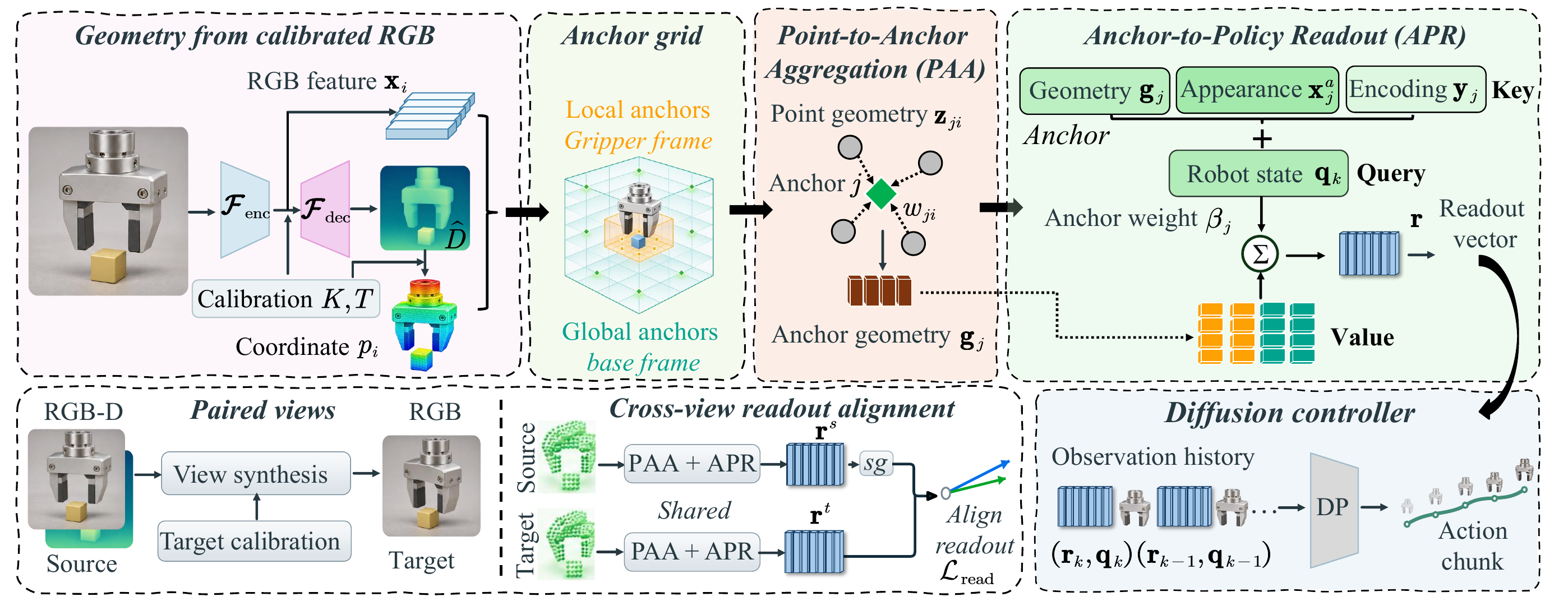}
    \caption{Overall pipeline of ReGeoAnch. At deployment, ReGeoAnch extracts RGB features \(\mathbf x_i\) for pixel \(i\) and predicts a depth map \(\hat D\) from an RGB image \(I\). Using the target camera calibration \((K,T)\), it back-projects depth into a 3D point \(p_i\), as described in Sec.~\ref{sec:method-geometry}. Following Sec.~\ref{sec:method-anchors}, we define global and local anchors in the robot base and gripper frames as shared spatial references. 
PAA encodes point geometry \(\mathbf z_{ji}\) and pools the resulting values with spatial weights to form anchor features \(\mathbf g_j\). APR uses appearance and robot state to weight these features for the policy readout \(\mathbf r\) (Sec.~\ref{sec:method-readout}). The diffusion controller uses readout and robot-state histories to generate an action chunk.
During training, we synthesize target RGB views from source RGB-D observations and target calibration, and align the resulting APR readouts, as described in Sec.~\ref{sec:method-consistency}.}
    \label{fig:method}
    \vspace{-4pt}
\end{figure*}

Our training set \(\mathcal{D}_s\) consists of RGB-D demonstrations recorded from a single calibrated source viewpoint with intrinsics \(K^s\) and camera-to-base pose \(T^s\). Each demonstration contains an RGB image \(I^s_k\), a depth map \(D^s_k\), a robot proprioceptive state \(\mathbf{q}_k\), and an action \(\mathbf a_k\) at step \(k\). \textit{The goal is to train a policy for a target viewpoint whose calibration is known during training, without collecting additional target-view demonstrations}. At deployment, the policy uses target RGB observations, camera calibration, and robot proprioception to predict actions. We use superscript \(s\) to indicate the source viewpoint. Fig.~\ref{fig:method} illustrates the overall pipeline.

\subsection{Depth from a Calibrated RGB View}
\label{sec:method-geometry}

We predict depth to locate observed surfaces in robot coordinates, allowing shared anchors to organize evidence across viewpoints. Our lightweight depth decoder \(\boldsymbol{\mathcal F}_{\mathrm{dec}}\) reuses multiscale RGB features from the policy's ResNet-18 encoder \(\boldsymbol{\mathcal F}_{\mathrm{enc}}\)~\cite{he2016resnet}. At each scale, we condition these features on calibrated rays~\cite{piccinelli2024unidepth} by encoding the camera origin \(o\) and each pixel's viewing direction in the robot base frame using intrinsics \(K\) and camera-to-base pose \(T=(R,o)\). The decoder uses 128-channel residual fusion blocks~\cite{ranftl2021dpt,yang2024depthanythingv2} and a single low-resolution self-attention block to predict optical-axis depth \(\hat D=\boldsymbol{\mathcal F}_{\mathrm{dec}}(\boldsymbol{\mathcal F}_{\mathrm{enc}}(I),K,T)\). For pixel \(u=(u_x,u_y,1)^\top\), depth \(D(u)\) locates a surface point along its ray~\cite{hartley2004multiple},
\begin{equation}
    p(u,D)=o+D(u)\,R K^{-1}u .
    \label{eq:regeo-lift}
\end{equation}
For anchor aggregation, we pair each predicted surface point \(p_i=p(u_i,\hat D)\) with its RGB feature \(\mathbf x_i\), sampled bilinearly from the encoder maps at pixel \(u_i\). Appendix~\ref{sec:decoder-details} describes the decoder and the construction of \(\mathbf x_i\).

\subsection{Global and Local Geometric Anchors}
\label{sec:method-anchors}

As shown in Fig.~\ref{fig:method}, we define a global grid in the robot base frame for workspace context and a finer local grid that follows the current gripper pose to capture nearby geometry. At a given robot state, each anchor represents the same physical neighborhood across viewpoints, although its visible points and pooled features may differ.

Let \(j\) index all anchors and \(f=f(j)\in\{\mathrm{G},\mathrm{L}\}\) identify its global or local grid. The center \(\mathbf c_j\), point coordinate \(p_i^f\), and spacing \(\boldsymbol\Delta_f\) use frame \(f\), with \(p_i^{\mathrm{G}}=p_i\). Superscript \(\mathrm{L}\) denotes gripper-frame coordinates.

Unlike fixed-workspace voxel methods~\cite{peract}, our approach adds a gripper-local grid for finer manipulation and uses these anchors as spatial references.

\subsection{Decoupled Appearance-Guided Geometric Readout}
\label{sec:method-readout}

Scene geometry provides a common reference across viewpoints. Yet single-view reconstruction is partial and potentially inaccurate, while RGB cues help identify task-relevant evidence. Directly incorporating RGB features into values may increase sensitivity to appearance changes. PAA constructs local geometric descriptors using spatial weights, leaving task-dependent selection to APR. APR uses appearance and robot state to select geometric values for action prediction.

\paragraph{Point-to-Anchor Aggregation (PAA).}
Each anchor pools geometry from the points visible in its neighborhood. For anchor \(j\) in grid \(f=f(j)\) and point \(i\) expressed in its grid frame, the spatial weight is
\begin{equation}
    w_{ji}=\tau_{fi}\,\operatorname{Tri}_f(p_i^f,\mathbf c_j).
    \label{eq:regeo-spatial-weight}
\end{equation}
Here \(\operatorname{Tri}_f\) gives the trilinear weight of center \(\mathbf c_j\) based on its offset from the point, normalized over the point's valid neighboring centers (at most eight)~\cite{heinrich2023chasing}. This soft assignment avoids hard switches between single voxels as predicted points move. As visibility changes, each anchor pools available points from the same physical region. The taper \(\tau_{fi}\) reduces support near the boundary, as detailed in Appendix~\ref{sec:attention-details}.

For each contributing pair, point geometry \(\mathbf z_{ji}\) records the gripper-frame position of point \(i\) and its offset from the anchor:
\begin{equation}
    \mathbf z_{ji} =
    \bigl[p_i^{\mathrm{L}},\; p_i^{\mathrm{L}}-\mathbf c_j^{\mathrm{L}},\;
    (p_i^f-\mathbf c_j)\oslash\boldsymbol\Delta_f\bigr]\in\mathbb R^9 .
    \label{eq:regeo-point-geometry}
\end{equation}
The offset is expressed in both gripper coordinates and grid cell widths, with \(\oslash\) denoting componentwise division. With exact geometry and a fixed robot state, the descriptor of a point relative to the same anchor is unchanged across viewpoints. For each occupied anchor, PAA pools the encoded point geometry with the spatial weights from \cref{eq:regeo-spatial-weight}:
\begin{equation}
    \mathbf g_j=e_j\,\boldsymbol{\phi}_g\!\left(
      \frac{\sum_i w_{ji}\,\boldsymbol{\phi}_z(\mathbf z_{ji})}
           {\sum_i w_{ji}}\right),
    \qquad e_j=\max_i w_{ji}.
    \label{eq:regeo-anchor-feature}
\end{equation}
Normalized pooling discards the magnitude of spatial support, so weakly supported anchors can still yield full-strength features. We therefore scale each feature by \(e_j\), attenuating it when all contributing points have small spatial weights.
For each occupied anchor, we use the same spatial weights to pool the projected RGB features into an appearance descriptor for the APR key:
\begin{equation}
    \mathbf x_j^a =
    \frac{\sum_i w_{ji}\,\boldsymbol{\phi}_x(\mathbf x_i)}{\sum_i w_{ji}} .
    \label{eq:regeo-anchor-appearance}
\end{equation}
The mappings \(\boldsymbol{\phi}_z\), \(\boldsymbol{\phi}_g\), and \(\boldsymbol{\phi}_x\) are separate two-layer MLPs with GELU activation.

\paragraph{Anchor-to-Policy Readout (APR).}
PAA provides potentially more viewpoint-stable geometric features for each anchor. APR selects anchors for action prediction by assigning distinct roles to appearance and geometry. 

The core of APR is an attention mechanism: at step \(k\), APR derives a single query from the robot state \(\mathbf q_k\), which is independent of the viewpoint and changes as manipulation progresses. The query is matched against each candidate anchor's key, which combines its appearance, geometry, and encoding. These features provide task-relevant visual cues and distinguish similar-looking regions by their shape and location relative to the gripper. Finally, the value carries geometry into the readout:
\begin{equation}
    \mathbf Q^a=\boldsymbol{\psi}_Q(\mathbf q_k),
    \mathbf K_j^a=\boldsymbol{\psi}_K(\mathbf g_j,\mathbf x_j^a,\mathbf y_j),
    \mathbf v_j^a=\boldsymbol{\psi}_V(\mathbf g_j).
    \label{eq:regeo-apr-qkv}
\end{equation}
Here \(\mathbf y_j\) encodes \((\mathbf c_j^{\mathrm{L}},f,\boldsymbol\Delta_f)\). For each head of width \(d_h\), APR combines spatial support with scaled dot-product attention~\cite{vaswani2017attention}:
\begin{gather}
    \beta_{j}=
    \frac{e_j\,\exp\!\bigl(\mathbf Q^a\cdot\mathbf K_j^a/\sqrt{d_h}\bigr)}
         {1+\sum_{j'} e_{j'}\,\exp\!\bigl(\mathbf Q^a\cdot\mathbf K_{j'}^a/\sqrt{d_h}\bigr)},
    \;\mathbf m_{\mathrm r}=\sum_j\beta_{j}\mathbf v^a_{j}.
    \label{eq:regeo-policy}
\end{gather}
The constant \(1\) lets APR assign less total attention to anchors when their spatial support is weak. To retain the strength of observed evidence after head fusion, we scale the output of \(\boldsymbol{\psi}_R\) by the mean attention mass on observed anchors:
\begin{equation}
    \mathbf{r}=
    \operatorname{mean}_h\!\left(\sum_j\beta_{j}^{(h)}\right)
    \boldsymbol{\psi}_R\!\left(\operatorname{concat}_h\mathbf{m}_{\mathrm r}^{(h)}\right).
    \label{eq:regeo-readout}
\end{equation}
 We concatenate each readout with its robot state and condition the diffusion controller~\cite{chi2023diffusion} on the resulting observation history to predict an action chunk, as illustrated in Fig.~\ref{fig:method}.

APR uses affine projections with layer normalization in \(\boldsymbol{\psi}_Q,\boldsymbol{\psi}_K\), a bias-free linear projection in \(\boldsymbol{\psi}_V\), and a separate two-layer MLP with GELU activation within \(\boldsymbol{\psi}_R\). Appendix~\ref{sec:network-architecture} gives implementation details.

\subsection{Cross-View Readout Consistency}
\label{sec:method-consistency}

Two views of the same manipulation step share an action label, yet changes in visibility and appearance can alter the geometric readout used by the controller. We therefore align the source and target readout directions during paired training, rather than requiring all visible features to match.

\begin{figure}[t]
    \centering
    \includegraphics[width=\columnwidth]{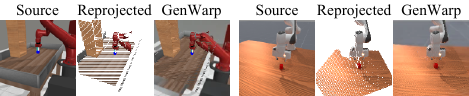}
    \caption{Target-view synthesis for \emph{Shelf Place} (left) and \emph{PickCube} (right). GenWarp~\cite{seo2024genwarp} fills the white regions in each reprojection while preserving the reprojected source colors.}
    \label{fig:genwarp-pair}
    \vspace{-4pt}
\end{figure}

\paragraph{Paired observations.}
Measured source depth \(D^s\) and camera calibration support reprojection of valid workspace points from source observation \(s\) into target RGB view \(t\). Pretrained GenWarp~\cite{seo2024genwarp} efficiently completes uncovered regions without additional training, as shown in \cref{fig:genwarp-pair}. GenWarp enables faster view synthesis than VISTA~\cite{vista} and RoboNVS~\cite{robonvs}. Retained reprojections provide the target depth labels \(D^t\) used in \cref{sec:method-objective}.

\paragraph{Readout alignment.}
The shared policy computes \(\mathbf r^s\) and \(\mathbf r^t\) from the geometry predicted in each view at the same robot state. We minimize the cosine distance
\begin{equation}
    \mathcal L_{\mathrm{read}} =
    1-\cos\!\left(\mathbf{r}^t,\operatorname{sg}(\mathbf{r}^s)\right),
    \label{eq:regeo-read}
\end{equation}
averaged over observation frames. After source pretraining, the source readout serves as the reference for this loss, avoiding a direct pull toward artifacts in synthesized target views. The stop-gradient \(\operatorname{sg}\) blocks only its direct source-side gradient. Both views still share model parameters.

\subsection{Training Objective}
\label{sec:method-objective}
Both views share the sampled diffusion step \(\tau\) and Gaussian noise \(\boldsymbol\epsilon\) for each action chunk. Given the readout and robot-state history for view \(\nu \in \left\{s,t\right\}\), the Diffusion Policy head~\cite{chi2023diffusion} outputs \(\hat{\boldsymbol\epsilon}_k^{\nu}(\tau)\), its estimate of the noise \(\boldsymbol\epsilon_k\) added to action \(\mathbf a_k\). We minimize the expected weighted denoising error,
\begin{equation}
    \mathcal L_{\mathrm{act}}
    =\mathbb E_{\tau,\boldsymbol\epsilon}\!\left[
      \tfrac12\sum_{\nu\in\{s,t\}}\frac{\sum_k\omega_k\,
      \operatorname{MSE}(\hat{\boldsymbol\epsilon}_k^{\nu}(\tau),\boldsymbol\epsilon_k)}
      {\sum_k\omega_k}\right].
    \label{eq:regeo-action-loss}
\end{equation}
 The adaptive weighting rule assigns larger \(\omega_k\) to steps from stages with higher source-view denoising MSE over past training updates. Appendix~\ref{sec:action-weighting} gives the details.

We supervise depth in 3D because anchor assignment and geometric values depend on surface positions. At each labeled pixel \(u\), \cref{eq:regeo-lift} gives the 3D error \(d_u^\nu=\lVert p(u,\hat D^\nu)-p(u,D^\nu)\rVert_2\). We supervise valid source depths and retained target reprojections with
\begin{equation}
    \mathcal L_{\mathrm{depth}}
    =\tfrac12\sum_{\nu\in\{s,t\}}\operatorname{mean}_{u}\!
    \left[\operatorname{SmoothL1}(d_u^\nu)\right].
    \label{eq:regeo-depth-loss}
\end{equation}
The mean over valid pixels is weighted, with greater weight on source workspace points as detailed in Appendix~\ref{sec:cross-view-supervision-details}.

We first pretrain on source demonstrations using action and depth supervision. During paired training, we add readout consistency:
\begin{equation}
    \mathcal L =
    \mathcal L_{\mathrm{act}} + \lambda_d\mathcal L_{\mathrm{depth}}
    + \lambda_r\mathcal L_{\mathrm{read}} .
    \label{eq:regeo-total}
\end{equation}

\section{Experiments}
\label{sec:experiments}
\begingroup
\makeatletter
\renewcommand\paragraph{\@startsection{paragraph}{4}{\z@}%
  {3ex}{-1em}{\normalfont\normalsize\bfseries}}
\makeatother
\setlength{\textfloatsep}{8pt plus 1pt minus 2pt}
\setlength{\intextsep}{8pt plus 1pt minus 2pt}
\setlength{\floatsep}{6pt plus 1pt minus 2pt}

\begin{figure}[!t]
    \centering
    \includegraphics[width=\columnwidth]{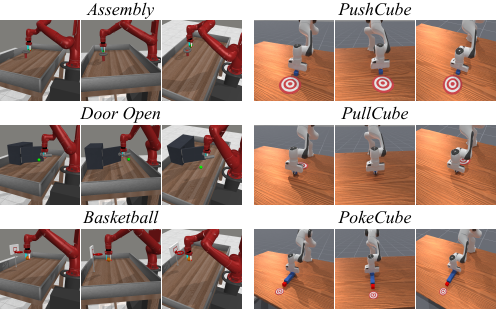}
    \caption{Manipulation tasks and evaluation viewpoints. Each triplet shows the same state at the source viewpoint, after a $-30^\circ$ azimuth shift, and after a $+20^\circ$ elevation shift, from left to right.}
    \label{fig:task-views}
    \vspace{-1pt}
\end{figure}

\begin{table*}[t]
    \centering
    \caption{Target-view success rates (\%) for azimuth and elevation changes on MetaWorld, and for azimuth, elevation, and distance-scale changes on ManiSkill3. \textbf{Bold} and \underline{underlined} entries indicate the best and second-best results, including ties.}
    \label{tab:main-results}
    \begingroup
    \small
    \setlength{\tabcolsep}{0.5pt}
    \renewcommand{\arraystretch}{1.02}
    \begin{tabular*}{0.8\textwidth}{@{\extracolsep{\fill}}lc*{15}{r}@{}}
    \toprule
    & & \multicolumn{9}{c}{MetaWorld} & \multicolumn{5}{c}{ManiSkill3} & \\
    \cmidrule(lr){3-11}\cmidrule(l){12-16}
    \textbf{Method} & Year & \makebox[20.5pt][c]{\textit{\shortstack{Bask.\\ball}}} & \makebox[20.5pt][c]{\textit{\shortstack{Bin\\Pick}}} & \makebox[20.5pt][c]{\textit{\shortstack{Pick\\Place}}}
    & \makebox[20.5pt][c]{\textit{\shortstack{Shelf\\Place}}} & \makebox[20.5pt][c]{\textit{\shortstack{Coff.\\Btn.}}} & \makebox[20.5pt][c]{\textit{\shortstack{Door\\Open}}}
    & \makebox[20.5pt][c]{\textit{\shortstack{Win.\\Close}}} & \makebox[20.5pt][c]{\textit{Asm.}} & \makebox[20.5pt][c]{\textbf{Avg.}}
    & \makebox[20.5pt][c]{\textit{\shortstack{Pick\\Cube}}} & \makebox[20.5pt][c]{\textit{\shortstack{Push\\Cube}}} & \makebox[20.5pt][c]{\textit{\shortstack{Pull\\Cube}}}
    & \makebox[20.5pt][c]{\textit{\shortstack{Poke\\Cube}}} & \makebox[20.5pt][c]{\textbf{Avg.}} & \makebox[20.5pt][c]{\textbf{All}} \\
    \midrule
    DP~\cite{chi2023diffusion} & 2023 & \makebox[20.5pt][r]{3.5} & \makebox[20.5pt][r]{6.4} & \makebox[20.5pt][r]{7.8} & \makebox[20.5pt][r]{2.3} & \makebox[20.5pt][r]{84.7} & \makebox[20.5pt][r]{44.6} & \makebox[20.5pt][r]{90.6} & \makebox[20.5pt][r]{2.1} & \makebox[20.5pt][r]{30.3} & \makebox[20.5pt][r]{53.8} & \makebox[20.5pt][r]{64.7} & \makebox[20.5pt][r]{58.3} & \makebox[20.5pt][r]{33.8} & \makebox[20.5pt][r]{52.7} & \makebox[20.5pt][r]{41.5} \\
    DP\({}^{\dagger}\)~\cite{chi2023diffusion} & 2023 & \makebox[20.5pt][r]{23.9} & \makebox[20.5pt][r]{62.3} & \makebox[20.5pt][r]{38.6} & \makebox[20.5pt][r]{24.0} & \makebox[20.5pt][r]{94.1} & \makebox[20.5pt][r]{95.6} & \makebox[20.5pt][r]{\underline{98.6}} & \makebox[20.5pt][r]{62.7} & \makebox[20.5pt][r]{62.5} & \makebox[20.5pt][r]{79.6} & \makebox[20.5pt][r]{88.0} & \makebox[20.5pt][r]{94.4} & \makebox[20.5pt][r]{84.0} & \makebox[20.5pt][r]{86.5} & \makebox[20.5pt][r]{74.5} \\
    DP3~\cite{ze2024dp3} & 2024 & \makebox[20.5pt][r]{23.5} & \makebox[20.5pt][r]{\underline{85.0}} & \makebox[20.5pt][r]{61.5} & \makebox[20.5pt][r]{42.1} & \makebox[20.5pt][r]{99.6} & \makebox[20.5pt][r]{\underline{99.3}} & \makebox[20.5pt][r]{\textbf{100.0}} & \makebox[20.5pt][r]{63.7} & \makebox[20.5pt][r]{71.8} & \makebox[20.5pt][r]{60.0} & \makebox[20.5pt][r]{76.9} & \makebox[20.5pt][r]{92.8} & \makebox[20.5pt][r]{61.7} & \makebox[20.5pt][r]{72.9} & \makebox[20.5pt][r]{72.3} \\
    3D-DA~\cite{ke2025diffuser} & 2025 & \makebox[20.5pt][r]{19.8} & \makebox[20.5pt][r]{82.4} & \makebox[20.5pt][r]{58.5} & \makebox[20.5pt][r]{43.7} & \makebox[20.5pt][r]{\underline{99.9}} & \makebox[20.5pt][r]{97.4} & \makebox[20.5pt][r]{\textbf{100.0}} & \makebox[20.5pt][r]{79.1} & \makebox[20.5pt][r]{72.6} & \makebox[20.5pt][r]{84.8} & \makebox[20.5pt][r]{81.6} & \makebox[20.5pt][r]{94.8} & \makebox[20.5pt][r]{85.9} & \makebox[20.5pt][r]{86.8} & \makebox[20.5pt][r]{\underline{79.7}} \\
    ReViWo~\cite{pang2025reviwo} & 2025 & \makebox[20.5pt][r]{\underline{42.5}} & \makebox[20.5pt][r]{83.1} & \makebox[20.5pt][r]{48.3} & \makebox[20.5pt][r]{46.7} & \makebox[20.5pt][r]{\textbf{100.0}} & \makebox[20.5pt][r]{98.8} & \makebox[20.5pt][r]{\textbf{100.0}} & \makebox[20.5pt][r]{73.9} & \makebox[20.5pt][r]{74.2} & \makebox[20.5pt][r]{32.2} & \makebox[20.5pt][r]{47.2} & \makebox[20.5pt][r]{54.0} & \makebox[20.5pt][r]{42.2} & \makebox[20.5pt][r]{43.9} & \makebox[20.5pt][r]{59.0} \\
    CLASS~\cite{lee2025class} & 2025 & \makebox[20.5pt][r]{29.1} & \makebox[20.5pt][r]{46.6} & \makebox[20.5pt][r]{\underline{79.9}} & \makebox[20.5pt][r]{\underline{55.7}} & \makebox[20.5pt][r]{\textbf{100.0}} & \makebox[20.5pt][r]{91.0} & \makebox[20.5pt][r]{\textbf{100.0}} & \makebox[20.5pt][r]{\textbf{97.3}} & \makebox[20.5pt][r]{\underline{74.9}} & \makebox[20.5pt][r]{66.7} & \makebox[20.5pt][r]{84.4} & \makebox[20.5pt][r]{94.1} & \makebox[20.5pt][r]{65.0} & \makebox[20.5pt][r]{77.6} & \makebox[20.5pt][r]{76.3} \\
    Cam-DP~\cite{jiang2026camera} & 2026 & \makebox[20.5pt][r]{21.8} & \makebox[20.5pt][r]{76.0} & \makebox[20.5pt][r]{57.3} & \makebox[20.5pt][r]{39.8} & \makebox[20.5pt][r]{99.7} & \makebox[20.5pt][r]{98.0} & \makebox[20.5pt][r]{\textbf{100.0}} & \makebox[20.5pt][r]{85.3} & \makebox[20.5pt][r]{72.2} & \makebox[20.5pt][r]{82.3} & \makebox[20.5pt][r]{86.5} & \makebox[20.5pt][r]{93.5} & \makebox[20.5pt][r]{81.6} & \makebox[20.5pt][r]{86.0} & \makebox[20.5pt][r]{79.1} \\
    RAVEN~\cite{klee2026raven} & 2026 & \makebox[20.5pt][r]{21.7} & \makebox[20.5pt][r]{14.0} & \makebox[20.5pt][r]{\textbf{80.5}} & \makebox[20.5pt][r]{39.8} & \makebox[20.5pt][r]{\textbf{100.0}} & \makebox[20.5pt][r]{89.6} & \makebox[20.5pt][r]{97.5} & \makebox[20.5pt][r]{57.5} & \makebox[20.5pt][r]{62.6} & \makebox[20.5pt][r]{\underline{90.5}} & \makebox[20.5pt][r]{\underline{89.4}} & \makebox[20.5pt][r]{\underline{96.8}} & \makebox[20.5pt][r]{\textbf{91.0}} & \makebox[20.5pt][r]{\underline{91.9}} & \makebox[20.5pt][r]{77.2} \\
    \midrule
    \textbf{ReGeoAnch} & \textemdash & \makebox[20.5pt][r]{\textbf{69.7}} & \makebox[20.5pt][r]{\textbf{91.2}} & \makebox[20.5pt][r]{68.8} & \makebox[20.5pt][r]{\textbf{72.2}} & \makebox[20.5pt][r]{\textbf{100.0}} & \makebox[20.5pt][r]{\textbf{100.0}} & \makebox[20.5pt][r]{\textbf{100.0}} & \makebox[20.5pt][r]{\underline{91.3}} & \makebox[20.5pt][r]{\textbf{86.6}} & \makebox[20.5pt][r]{\textbf{95.8}} & \makebox[20.5pt][r]{\textbf{97.3}} & \makebox[20.5pt][r]{\textbf{99.9}} & \makebox[20.5pt][r]{\underline{89.4}} & \makebox[20.5pt][r]{\textbf{95.6}} & \makebox[20.5pt][r]{\textbf{91.1}} \\
    \bottomrule
    \end{tabular*}
    \endgroup
    \vspace{-5pt}
\end{table*}

\subsection{Experimental Setup}
\label{sec:experimental-setup}

\vspace{-1pt}
\paragraph{Tasks.}
We evaluate eight MetaWorld~\cite{metaworld} tasks in MuJoCo and four ManiSkill3~\cite{maniskill3} tasks in SAPIEN. MetaWorld uses a 7-DoF Sawyer arm, while ManiSkill3 uses a 7-DoF Franka Emika Panda, both with parallel-finger grippers. Actions specify 3D end-effector displacements and a gripper command. The MetaWorld tasks cover grasping and placement in \emph{Basketball}, \emph{Bin Picking}, \emph{Pick Place}, and \emph{Shelf Place}; button pressing, door opening, and window closing in \emph{Coffee Button}, \emph{Door Open}, and \emph{Window Close}; and fitting a nut onto a peg in \emph{Assembly}. In ManiSkill3, \emph{PickCube} requires grasping and moving a cube to a 3D goal, while \emph{PushCube} and \emph{PullCube} require pushing and pulling it to a tabletop target. \emph{PokeCube} requires grasping a peg and using it to push a cube to its target. Figure~\ref{fig:task-views} shows six tasks from source and evaluation viewpoints. We use 400 demonstrations per task. The image resolution is \(128\times128\).
\vspace{-6pt}

\begin{table}[tb]
    \centering
    \caption{Mean success rates (\%) by camera change.}
    \label{tab:axis-summary}
    \begingroup
    \small
    \setlength{\tabcolsep}{1pt}
    \renewcommand{\arraystretch}{1.02}
    \begin{tabular*}{0.8\columnwidth}{@{\extracolsep{\fill}}l*{5}{c}@{}}
    \toprule
    & \multicolumn{2}{c}{MetaWorld} & \multicolumn{3}{c}{ManiSkill3} \\
    \cmidrule(lr){2-3}\cmidrule(l){4-6}
    \textbf{Method} & \makebox[24pt][c]{Azim.} & \makebox[24pt][c]{Elev.}
    & \makebox[24pt][c]{Azim.} & \makebox[24pt][c]{Elev.} & \makebox[24pt][c]{Scale} \\
    \midrule
    DP~\cite{chi2023diffusion} & \makebox[24pt][c]{32.3} & \makebox[24pt][c]{28.2} & \makebox[24pt][c]{42.0} & \makebox[24pt][c]{51.3} & \makebox[24pt][c]{64.6} \\
    DP\({}^{\dagger}\)~\cite{chi2023diffusion} & \makebox[24pt][c]{66.9} & \makebox[24pt][c]{58.0} & \makebox[24pt][c]{81.2} & \makebox[24pt][c]{85.3} & \makebox[24pt][c]{93.0} \\
    DP3~\cite{ze2024dp3} & \makebox[24pt][c]{\underline{74.9}} & \makebox[24pt][c]{68.8} & \makebox[24pt][c]{74.5} & \makebox[24pt][c]{79.6} & \makebox[24pt][c]{64.5} \\
    3D-DA~\cite{ke2025diffuser} & \makebox[24pt][c]{73.4} & \makebox[24pt][c]{71.9} & \makebox[24pt][c]{80.4} & \makebox[24pt][c]{86.2} & \makebox[24pt][c]{93.8} \\
    ReViWo~\cite{pang2025reviwo} & \makebox[24pt][c]{74.7} & \makebox[24pt][c]{73.6} & \makebox[24pt][c]{28.4} & \makebox[24pt][c]{38.0} & \makebox[24pt][c]{65.3} \\
    CLASS~\cite{lee2025class} & \makebox[24pt][c]{73.7} & \makebox[24pt][c]{\underline{76.2}} & \makebox[24pt][c]{71.7} & \makebox[24pt][c]{67.3} & \makebox[24pt][c]{93.7} \\
    Cam-DP~\cite{jiang2026camera} & \makebox[24pt][c]{71.7} & \makebox[24pt][c]{72.7} & \makebox[24pt][c]{78.7} & \makebox[24pt][c]{86.7} & \makebox[24pt][c]{92.5} \\
    RAVEN~\cite{klee2026raven} & \makebox[24pt][c]{56.5} & \makebox[24pt][c]{68.6} & \makebox[24pt][c]{\underline{83.6}} & \makebox[24pt][c]{\underline{92.4}} & \makebox[24pt][c]{\textbf{99.8}} \\
    \midrule
    \textbf{ReGeoAnch} & \makebox[24pt][c]{\textbf{92.0}} & \makebox[24pt][c]{\textbf{81.3}} & \makebox[24pt][c]{\textbf{94.5}} & \makebox[24pt][c]{\textbf{94.4}} & \makebox[24pt][c]{\underline{98.0}} \\
    \bottomrule
    \end{tabular*}
    \endgroup
    \vspace{-3pt}
\end{table}

\vspace{-3pt}
\paragraph{Evaluation protocol.}
We evaluate azimuth, elevation, and distance changes independently from the source view. For each camera setting, we report the success rate over 100 rollouts, all using the same fixed camera pose. Unless stated otherwise, mean success rates exclude \(0^\circ\) offsets and scale \(1\). Source-view results are reported separately in Table~\ref{tab:source-target-summary}. All methods use the task budgets in Table~\ref{tab:episode-budgets}.
(1) \emph{Elevation.} We vary elevation in \(5^\circ\) increments from \(0^\circ\) to \(+30^\circ\) on MetaWorld and from \(-10^\circ\) to \(+20^\circ\) on ManiSkill3.
(2) \emph{Azimuth.} On both benchmarks, we test offsets of \(0^\circ,-5^\circ,-10^\circ,-15^\circ,-20^\circ\), and \(-30^\circ\).
(3) \emph{Distance scale.} On ManiSkill3, we test camera distances of \(\{0.95,0.975,1,1.05,1.1,1.15,1.2\}\) times the source distance from the look-at point, keeping others fixed.
\vspace{-10pt}

\paragraph{Baselines.}
(1) \emph{Diffusion Policy (DP)}~\cite{chi2023diffusion} trains on the original source-view demonstrations.
(2) \emph{DP\({}^{\dagger}\)} trains the RGB policy on both source views and target views synthesized by GenWarp~\cite{seo2024genwarp} to measure the effect of view augmentation.
(3) \emph{Camera-conditioned DP (Cam-DP)}~\cite{jiang2026camera} encodes each RGB image together with its calibrated viewing rays.
(4) \emph{RAVEN (RAV.)}~\cite{klee2026raven} uses calibrated camera rays in its equivariant attention layers and predicts actions with a flow decoder.
(5) \emph{DP3}~\cite{ze2024dp3} encodes a point cloud constructed from predicted depth for RGB-only deployment, using the same depth-decoder architecture as ReGeoAnch (Sec.~\ref{sec:method-geometry}).
(6) \emph{3D Diffuser Actor (3D-DA)}~\cite{ke2025diffuser} uses depth-lifted visual features and 3D relative attention with the same predicted-depth decoder and Diffusion Policy controller as ours.
(7) For \emph{ReViWo (ReV.)}~\cite{pang2025reviwo}, we pair its pretrained encoder with a diffusion controller and fine-tune the encoder during policy training.
(8) For \emph{CLASS (CLS.)}~\cite{lee2025class}, we pretrain the diffusion controller's visual encoder by aligning features of observations followed by similar action sequences.

Within each task, ReGeoAnch and all baselines except RAVEN use the same DP controller implementation, action horizon, and diffusion schedule. Conditioning widths depend on the visual representation.

\vspace{-4pt}
\paragraph{Training details.}
We train ReGeoAnch on NVIDIA GeForce RTX 5090 GPUs using AdamW with a batch size of 128. The standard schedule uses 30k source-view updates followed by 20k paired-view updates. During paired training, the learning rate is \(10^{-4}\), with \(10^{-5}\) for the MetaWorld RGB backbone. The global and local grids use \(12\times9\times9=972\) and \(16^3=4096\) anchors, respectively (5,068 total). Paired training uses \(\lambda_d=\lambda_r=1\).

To control for gains from additional synthesized training images, all baselines except standard DP reuse the same view pairs and action labels as ReGeoAnch. DP\({}^{\dagger}\) follows the same staged schedule, and standard DP uses the 30k source-view stage. DP3 and 3D-DA use the same depth-decoder architecture and depth supervision as ReGeoAnch. Cam-DP, RAVEN, DP3, 3D-DA, ReViWo, and CLASS receive 50k joint source/target updates. ReViWo initializes its encoder from official pretrained weights. CLASS receives 50 epochs of contrastive pretraining on the task's paired images.

A single augmented policy is trained per task and camera-change axis using multiple GenWarp camera poses, then evaluated across all test poses. We also interpolate between source and target poses.

\subsection{Main Results}
\label{sec:quantitative-results}
\label{sec:viewpoint-results}

\vspace{1pt}
\paragraph{Viewpoint robustness.}
Table~\ref{tab:main-results} shows that ReGeoAnch reaches 86.6\% on MetaWorld and 95.6\% on ManiSkill3, with 91.1\% overall versus 79.7\% for 3D-DA, the strongest baseline overall. It obtains the best or joint-best score on nine of twelve tasks. ReGeoAnch also improves on DP\({}^{\dagger}\) by 16.6 percentage points using the same GenWarp views, indicating benefits from overall architecture design beyond view augmentation alone. With the same depth-decoder architecture and supervision, ReGeoAnch outperforms DP3 and 3D-DA. The latter reaches 72.6\% on MetaWorld and 86.8\% on ManiSkill3. These comparisons support the benefit of organizing and reading geometry through shared anchors. Table~\ref{tab:axis-summary} shows that ReGeoAnch maintains consistently strong average performance under diverse viewpoint changes, including azimuth, elevation, and distance scale. Figure~\ref{fig:viewpoint-curves} further details its performance across individual angular offsets. Per-view results are provided in Appendix~\ref{sec:supp-detailed-view-results}.
\begin{figure}[!t]
    \centering
    \includegraphics[width=\columnwidth]{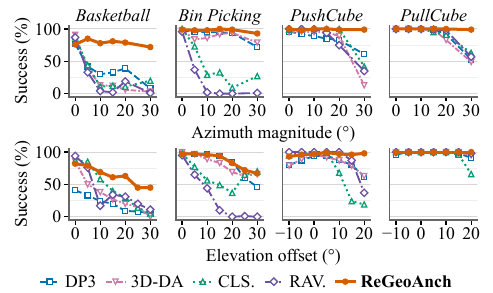}
    \caption{Success rates across camera viewpoints. Top: azimuth magnitudes; bottom: elevation offsets.}
    \label{fig:viewpoint-curves}
    \vspace{-4pt}
\end{figure}

\suppressfloats[t]
\begingroup
\setlength{\intextsep}{6pt}
\begin{figure}[t]
    \centering
    \includegraphics[width=0.95\columnwidth]{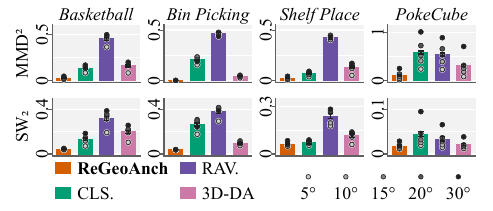}
    \vspace{-3pt}
    \caption{Action-distribution stability across viewpoints. Bars average non-source views with 95\% confidence intervals; dots show view means, shaded by azimuth magnitude.}
    \label{fig:action-distribution}
    \vspace{-1pt}
\end{figure}
\endgroup

\vspace{-8pt}
\paragraph{Action-distribution stability.}
To assess the effect of viewpoint changes on action prediction, we additionally use the source view at inference solely for diagnosis. For each trained policy, we keep the physical state fixed and repeatedly sample action chunks under source and target observations. We compare the resulting action distributions using squared maximum mean discrepancy (MMD$^2$) and sliced Wasserstein-2 (SW$_2$) for four tasks, as illustrated in Fig. \ref{fig:action-distribution}. On the four displayed tasks, ReGeoAnch yields lower mean MMD$^2$ and SW$_2$ than all compared methods, indicating more stable action generation across viewpoints. CLASS aligns visual features through action-sequence similarity, but does not explicitly constrain how corresponding physical evidence is aggregated. ReGeoAnch instead reads shared geometric anchors via PAA and APR with final readout alignment across views, encouraging more stable use of task-relevant geometry.
Figure~\ref{fig:supp-action-distribution-curves} gives the angular trends for more evaluated tasks.

\vspace{-5pt}
\paragraph{Success under control-step budgets.}
At 60\% and 80\% of the task step limits, ReGeoAnch leads on MetaWorld (64.1\%, 85.6\%) and ManiSkill3 (52.1\%, 90.4\%), respectively. See Fig.~\ref{fig:supp-completion-budget-bars} for details.

\vspace{-5pt}
\paragraph{Viewpoint extrapolation.}
Table~\ref{tab:maniskill-extrapolation-axes} shows ReGeoAnch leading ManiSkill3 baselines in mean success at \(+30^\circ\) elevation and \(-30^\circ\) azimuth. Neither viewpoint is used during training.

\vspace{-5pt}
\paragraph{Qualitative comparison.}
Figure~\ref{fig:qualitative-comparison} compares rollouts with matched initial states on \emph{Basketball} and \emph{PushCube}. ReGeoAnch succeeds in both selected cases, while Cam-DP and RAVEN fail within the same task budgets. Additional rollout videos are available in the supplementary material.

\begin{figure}[!t]
    \centering
    \includegraphics[width=0.95\columnwidth]{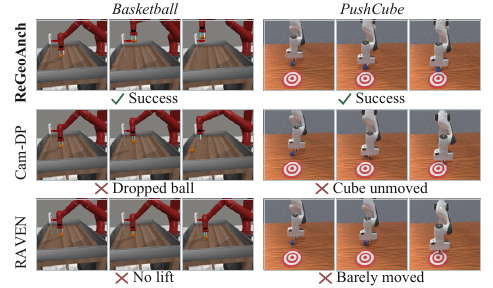}
    \caption{Rollouts at a \(-30^\circ\) azimuth offset. Within each task, methods share the initial state and frames progress left to right. }
    \label{fig:qualitative-comparison}
    \vspace{-1pt}
\end{figure}

\subsection{Analysis}
\label{sec:analysis}

\vspace{1pt}
\paragraph{Component ablations.}
Table~\ref{tab:loss-structure-ablation} shows that learned anchor selection has the largest effect on target-view success across six tasks. Fixed APR weights anchors by spatial support alone, ignoring the learned query-key scores in \cref{eq:regeo-policy}. Under the same readout consistency loss, success drops from 86.3\% to 59.4\%, indicating that alignment cannot replace learned selection within this adaptation budget. Global-only anchors retain most of the full model's performance and outperform local-only anchors. Their broader coverage of task-relevant geometry beyond the gripper neighborhood contributes to this gap. Cross-view readout alignment further raises success from 83.0\% to 86.8\%.

\begin{table}[!htb]
    \centering
    \caption{Ablations on six azimuth tasks of MetaWorld and ManiSkill3.}
    \label{tab:loss-structure-ablation}
    \vspace{-3pt}
    \small
    \setlength{\tabcolsep}{0pt}
    \begin{tabular*}{.87\columnwidth}{@{\extracolsep{\fill}}ccccr@{}}
        \toprule
        $\mathcal L_{\mathrm{read}}$ & Global grid & Local grid & APR & Success (\%) \\
        \midrule
        $\times$ & $\checkmark$ & $\checkmark$ & $\checkmark$ & 83.0 \\
        $\checkmark$ & $\checkmark$ & $\times$ & $\checkmark$ & 85.9 \\
        $\checkmark$ & $\times$ & $\checkmark$ & $\checkmark$ & 76.3 \\
        $\checkmark$ & $\checkmark$ & $\checkmark$ & $\times$ & 59.4 \\
        \midrule
        $\pmb{\checkmark}$ & $\pmb{\checkmark}$ & $\pmb{\checkmark}$ & $\pmb{\checkmark}$ & \textbf{86.8} \\
        \bottomrule
    \end{tabular*}
    \vspace{-4pt}
\end{table}

\vspace{-5pt}
\paragraph{APR design choices.}
Table~\ref{tab:apr-design-choices} compares APR key and value features on four ManiSkill3 tasks. With geometric values retained, removing the direct RGB input from the key lowers mean success from 94.5\% to 86.5\%, suggesting that appearance helps select task-relevant anchors. With RGB retained in the key, replacing geometric values with appearance lowers success to 80.1\%, supporting the role of geometry in providing stable spatial information across different viewpoints for action prediction.

\begingroup
\setlength{\intextsep}{6pt}
\begin{table}[!htb]
    \centering
    \caption{APR key--value ablations on ManiSkill3 azimuths.}
    \label{tab:apr-design-choices}
    \vspace{-3pt}
    \small
    \setlength{\tabcolsep}{0.5pt}
    \renewcommand{\arraystretch}{1.0}
    \begin{tabular*}{0.78\columnwidth}{@{\extracolsep{\fill}}clr@{}}
        \toprule
        RGB in key & Value features & Success (\%) \\
        \midrule
        $\checkmark$ & Geometry & \textbf{94.5} \\
        $\times$ & Geometry & 86.5 \\
        $\checkmark$ & Appearance & 80.1 \\
        \bottomrule
    \end{tabular*}
    \vspace{-2pt}
\end{table}
\endgroup

\vspace{-4pt}
\paragraph{Policy interventions.}
We apply four interventions throughout inference on frozen policies across MetaWorld target azimuths (Table~\ref{tab:attention-interventions}).
(A) \emph{Fixed query.} APR uses its initial query throughout each rollout.
(B) \emph{Swapped keys.} We exchange complete APR keys between anchors, keeping their geometric values in place.
(C) \emph{Shared appearance.} Each anchor in APR receives the same image-wide mean RGB feature.
(D) \emph{Uniform point pooling.} In contrast to PAA, we pool points uniformly within each anchor's neighborhood.
Fixing queries or disrupting key-value correspondence causes the largest success drops. The combined appearance intervention also reduces success substantially, while uniform point pooling has a smaller effect.

\begingroup
\setlength{\intextsep}{6pt}
\begin{table}[!htb]
    \centering
    \caption{MetaWorld azimuth interventions; full policy: \textbf{92.0}\%.}
    \label{tab:attention-interventions}
    \vspace{-3pt}
    \small
    \setlength{\tabcolsep}{0.5pt}
    \renewcommand{\arraystretch}{1.0}
    \begin{tabular*}{\columnwidth}{@{\extracolsep{\fill}}lr@{\hspace{1pt}}||@{\hspace{1pt}}lr@{}}
    \toprule
    \textbf{Intervention} & Success (\%) & \textbf{Intervention} & Success (\%) \\
    \midrule
    A: Fixed query & 28.8 & C: Shared appearance & 54.7 \\
    B: Swapped keys & 32.8 & D: Uniform pooling & \underline{87.6} \\
    \bottomrule
    \end{tabular*}
    \vspace{-4pt}
\end{table}
\endgroup

\vspace{-5pt}
\paragraph{Decoupled readout versus MLP fusion.}
We replace APR with a per-anchor MLP that jointly processes geometry, appearance, position, and robot state, then averages its outputs over valid anchors. This replaces APR's explicit separation of appearance-guided selection and geometric values with joint feature fusion. The MLP only achieves 56.5\% target-view success, compared with 94.5\% for APR (Table~6). Together with the key-value ablations, this result supports decoupling appearance-guided selection from geometric content as a key design for viewpoint generalization.

\begingroup
\setlength{\intextsep}{6pt}
\begin{table}[!htb]
    \centering
    \caption{APR versus anchor MLP on ManiSkill3 azimuths.}
    \label{tab:anchor-mlp}
    \vspace{-3pt}
    \small
    \setlength{\tabcolsep}{0.5pt}
    \renewcommand{\arraystretch}{1.0}
    \begin{tabular*}{0.72\columnwidth}{@{\extracolsep{\fill}}lcc@{}}
        \toprule
        \textbf{Readout} & APR & Anchor MLP \\
        \midrule
        Success (\%) & \textbf{94.5} & 56.5 \\
        \bottomrule
    \end{tabular*}
    \vspace{-4pt}
\end{table}
\endgroup

\vspace{-4pt}
\paragraph{Analysis of depth prediction and view synthesis.}
Figure~\ref{fig:perception-fidelity} examines depth and view-synthesis errors on \textit{PickCube} at a $-30^\circ$ azimuth offset. Against ground truth, the workspace mean absolute error (MAE) is 8.32\,mm for depth predicted from target RGB and 0.11 for RGB synthesized from source RGB-D on a $[0,1]$ scale. Despite imperfect synthesized training views and predicted depth at deployment, the policy achieves 94\% success using simulator target RGB at this viewpoint (Table~\ref{tab:supp-ms-angles}). Together with the ablations, these results support task-relevant geometric readout  rather than error-free reconstruction. Figures~\ref{fig:supp-depth-prediction} and~\ref{fig:supp-view-synthesis} provide further visualizations. 

\begingroup
\setlength{\intextsep}{6pt}
\begin{figure}[!htb]
    \centering
    \includegraphics[width=0.9\columnwidth]{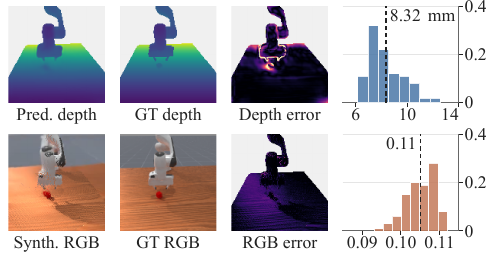}
    \vspace{-3pt}
    \caption{Predicted depth and synthesized RGB versus ground-truth depth and simulator RGB on \emph{PickCube} at a \(-30^\circ\) azimuth offset. Histograms show per-trajectory MAE (depth in mm); dashed lines mark means.}
    \label{fig:perception-fidelity}
    \vspace{-1pt}
\end{figure}
\endgroup

\vspace{-4pt}
\paragraph{Calibration robustness.}
We perturb the calibration supplied to frozen policies while keeping the physical camera fixed. We jointly increase translation, rotation, and focal-length errors from $(0.1\,\mathrm{mm},0.01^\circ,0.02\%)$ to $(2\,\mathrm{mm},0.2^\circ,0.5\%)$ over five levels (L1--L5). Across target azimuths, mean success remains at least 93.5\%, within 1.0 percentage point of the 94.5\% with exact calibration, indicating robustness to the tested errors (Table~\ref{tab:calibration-sensitivity}).

\begingroup
\setlength{\intextsep}{6pt}
\begin{table}[!htb]
    \centering
    \caption{Calibration sensitivity on ManiSkill3 azimuths.}
    \label{tab:calibration-sensitivity}
    \vspace{-3pt}
    \small
    \setlength{\tabcolsep}{0.5pt}
    \renewcommand{\arraystretch}{1.0}
    \begin{tabular*}{0.78\columnwidth}{@{\extracolsep{\fill}}p{54pt}*{6}{c}@{}}
    \toprule
    \textbf{Setting} & Clean & L1 & L2 & L3 & L4 & L5 \\
    \midrule
    Success (\%) & \textbf{94.5} & 94.4 & 94.1 & 94.2 & 94.4 & 93.5 \\
    \bottomrule
    \end{tabular*}
    \vspace{-4pt}
\end{table}
\endgroup

\vspace{-4pt}
\paragraph{Time analysis.}
ReGeoAnch uses \(21.23\)M parameters: \(11.18\)M for the RGB encoder, \(3.18\)M for depth decoding, \(0.74\)M for anchor representation and readout, and \(6.13\)M for the diffusion head. It predicts an action chunk in \(94.3\,\mathrm{ms}\) on an RTX 5090. Comparisons with baselines are summarized in Table~\ref{tab:efficiency}.

\endgroup

\section{Conclusion}
\label{sec:conclusion}

We presented ReGeoAnch for viewpoint-robust imitation learning from single-view RGB-D demonstrations with calibrated RGB-only deployment. Global and gripper-local anchors aggregate predicted geometry, while appearance guides the readout and cross-view consistency aligns its representations. Across 12 MetaWorld and ManiSkill3 tasks, ReGeoAnch achieves \(91.1\%\) overall success versus \(79.7\%\) for 3D-DA. Ablations support the importance of learned anchor selection and geometric values for viewpoint robustness.

{\small\bibliographystyle{ieeenat_fullname_citeorder}\bibliography{main}}
\clearpage
\setcounter{figure}{0}
\setcounter{table}{0}
\setcounter{equation}{0}
\renewcommand{\thefigure}{S\arabic{figure}}
\renewcommand{\thetable}{S\arabic{table}}
\renewcommand{\theequation}{S\arabic{equation}}
\appendix

\section{Supplementary Method}
\label{sec:readout-details}

ReGeoAnch constructs geometric evidence from calibrated RGB observations, aggregates it at shared anchors, and learns consistent readouts across views. We supplement this pipeline with the feature construction and boundary handling omitted from the main text, followed by the computation of cross-view supervision and adaptive action weights.

\subsection{Depth decoder and appearance features}
\label{sec:decoder-details}

\begin{figure}[!ht]
    \centering
    \includegraphics[width=\columnwidth]{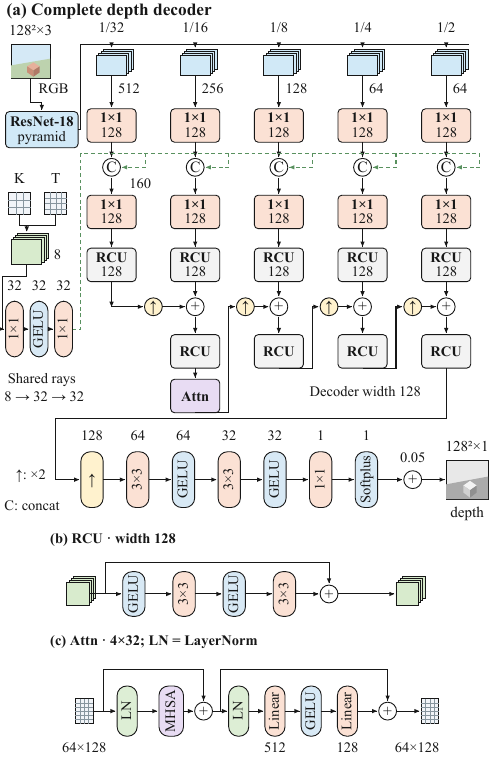}
    \caption{Ray-conditioned depth decoder. (a) Complete graph; kernel labels denote convolutions, C concatenation, and \(\uparrow\) \(2\times\) upsampling. (b) RCU, width 128. (c) Pre-LN four-head self-attention (MHSA) and FFN \(128\!\rightarrow\!512\!\rightarrow\!128\). Input: \(128\times128\) RGB.}
    \label{fig:depth-decoder}
\end{figure}

The decoder in \cref{fig:depth-decoder} incorporates calibration throughout the feature pyramid. At each scale, eight ray channels encode normalized camera coordinates, a unit direction in the robot base frame, and the camera origin. A shared MLP projects these channels for fusion with the corresponding RGB features. Four-head spatial attention follows the stride-16 fusion, and a convolutional head produces the depth map.

Point appearance features retain finer image detail through a full-resolution \(3\times3\) convolution followed by GELU. At each sampled pixel, we concatenate this feature with projected features from the ResNet-18 stem and its first two stages~\cite{he2016resnet}. The deeper pyramid stages contribute to depth decoding.

\subsection{Anchor encoding and support}
\label{sec:attention-details}

The anchor encoder combines the gripper-frame center \(\mathbf c_j^{\mathrm{L}}\), cell widths \(\boldsymbol\Delta_f\), and grid identity \(f=f(j)\in\{\mathrm{G},\mathrm{L}\}\):
\begin{equation}
 \mathbf y_j=\operatorname{MLP}\!\left([
 \operatorname{PE}(\mathbf c_j^{\mathrm{L}}),\,
 \log\boldsymbol\Delta_f]\right)+\mathbf E_f,
 \label{eq:supp-anchor-encoding}
\end{equation}
The anchor encoder uses the same metric convention as the point descriptor in Sec.~\ref{sec:method-readout}. PE concatenates raw coordinates with Fourier features, and the MLP is shared across both grids. \(\mathbf E_f\) is a learnable grid-identity embedding shared by all anchors in grid \(f\).

\paragraph{Trilinear support and boundary taper.}
For a valid point \(p_i^f\), the raw trilinear coefficient of center \(\mathbf c_j\) is
\begin{equation}
  \tilde w_{ji}=\prod_{d=1}^3
    \left(1-\frac{|p^f_{i,d}-c_{j,d}|}{\Delta_{f,d}}\right)_+ ,
  \label{eq:supp-trilinear-raw}
\end{equation}
where \((x)_+=\max(x,0)\). Let \(\mathcal N_f(i)\) contain the in-grid centers with positive coefficients. For \(j\in\mathcal N_f(i)\), we set
\begin{equation}
  \operatorname{Tri}_f(p_i^f,\mathbf c_j)
    =\tilde w_{ji}\Big/\sum_{k\in\mathcal N_f(i)}\tilde w_{ki}.
  \label{eq:supp-trilinear-normalized}
\end{equation}
All other centers receive zero weight. For a point inside grid \(f\), let \(b_{fi,d}\) be its distance to the nearer box face along axis \(d\). The grid taper is \(t_{fi}=\prod_{d=1}^3\min(1,2b_{fi,d}/\Delta_{f,d})\) inside the grid and zero elsewhere. In \cref{eq:regeo-spatial-weight}, global anchors use \(\tau_{fi}=t_{\mathrm{G},i}\), and local anchors use \(\tau_{fi}=t_{\mathrm{G},i}\,t_{\mathrm{L},i}\). Invalid points have zero support.

\subsection{Network architecture}
\label{sec:network-architecture}

\begin{figure}[!htb]
    \centering
    \includegraphics[width=\columnwidth]{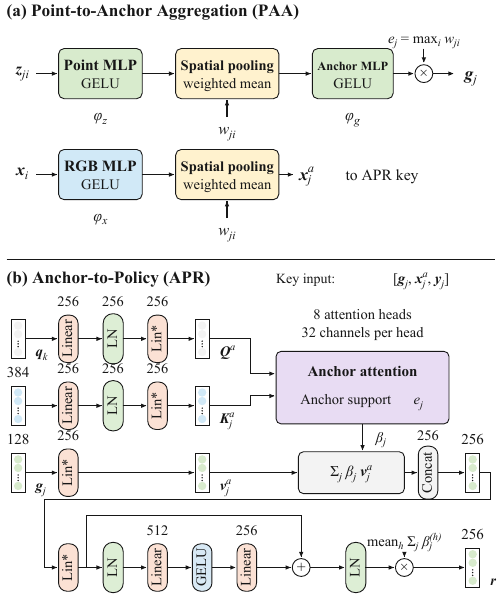}
    \caption{Geometric aggregation and readout networks. (a) PAA pools point geometry and RGB features using spatial weights; support \(e_j\) scales the geometric output. (b) APR uses appearance-guided anchor attention and gates the readout by the mean attention mass. Numbers denote channel widths; LN denotes LayerNorm; Lin* is bias-free Linear.}
    \label{fig:attention-architecture}
\end{figure}

Figures~\ref{fig:depth-decoder} and~\ref{fig:attention-architecture} specify the network layers and channel widths. MLPs use GELU; depth attention uses pre-LN and zero dropout. APR concatenates \([\mathbf g_j,\mathbf x_j^a,\mathbf y_j]\in\mathbb R^{384}\) as its key input. The constant \(1\) in the denominator of \cref{eq:regeo-policy} allows the total attention on observed anchors to be less than one.

\subsection{Supervision reduction}
\label{sec:cross-view-supervision-details}

For valid depth labels, we use
\begin{equation}
 \ell_d^\nu=
 \frac{\sum_{u\in\mathcal V^\nu} a_u^\nu\,
       \operatorname{SmoothL1}(d_u^\nu)}
      {\sum_{u\in\mathcal V^\nu} a_u^\nu}.
 \label{eq:supp-depth-reduction}
\end{equation}
Here, \(\mathcal V^\nu\) contains valid sampled depth labels. For source points, \(a_u^s=5\) inside the workspace and \(a_u^s=1\) outside; retained target reprojections use \(a_u^t=1\). We average over valid observations and views. Target depth predictions use continuous projected coordinates.

\subsection{Adaptive action-loss weighting}
\label{sec:action-weighting}

For MetaWorld, we emphasize demonstration stages with higher historical source-view denoising error. We partition normalized demonstration progress into ten bins and a separate initial-step bin. For each training window, its valid-action noise MSE is divided by a running reference for the same diffusion step. We maintain an exponential moving average (EMA) of this normalized error as the difficulty \(\delta_g\) of bin \(g\).

Let \(n_g\) be the number of observed windows in bin \(g\), and let \(\bar\delta=\sum_g n_g\delta_g/\sum_g n_g\) be the observation-weighted mean difficulty. Using statistics from previous updates, the eligible-bin weight and the normalized action weight in \cref{eq:regeo-action-loss} are
\begin{equation}
\begin{aligned}
 \eta_g&=1-\mu+\mu\operatorname{clip}
 \left(\delta_g/\bar\delta,\,c^{-1},\,c\right),\\
 \omega_k&=\eta_{g(W)}\,
 \frac{\sum_{W'\in\mathcal B}|W'|}
      {\sum_{W'\in\mathcal B}|W'|\,\eta_{g(W')}} ,\quad k\in W.
\end{aligned}
 \label{eq:supp-bin-weights}
\end{equation}
Here, \(g(W)\) is the progress bin of window \(W\), \(|W|\) is its number of valid action steps, and \(\mathcal B\) is the effective batch. The sums in \cref{eq:regeo-action-loss} pool valid window--step pairs across \(\mathcal B\), and the normalization keeps their mean weight at one. We use \(\mu=0.5\), \(c=5\), and EMA decay \(0.95\). Bins use \(\eta_g=1\) during the first 200 updates or until they contain 16 observations. Only source-view errors update the statistics; paired source and synthesized views share the same window weights. ManiSkill3 uses unit weights. Table~\ref{tab:regeo-hyperparameters} summarizes these settings.

\section{Experimental Settings}
\label{sec:supp-experimental-settings}

\subsection{Hyperparameters and evaluation budgets}
\label{sec:hyperparameters}

Table~\ref{tab:regeo-hyperparameters} gives additional controller settings, benchmark-specific training parameters, and source camera poses; Table~\ref{tab:episode-budgets} lists the control-step limits used by all compared methods.

\begin{table*}[t]
    \centering
    \caption{Additional training settings and source camera poses. Training-view intervals include the source pose and synthesized interpolations. Camera positions and look-at points use simulator world coordinates; angles follow the convention in Sec.~\ref{sec:source-cameras}. Within ManiSkill3, Others denotes \emph{PickCube}, \emph{PushCube}, and \emph{PullCube}.}
    \label{tab:regeo-hyperparameters}
    \begingroup
    \small
    \setlength{\tabcolsep}{2.5pt}
    \renewcommand{\arraystretch}{1.08}
    \begin{tabular*}{0.96\textwidth}{@{\extracolsep{\fill}}
        >{\raggedright\arraybackslash}p{0.30\textwidth}
        >{\raggedright\arraybackslash}p{0.28\textwidth}
        >{\raggedright\arraybackslash}p{0.28\textwidth}@{}}
        \toprule
        \textbf{Setting} & MetaWorld & ManiSkill3 \\
        \midrule
        Observation / prediction horizons & \(2/16\) & \(2/16\) \\
        Execution horizon & 8 & 8 \\
        DDPM steps / schedule & 100 / squared cosine & 100 / squared cosine \\
        APR readout layers & 1 & 1 \\
        \midrule
        Action weighting & Adaptive & Unit weights \\
        Included action steps & Non-padding only & All entries \\
        Adaptive \((\rho,\mu,c)\) & \((0.95,0.5,5)\) & --- \\
        Warm-up / minimum observations & \(T_w=200,\ n_{\min}=16\) & --- \\
        Progress bins & 10 + initial-step bin & --- \\
        Training azimuth offsets & \([-30^\circ,0^\circ]\) & \([-20^\circ,0^\circ]\) \\
        Training elevation offsets & \([0^\circ,30^\circ]\) & \([-10^\circ,20^\circ]\) \\
        Training distance scale & --- & \([0.95,1.2]\) \\
        \bottomrule
    \end{tabular*}
    \par\vspace{6pt}
    \begin{tabular*}{0.96\textwidth}{@{\extracolsep{\fill}}lccrrr@{}}
        \toprule
        \textbf{Source camera / task group} & Position (m) & Look-at (m) & Azimuth & Elevation & Distance (m) \\
        \midrule
        MetaWorld (all eight tasks) & \((-1.3022,-0.0394,0.5130)\) & \((0,0.5,0)\) & \(-157.50^\circ\) & \(20.00^\circ\) & 1.500 \\
        ManiSkill3: Others & \((0.5768,0.2690,0.5000)\) & \((0,0,0.12)\) & \(25.00^\circ\) & \(30.84^\circ\) & 0.741 \\
        ManiSkill3: \emph{PokeCube} & \((0.7200,0.3358,0.6200)\) & \((0.03,0,0.14)\) & \(25.95^\circ\) & \(32.03^\circ\) & 0.905 \\
        \bottomrule
    \end{tabular*}
    \endgroup
\end{table*}

\begin{table}[!ht]
    \centering
    \caption{Task-specific control-step limits. Shared by all compared methods.}
    \label{tab:episode-budgets}
    \small
    \begin{tabular*}{0.95\columnwidth}{@{\extracolsep{\fill}}llr@{}}
        \toprule
        \textbf{Benchmark} & Task & Steps \\
        \midrule
        MetaWorld & All except \emph{Bin Picking} & 150 \\
        MetaWorld & \emph{Bin Picking} & 250 \\
        ManiSkill3 & \emph{PickCube} & 100 \\
        ManiSkill3 & \emph{PushCube} & 100 \\
        ManiSkill3 & \emph{PullCube} & 50 \\
        ManiSkill3 & \emph{PokeCube} & 200 \\
        \bottomrule
    \end{tabular*}
\end{table}

\subsection{Baseline comparison protocol}
\label{sec:baseline-protocol}
All methods use the same source demonstrations, task budgets, and success criteria; augmented baselines share the synthesized views and action labels with ReGeoAnch. DP3 and 3D-DA use the same depth decoder and diffusion controller as ReGeoAnch. RAVEN, ReGeoAnch, and DP\({}^{\dagger}\) are evaluated under matched initial states, camera calibrations, and episode limits. On ManiSkill3, RAVEN uses end-effector pose and gripper-state inputs, with additional goal and grasp-state inputs for \emph{PickCube}. It uses the measured gripper orientation on MetaWorld \emph{Pick Place}. We adopted these task-specific settings after observing better performance for RAVEN in our experiments.

\subsection{Source camera poses}
\label{sec:source-cameras}
\begin{figure*}[p]
    \centering
    \includegraphics[width=\textwidth]{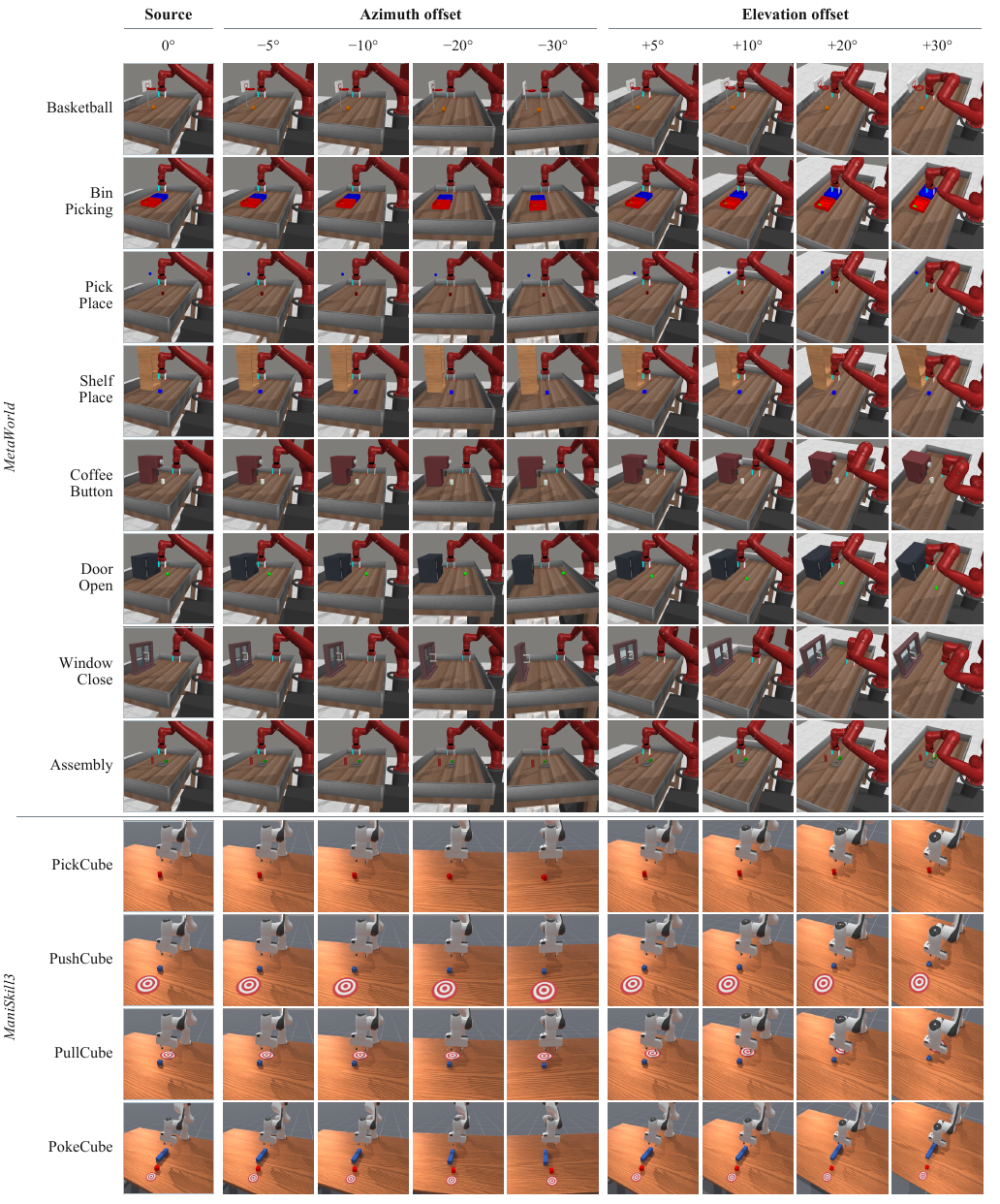}
    \caption{Task scenes under camera changes. Each row shows one frozen initial state at the source pose, four azimuth offsets, and four elevation offsets. Offsets are relative to the source camera; the other angle, look-at point, distance, and field of view remain fixed.}
    \label{fig:supp-task-camera-grid}
\end{figure*}

Table~\ref{tab:regeo-hyperparameters} lists source camera positions and look-at points in simulator world coordinates (meters). Azimuth is measured from \(+x\) toward \(+y\) along the vector from the look-at point to the camera, elevation is measured above the horizontal plane, and distance is the length of that vector. We use this world-coordinate convention for both benchmarks.
Figure~\ref{fig:supp-task-camera-grid} shows all twelve tasks under independent azimuth and elevation changes from these source poses.

\section{Additional Experiments}
\label{sec:additional-experiments}

Table~\ref{tab:source-target-summary} compares source and target means with equal weights for axes and tasks. Source results use each axis-specific policy at its source pose. ReGeoAnch has the smallest source-to-target drop on both benchmarks despite not having the highest source-view scores, supporting its ability to retain task performance as the camera moves.

\begin{table}[!htb]
    \centering
    \caption{Source- and target-view mean success (\%). Target matches Table~\ref{tab:main-results}.}
    \label{tab:source-target-summary}
    \begingroup
    \small
    \setlength{\tabcolsep}{1pt}
    \renewcommand{\arraystretch}{1.0}
    \begin{tabular*}{0.95\columnwidth}{@{\extracolsep{\fill}}lrrrr@{}}
    \toprule
    \textbf{Method} & \multicolumn{2}{c}{MetaWorld} & \multicolumn{2}{c}{ManiSkill3} \\
    \cmidrule(lr){2-3}\cmidrule(l){4-5}
    & Source & Target & Source & Target \\
    \midrule
    DP & 95.7 & 30.3 & 94.3 & 52.7 \\
    DP\({}^{\dagger}\) & \underline{96.3} & 62.5 & 96.8 & 86.5 \\
    DP3 & 90.0 & 71.8 & 92.8 & 72.9 \\
    3D-DA & \textbf{96.6} & 72.6 & 96.3 & 86.8 \\
    ReViWo & 87.3 & 74.2 & 81.8 & 43.9 \\
    CLASS & 95.7 & \underline{74.9} & 94.4 & 77.6 \\
    Cam-DP & \textbf{96.6} & 72.2 & 96.9 & 86.0 \\
    RAVEN & \textbf{96.6} & 62.6 & \textbf{99.8} & \underline{91.9} \\
    \midrule
    \textbf{ReGeoAnch} & 94.6 & \textbf{86.6} & \underline{97.8} & \textbf{95.6} \\
    \bottomrule
    \end{tabular*}
    \endgroup
\end{table}

\subsection{Viewpoint extrapolation}
\label{sec:supp-viewpoint-extrapolation}

We evaluate unchanged ManiSkill3 policies at \(+30^\circ\) elevation and \(-30^\circ\) azimuth, beyond the respective synthesized training ranges \([-10^\circ,+20^\circ]\) and \([-20^\circ,0^\circ]\). Table~\ref{tab:maniskill-extrapolation-axes} shows that ReGeoAnch achieves the highest mean success under both extrapolation settings. These frozen-policy results support transfer beyond the synthesized training range, with greater sensitivity to elevation than azimuth changes.

\begin{table*}[t]
    \centering
    \caption{Per-task viewpoint extrapolation on ManiSkill3. Success rates (\%) use 100 rollouts per task and view; Avg. weights the four tasks equally. Bold and underlined scores mark first and second place in each column, including ties.}
    \label{tab:maniskill-extrapolation-axes}
    \begingroup
    \small
    \setlength{\tabcolsep}{0.5pt}
    \renewcommand{\arraystretch}{1.03}
    \begin{tabular*}{0.94\textwidth}{@{\extracolsep{\fill}}p{66pt}*{10}{r}@{}}
    \toprule
    & \multicolumn{5}{c}{Elevation $+30^\circ$} & \multicolumn{5}{c}{Azimuth $-30^\circ$} \\
    \cmidrule(lr){2-6}\cmidrule(l){7-11}
    \textbf{Method} & \makebox[24pt][c]{\textit{\shortstack{Pick\\Cube}}} & \makebox[24pt][c]{\textit{\shortstack{Push\\Cube}}} & \makebox[24pt][c]{\textit{\shortstack{Pull\\Cube}}} & \makebox[24pt][c]{\textit{\shortstack{Poke\\Cube}}} & \makebox[24pt][c]{\textbf{Avg.}} & \makebox[24pt][c]{\textit{\shortstack{Pick\\Cube}}} & \makebox[24pt][c]{\textit{\shortstack{Push\\Cube}}} & \makebox[24pt][c]{\textit{\shortstack{Pull\\Cube}}} & \makebox[24pt][c]{\textit{\shortstack{Poke\\Cube}}} & \makebox[24pt][c]{\textbf{Avg.}} \\
    \midrule
    DP & \makebox[24pt][r]{0.0} & \makebox[24pt][r]{15.0} & \makebox[24pt][r]{0.0} & \makebox[24pt][r]{0.0} & \makebox[24pt][r]{3.8} & \makebox[24pt][r]{14.0} & \makebox[24pt][r]{15.0} & \makebox[24pt][r]{8.0} & \makebox[24pt][r]{1.0} & \makebox[24pt][r]{9.5} \\
    DP\({}^{\dagger}\) & \makebox[24pt][r]{0.0} & \makebox[24pt][r]{\textbf{41.0}} & \makebox[24pt][r]{19.0} & \makebox[24pt][r]{3.0} & \makebox[24pt][r]{15.8} & \makebox[24pt][r]{39.0} & \makebox[24pt][r]{44.0} & \makebox[24pt][r]{55.0} & \makebox[24pt][r]{20.0} & \makebox[24pt][r]{39.5} \\
    DP3 & \makebox[24pt][r]{2.0} & \makebox[24pt][r]{27.0} & \makebox[24pt][r]{16.0} & \makebox[24pt][r]{0.0} & \makebox[24pt][r]{11.2} & \makebox[24pt][r]{14.0} & \makebox[24pt][r]{\underline{61.0}} & \makebox[24pt][r]{53.0} & \makebox[24pt][r]{12.0} & \makebox[24pt][r]{35.0} \\
    3D-DA & \makebox[24pt][r]{2.0} & \makebox[24pt][r]{21.0} & \makebox[24pt][r]{13.0} & \makebox[24pt][r]{0.0} & \makebox[24pt][r]{9.0} & \makebox[24pt][r]{\underline{52.0}} & \makebox[24pt][r]{13.0} & \makebox[24pt][r]{49.0} & \makebox[24pt][r]{12.0} & \makebox[24pt][r]{31.5} \\
    ReViWo & \makebox[24pt][r]{0.0} & \makebox[24pt][r]{4.0} & \makebox[24pt][r]{0.0} & \makebox[24pt][r]{1.0} & \makebox[24pt][r]{1.2} & \makebox[24pt][r]{1.0} & \makebox[24pt][r]{3.0} & \makebox[24pt][r]{0.0} & \makebox[24pt][r]{0.0} & \makebox[24pt][r]{1.0} \\
    CLASS & \makebox[24pt][r]{0.0} & \makebox[24pt][r]{7.0} & \makebox[24pt][r]{1.0} & \makebox[24pt][r]{0.0} & \makebox[24pt][r]{2.0} & \makebox[24pt][r]{17.0} & \makebox[24pt][r]{42.0} & \makebox[24pt][r]{\underline{63.0}} & \makebox[24pt][r]{0.0} & \makebox[24pt][r]{30.5} \\
    Cam-DP & \makebox[24pt][r]{1.0} & \makebox[24pt][r]{\underline{40.0}} & \makebox[24pt][r]{18.0} & \makebox[24pt][r]{0.0} & \makebox[24pt][r]{14.8} & \makebox[24pt][r]{39.0} & \makebox[24pt][r]{40.0} & \makebox[24pt][r]{31.0} & \makebox[24pt][r]{2.0} & \makebox[24pt][r]{28.0} \\
    RAVEN & \makebox[24pt][r]{\underline{24.0}} & \makebox[24pt][r]{0.0} & \makebox[24pt][r]{\underline{50.0}} & \makebox[24pt][r]{\underline{37.0}} & \makebox[24pt][r]{\underline{27.8}} & \makebox[24pt][r]{31.0} & \makebox[24pt][r]{35.0} & \makebox[24pt][r]{57.0} & \makebox[24pt][r]{\underline{52.0}} & \makebox[24pt][r]{\underline{43.8}} \\
    \midrule
    \textbf{ReGeoAnch} & \makebox[24pt][r]{\textbf{38.0}} & \makebox[24pt][r]{33.0} & \makebox[24pt][r]{\textbf{96.0}} & \makebox[24pt][r]{\textbf{60.0}} & \makebox[24pt][r]{\textbf{56.8}} & \makebox[24pt][r]{\textbf{94.0}} & \makebox[24pt][r]{\textbf{99.0}} & \makebox[24pt][r]{\textbf{99.0}} & \makebox[24pt][r]{\textbf{56.0}} & \makebox[24pt][r]{\textbf{87.0}} \\
    \bottomrule
    \end{tabular*}
    \endgroup
\end{table*}

\subsection{Action-distribution stability}
\label{sec:supp-action-distribution}

\begin{figure*}[t]
    \centering
    \includegraphics[width=\textwidth]{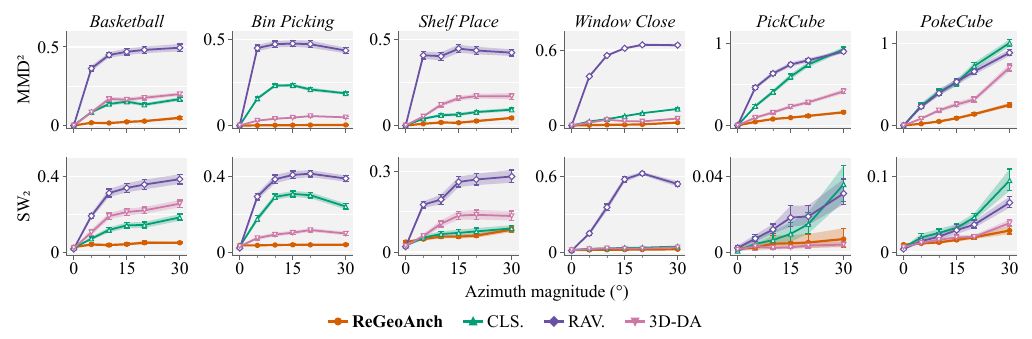}
    \caption{Action-distribution stability across all evaluated tasks and camera angles. Shading and error bars show 95\% confidence intervals.}
    \label{fig:supp-action-distribution-curves}
\end{figure*}\par

Figure~\ref{fig:supp-action-distribution-curves} reports angular trends for all six evaluated tasks, including the four shown in main-paper Figure~\ref{fig:action-distribution}. For each method and task, we average the five non-source-view distances within each episode and then average over all 100 episodes. We compute 95\% confidence intervals from 10,000 episode-bootstrap resamples, sharing episode indices across methods and views within each task. Dots show view-specific means, with darker shades representing larger azimuth magnitudes. Metric kernels and native command coordinates are unchanged, and distances are not pooled across tasks. All methods use matched physical states and the same metric settings within each task. ReGeoAnch has lower target-view mean MMD$^2$ than 3D-DA on all six tasks, and lower SW$_2$ on five tasks. On \emph{PickCube}, mean SW$_2$ is 0.00472 for ReGeoAnch and 0.00309 for 3D-DA; the paired difference has a 95\% interval of $[-0.00027,0.00470]$, so this comparison is inconclusive. The smaller cross-view shifts complement the success results by indicating more stable action generation at matched physical states.

\subsection{Success under control-step budgets}
\begin{figure}[t]
    \centering
    \includegraphics[width=\columnwidth]{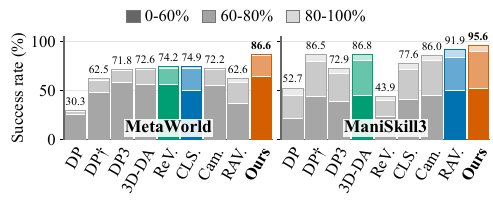}
    \caption{Success under control-step budgets. Segment boundaries give cumulative success at 60\%, 80\%, and 100\% budgets.}
    \label{fig:supp-completion-budget-bars}
\end{figure}

Figure~\ref{fig:supp-completion-budget-bars} compares cumulative success at 60\%, 80\%, and 100\% of each task's step limit using the same rollouts. ReGeoAnch leads both benchmarks at every budget, showing that its advantage persists when fewer control steps are available.

\par

\subsection{Qualitative policy comparison}
\label{sec:supp-qualitative-rollouts}
\begin{figure*}[p]
    \centering
    \includegraphics[width=\textwidth]{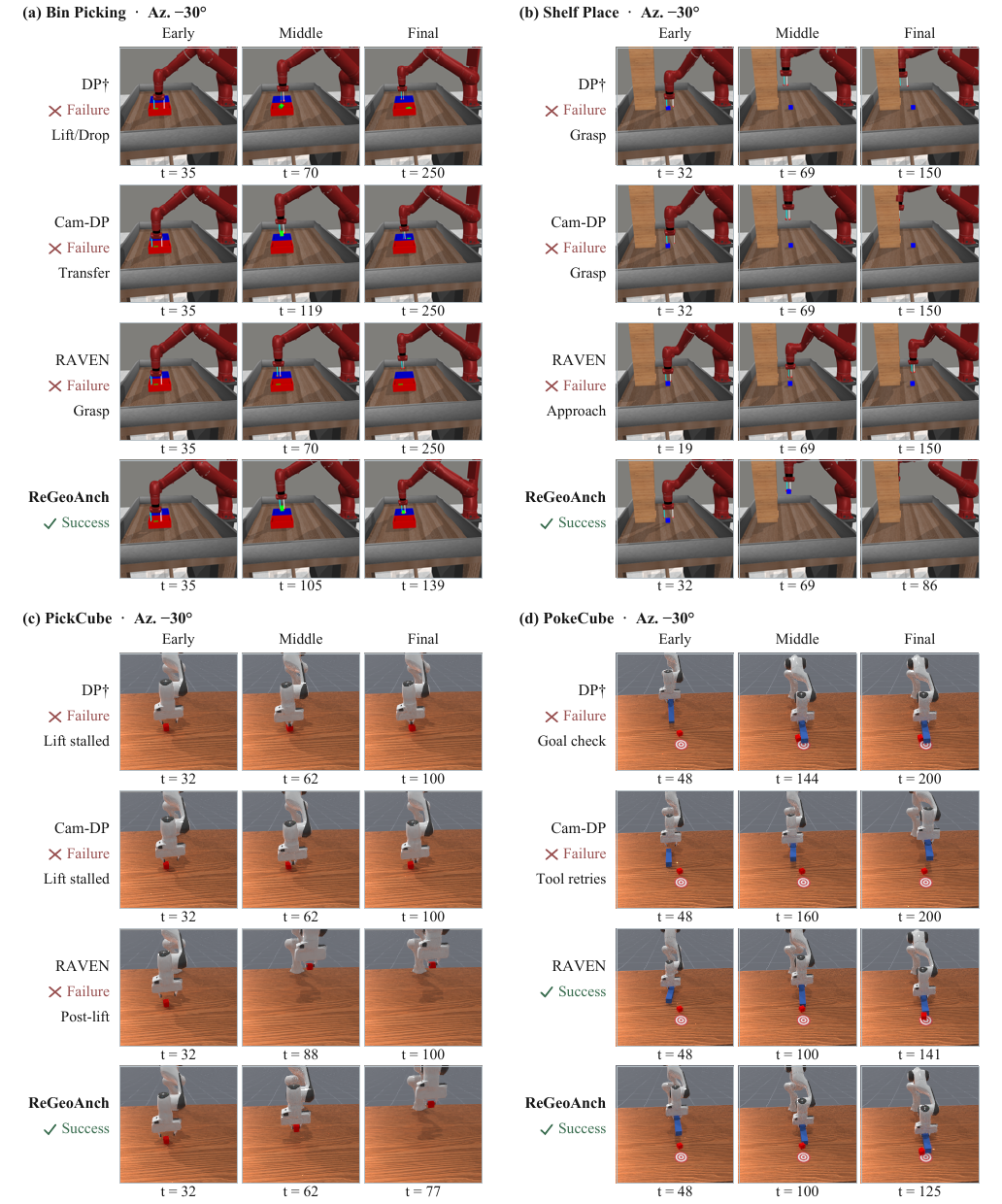}
    \caption{Failure stages in recorded rollouts. Methods share initial states at a \(-30^\circ\) azimuth offset. Labels locate failure stages; \(t\) denotes actual steps. PickCube RAVEN lifts the cube, and PokeCube DP\({}^{\dagger}\) moves it near the target, yet neither satisfies the full check requiring goal proximity and robot settling. Keyframes follow events; the final column shows termination.}
    \label{fig:supp-failure-comparison}
\end{figure*}

Figure~\ref{fig:supp-failure-comparison} locates failures during approach, grasping, lifting, transfer, and final task completion. ReGeoAnch completes all four selected cases. Failures among the compared policies occur at different interaction stages, showing that a successful approach or lift alone does not ensure task completion.

\subsection{Depth prediction and view synthesis}
\label{sec:supp-perception-comparison}
\begin{figure*}[p]
    \centering
    \includegraphics[width=\textwidth]{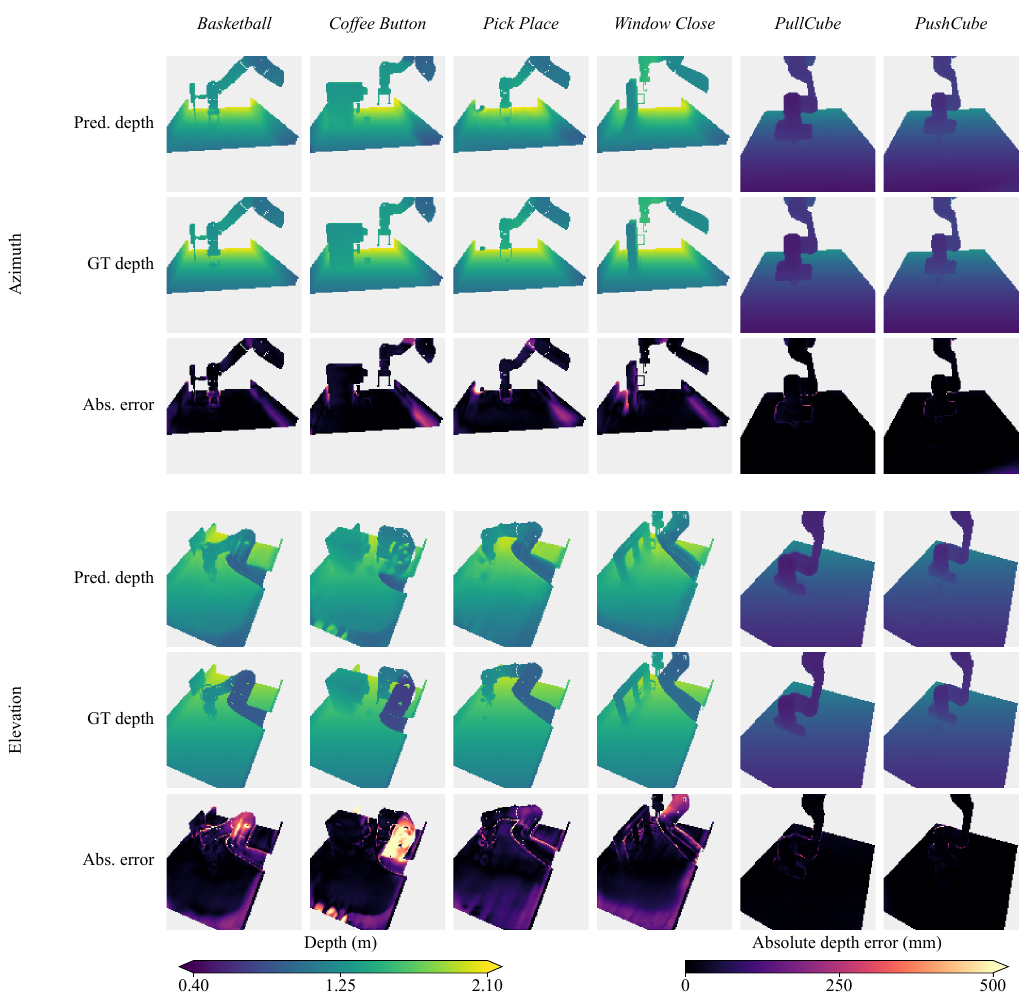}
    \caption{Depth predictions, GT depth and absolute errors across six tasks. Azimuth is \(-30^\circ\), while elevation is \(+30^\circ\) on MetaWorld and \(+20^\circ\) on ManiSkill3. Gray pixels lie outside the GT-defined workspace.}
    \label{fig:supp-depth-prediction}
\end{figure*}

\begin{figure*}[p]
    \centering
    \includegraphics[width=\textwidth]{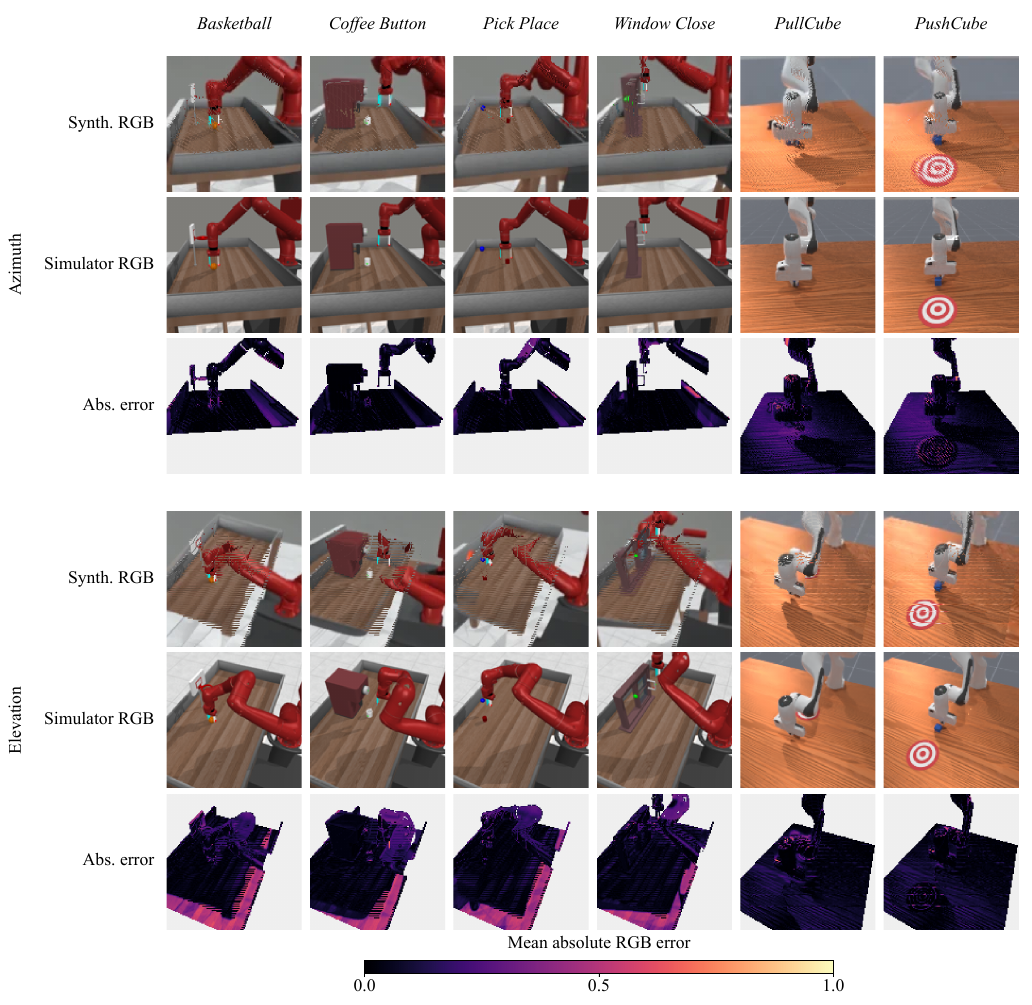}
    \caption{View synthesis for the same conditions as Figure~\ref{fig:supp-depth-prediction}. Rows compare synthesized RGB, simulator RGB and mean absolute channel errors. Error maps use the GT-defined workspace while RGB panels retain the full view.}
    \label{fig:supp-view-synthesis}
\end{figure*}

Figures~\ref{fig:supp-depth-prediction} and~\ref{fig:supp-view-synthesis} extend main-paper Figure~\ref{fig:perception-fidelity} to six tasks and two camera changes. Predictions and synthesized RGB images are compared with simulator ground truth at matched physical states. Predicted depth remains imperfect, and synthesized RGB differs from simulator renderings, yet ReGeoAnch maintains viewpoint generalization. The main-paper ablations support the role of its decoupled readout and paired learning: appearance guides the selection of task-relevant geometric values, while cross-view consistency stabilizes the resulting readouts.

\section{Detailed Results by Viewpoint}
\label{sec:supp-detailed-view-results}

Tables~\ref{tab:supp-mw-angles-1}--\ref{tab:supp-ms-angles} and~\ref{tab:supp-ms-distance} expand Table~\ref{tab:main-results} over angular offsets and camera distances, respectively. Entries are success rates (\%) over 100 rollouts. Each source column uses the policy evaluated on that axis. Axis averages exclude \(0^\circ\) or distance scale \(1.0\); Task Avg. reproduces Table~\ref{tab:main-results} by equally weighting the relevant axis averages, including distance for ManiSkill3. Averages are computed before rounding. ReGeoAnch's largest average gains occur under azimuth changes on both benchmarks. Under distance scaling, several methods remain strong and RAVEN achieves the highest mean.

\section{Deployment Cost}
\label{sec:supp-deployment-cost}

Table~\ref{tab:efficiency} compares parameter counts and inference times. ReGeoAnch adds \(14.7\,\mathrm{ms}\) per action chunk relative to DP, quantifying the deployment overhead of its additional geometric processing and readout.

\begingroup
\setlength{\intextsep}{6pt}
\begin{table}[!htb]
    \centering
    \caption{Parameter counts and mean latency per 16-step action chunk on an RTX 5090 (FP32, batch size 1). Native samplers are used; simulation and I/O are excluded.}
    \label{tab:efficiency}
    \small
    \setlength{\tabcolsep}{1pt}
    \renewcommand{\arraystretch}{1.02}
    \begin{minipage}[t]{0.48\columnwidth}
        \vspace{0pt}
        \centering
        \begin{tabular*}{\linewidth}{@{\extracolsep{\fill}}p{47pt}rr@{}}
            \toprule
            \textbf{Method} & \makebox[27pt][c]{\shortstack{Params.\\(M)}} & \makebox[27pt][c]{\shortstack{Time\\(ms)}} \\
            \midrule
            DP & \makebox[27pt][r]{17.4} & \makebox[27pt][r]{79.6} \\
            DP\({}^{\dagger}\) & \makebox[27pt][r]{17.4} & \makebox[27pt][r]{79.9} \\
            DP3 & \makebox[27pt][r]{19.2} & \makebox[27pt][r]{84.1} \\
            3D-DA & \makebox[27pt][r]{20.2} & \makebox[27pt][r]{85.2} \\
            & & \\
            \bottomrule
        \end{tabular*}
    \end{minipage}\hfill
    \begin{minipage}[t]{0.48\columnwidth}
        \vspace{0pt}
        \centering
        \begin{tabular*}{\linewidth}{@{\extracolsep{\fill}}p{47pt}rr@{}}
            \toprule
            \textbf{Method} & \makebox[27pt][c]{\shortstack{Params.\\(M)}} & \makebox[27pt][c]{\shortstack{Time\\(ms)}} \\
            \midrule
            ReViWo & \makebox[27pt][r]{11.8} & \makebox[27pt][r]{81.4} \\
            CLASS & \makebox[27pt][r]{22.8} & \makebox[27pt][r]{82.9} \\
            Cam-DP & \makebox[27pt][r]{16.0} & \makebox[27pt][r]{78.5} \\
            RAVEN & \makebox[27pt][r]{20.5} & \makebox[27pt][r]{74.8} \\
            \textbf{ReGeoAnch} & \makebox[27pt][r]{21.2} & \makebox[27pt][r]{94.3} \\
            \bottomrule
        \end{tabular*}
    \end{minipage}
\end{table}
\endgroup

\clearpage

\begin{table*}[p]
    \centering
    \caption{Per-view success rates (\%) on MetaWorld: tasks 1--4. Each condition uses 100 rollouts. Axis averages exclude source poses; Task Avg. equals the task score in Table~\ref{tab:main-results}. Bold and underlining mark first and second place, including ties.}
    \label{tab:supp-mw-angles-1}
    \begingroup
    \fontsize{9}{10.5}\selectfont
    \setlength{\tabcolsep}{0.5pt}
    \setlength{\aboverulesep}{1.5pt}
    \setlength{\belowrulesep}{1.5pt}
    \renewcommand{\arraystretch}{1.0}
    \begin{tabular*}{0.98\textwidth}{@{\extracolsep{\fill}}p{56pt}*{16}{r}@{}}
    \toprule
    & \multicolumn{7}{c}{Azimuth} & \multicolumn{8}{c}{Elevation} & \makebox[24pt][c]{Task} \\
    \cmidrule(lr){2-8}\cmidrule(lr){9-16}
    \textbf{Method} & \makebox[24pt][c]{$0^\circ$} & \makebox[24pt][c]{$-5^\circ$} & \makebox[24pt][c]{$-10^\circ$} & \makebox[24pt][c]{$-15^\circ$} & \makebox[24pt][c]{$-20^\circ$} & \makebox[24pt][c]{$-30^\circ$} & \makebox[24pt][c]{\textbf{Avg.}} & \makebox[24pt][c]{$0^\circ$} & \makebox[24pt][c]{$+5^\circ$} & \makebox[24pt][c]{$+10^\circ$} & \makebox[24pt][c]{$+15^\circ$} & \makebox[24pt][c]{$+20^\circ$} & \makebox[24pt][c]{$+25^\circ$} & \makebox[24pt][c]{$+30^\circ$} & \makebox[24pt][c]{\textbf{Avg.}} & \makebox[24pt][c]{\textbf{Avg.}} \\
    \midrule
    \multicolumn{17}{@{}l}{\textbf{(a) Basketball}} \\
    DP & \makebox[24pt][r]{86.0} & \makebox[24pt][r]{26.0} & \makebox[24pt][r]{8.0} & \makebox[24pt][r]{1.0} & \makebox[24pt][r]{0.0} & \makebox[24pt][r]{0.0} & \makebox[24pt][r]{7.0} & \makebox[24pt][r]{86.0} & \makebox[24pt][r]{0.0} & \makebox[24pt][r]{0.0} & \makebox[24pt][r]{0.0} & \makebox[24pt][r]{0.0} & \makebox[24pt][r]{0.0} & \makebox[24pt][r]{0.0} & \makebox[24pt][r]{0.0} & \makebox[24pt][r]{3.5} \\
    DP\({}^{\dagger}\) & \makebox[24pt][r]{\underline{89.0}} & \makebox[24pt][r]{\underline{71.0}} & \makebox[24pt][r]{\underline{51.0}} & \makebox[24pt][r]{23.0} & \makebox[24pt][r]{17.0} & \makebox[24pt][r]{14.0} & \makebox[24pt][r]{35.2} & \makebox[24pt][r]{84.0} & \makebox[24pt][r]{16.0} & \makebox[24pt][r]{17.0} & \makebox[24pt][r]{21.0} & \makebox[24pt][r]{13.0} & \makebox[24pt][r]{5.0} & \makebox[24pt][r]{3.0} & \makebox[24pt][r]{12.5} & \makebox[24pt][r]{23.9} \\
    DP3 & \makebox[24pt][r]{77.0} & \makebox[24pt][r]{42.0} & \makebox[24pt][r]{30.0} & \makebox[24pt][r]{33.0} & \makebox[24pt][r]{39.0} & \makebox[24pt][r]{8.0} & \makebox[24pt][r]{30.4} & \makebox[24pt][r]{41.0} & \makebox[24pt][r]{33.0} & \makebox[24pt][r]{24.0} & \makebox[24pt][r]{20.0} & \makebox[24pt][r]{9.0} & \makebox[24pt][r]{8.0} & \makebox[24pt][r]{5.0} & \makebox[24pt][r]{16.5} & \makebox[24pt][r]{23.5} \\
    3D-DA & \makebox[24pt][r]{\textbf{91.0}} & \makebox[24pt][r]{39.0} & \makebox[24pt][r]{13.0} & \makebox[24pt][r]{12.0} & \makebox[24pt][r]{6.0} & \makebox[24pt][r]{3.0} & \makebox[24pt][r]{14.6} & \makebox[24pt][r]{86.0} & \makebox[24pt][r]{50.0} & \makebox[24pt][r]{38.0} & \makebox[24pt][r]{27.0} & \makebox[24pt][r]{20.0} & \makebox[24pt][r]{15.0} & \makebox[24pt][r]{0.0} & \makebox[24pt][r]{25.0} & \makebox[24pt][r]{19.8} \\
    ReViWo & \makebox[24pt][r]{71.0} & \makebox[24pt][r]{61.0} & \makebox[24pt][r]{44.0} & \makebox[24pt][r]{\underline{40.0}} & \makebox[24pt][r]{\underline{47.0}} & \makebox[24pt][r]{\underline{30.0}} & \makebox[24pt][r]{\underline{44.4}} & \makebox[24pt][r]{68.0} & \makebox[24pt][r]{64.0} & \makebox[24pt][r]{53.0} & \makebox[24pt][r]{39.0} & \makebox[24pt][r]{\underline{38.0}} & \makebox[24pt][r]{\underline{28.0}} & \makebox[24pt][r]{\underline{21.0}} & \makebox[24pt][r]{\underline{40.5}} & \makebox[24pt][r]{\underline{42.5}} \\
    CLASS & \makebox[24pt][r]{87.0} & \makebox[24pt][r]{44.0} & \makebox[24pt][r]{11.0} & \makebox[24pt][r]{12.0} & \makebox[24pt][r]{11.0} & \makebox[24pt][r]{20.0} & \makebox[24pt][r]{19.6} & \makebox[24pt][r]{\underline{87.0}} & \makebox[24pt][r]{\textbf{85.0}} & \makebox[24pt][r]{\underline{58.0}} & \makebox[24pt][r]{40.0} & \makebox[24pt][r]{29.0} & \makebox[24pt][r]{14.0} & \makebox[24pt][r]{5.0} & \makebox[24pt][r]{38.5} & \makebox[24pt][r]{29.1} \\
    Cam-DP & \makebox[24pt][r]{87.0} & \makebox[24pt][r]{33.0} & \makebox[24pt][r]{15.0} & \makebox[24pt][r]{19.0} & \makebox[24pt][r]{12.0} & \makebox[24pt][r]{4.0} & \makebox[24pt][r]{16.6} & \makebox[24pt][r]{\underline{87.0}} & \makebox[24pt][r]{58.0} & \makebox[24pt][r]{46.0} & \makebox[24pt][r]{\underline{42.0}} & \makebox[24pt][r]{14.0} & \makebox[24pt][r]{2.0} & \makebox[24pt][r]{0.0} & \makebox[24pt][r]{27.0} & \makebox[24pt][r]{21.8} \\
    RAVEN & \makebox[24pt][r]{87.0} & \makebox[24pt][r]{33.0} & \makebox[24pt][r]{4.0} & \makebox[24pt][r]{2.0} & \makebox[24pt][r]{19.0} & \makebox[24pt][r]{1.0} & \makebox[24pt][r]{11.8} & \makebox[24pt][r]{\textbf{94.0}} & \makebox[24pt][r]{76.0} & \makebox[24pt][r]{17.0} & \makebox[24pt][r]{34.0} & \makebox[24pt][r]{31.0} & \makebox[24pt][r]{20.0} & \makebox[24pt][r]{11.0} & \makebox[24pt][r]{31.5} & \makebox[24pt][r]{21.7} \\
    \cmidrule(lr){1-17}
    \textbf{ReGeoAnch} & \makebox[24pt][r]{76.0} & \makebox[24pt][r]{\textbf{85.0}} & \makebox[24pt][r]{\textbf{78.0}} & \makebox[24pt][r]{\textbf{81.0}} & \makebox[24pt][r]{\textbf{79.0}} & \makebox[24pt][r]{\textbf{72.0}} & \makebox[24pt][r]{\textbf{79.0}} & \makebox[24pt][r]{82.0} & \makebox[24pt][r]{\underline{79.0}} & \makebox[24pt][r]{\textbf{69.0}} & \makebox[24pt][r]{\textbf{61.0}} & \makebox[24pt][r]{\textbf{63.0}} & \makebox[24pt][r]{\textbf{45.0}} & \makebox[24pt][r]{\textbf{45.0}} & \makebox[24pt][r]{\textbf{60.3}} & \makebox[24pt][r]{\textbf{69.7}} \\
    \midrule
    \multicolumn{17}{@{}l}{\textbf{(b) Bin Picking}} \\
    DP & \makebox[24pt][r]{\underline{99.0}} & \makebox[24pt][r]{15.0} & \makebox[24pt][r]{0.0} & \makebox[24pt][r]{1.0} & \makebox[24pt][r]{3.0} & \makebox[24pt][r]{0.0} & \makebox[24pt][r]{3.8} & \makebox[24pt][r]{\underline{99.0}} & \makebox[24pt][r]{42.0} & \makebox[24pt][r]{11.0} & \makebox[24pt][r]{0.0} & \makebox[24pt][r]{0.0} & \makebox[24pt][r]{0.0} & \makebox[24pt][r]{1.0} & \makebox[24pt][r]{9.0} & \makebox[24pt][r]{6.4} \\
    DP\({}^{\dagger}\) & \makebox[24pt][r]{\underline{99.0}} & \makebox[24pt][r]{82.0} & \makebox[24pt][r]{67.0} & \makebox[24pt][r]{82.0} & \makebox[24pt][r]{64.0} & \makebox[24pt][r]{68.0} & \makebox[24pt][r]{72.6} & \makebox[24pt][r]{\textbf{100.0}} & \makebox[24pt][r]{91.0} & \makebox[24pt][r]{69.0} & \makebox[24pt][r]{40.0} & \makebox[24pt][r]{45.0} & \makebox[24pt][r]{35.0} & \makebox[24pt][r]{32.0} & \makebox[24pt][r]{52.0} & \makebox[24pt][r]{62.3} \\
    DP3 & \makebox[24pt][r]{98.0} & \makebox[24pt][r]{\underline{95.0}} & \makebox[24pt][r]{\underline{95.0}} & \makebox[24pt][r]{\underline{95.0}} & \makebox[24pt][r]{\underline{94.0}} & \makebox[24pt][r]{72.0} & \makebox[24pt][r]{\underline{90.2}} & \makebox[24pt][r]{98.0} & \makebox[24pt][r]{\textbf{97.0}} & \makebox[24pt][r]{\textbf{97.0}} & \makebox[24pt][r]{\textbf{94.0}} & \makebox[24pt][r]{\underline{85.0}} & \makebox[24pt][r]{60.0} & \makebox[24pt][r]{46.0} & \makebox[24pt][r]{79.8} & \makebox[24pt][r]{\underline{85.0}} \\
    3D-DA & \makebox[24pt][r]{98.0} & \makebox[24pt][r]{84.0} & \makebox[24pt][r]{86.0} & \makebox[24pt][r]{92.0} & \makebox[24pt][r]{\underline{94.0}} & \makebox[24pt][r]{\underline{79.0}} & \makebox[24pt][r]{87.0} & \makebox[24pt][r]{\textbf{100.0}} & \makebox[24pt][r]{\textbf{97.0}} & \makebox[24pt][r]{89.0} & \makebox[24pt][r]{83.0} & \makebox[24pt][r]{69.0} & \makebox[24pt][r]{63.0} & \makebox[24pt][r]{66.0} & \makebox[24pt][r]{77.8} & \makebox[24pt][r]{82.4} \\
    ReViWo & \makebox[24pt][r]{98.0} & \makebox[24pt][r]{91.0} & \makebox[24pt][r]{91.0} & \makebox[24pt][r]{87.0} & \makebox[24pt][r]{85.0} & \makebox[24pt][r]{71.0} & \makebox[24pt][r]{85.0} & \makebox[24pt][r]{97.0} & \makebox[24pt][r]{\underline{93.0}} & \makebox[24pt][r]{82.0} & \makebox[24pt][r]{\underline{84.0}} & \makebox[24pt][r]{\textbf{89.0}} & \makebox[24pt][r]{\underline{70.0}} & \makebox[24pt][r]{69.0} & \makebox[24pt][r]{\underline{81.2}} & \makebox[24pt][r]{83.1} \\
    CLASS & \makebox[24pt][r]{\textbf{100.0}} & \makebox[24pt][r]{73.0} & \makebox[24pt][r]{29.0} & \makebox[24pt][r]{33.0} & \makebox[24pt][r]{9.0} & \makebox[24pt][r]{27.0} & \makebox[24pt][r]{34.2} & \makebox[24pt][r]{98.0} & \makebox[24pt][r]{77.0} & \makebox[24pt][r]{56.0} & \makebox[24pt][r]{49.0} & \makebox[24pt][r]{37.0} & \makebox[24pt][r]{65.0} & \makebox[24pt][r]{\underline{70.0}} & \makebox[24pt][r]{59.0} & \makebox[24pt][r]{46.6} \\
    Cam-DP & \makebox[24pt][r]{\underline{99.0}} & \makebox[24pt][r]{83.0} & \makebox[24pt][r]{90.0} & \makebox[24pt][r]{79.0} & \makebox[24pt][r]{70.0} & \makebox[24pt][r]{53.0} & \makebox[24pt][r]{75.0} & \makebox[24pt][r]{\underline{99.0}} & \makebox[24pt][r]{\underline{93.0}} & \makebox[24pt][r]{82.0} & \makebox[24pt][r]{80.0} & \makebox[24pt][r]{76.0} & \makebox[24pt][r]{58.0} & \makebox[24pt][r]{\textbf{73.0}} & \makebox[24pt][r]{77.0} & \makebox[24pt][r]{76.0} \\
    RAVEN & \makebox[24pt][r]{96.0} & \makebox[24pt][r]{37.0} & \makebox[24pt][r]{2.0} & \makebox[24pt][r]{0.0} & \makebox[24pt][r]{0.0} & \makebox[24pt][r]{1.0} & \makebox[24pt][r]{8.0} & \makebox[24pt][r]{95.0} & \makebox[24pt][r]{65.0} & \makebox[24pt][r]{44.0} & \makebox[24pt][r]{10.0} & \makebox[24pt][r]{0.0} & \makebox[24pt][r]{1.0} & \makebox[24pt][r]{0.0} & \makebox[24pt][r]{20.0} & \makebox[24pt][r]{14.0} \\
    \cmidrule(lr){1-17}
    \textbf{ReGeoAnch} & \makebox[24pt][r]{\underline{99.0}} & \makebox[24pt][r]{\textbf{98.0}} & \makebox[24pt][r]{\textbf{98.0}} & \makebox[24pt][r]{\textbf{100.0}} & \makebox[24pt][r]{\textbf{99.0}} & \makebox[24pt][r]{\textbf{93.0}} & \makebox[24pt][r]{\textbf{97.6}} & \makebox[24pt][r]{97.0} & \makebox[24pt][r]{\textbf{97.0}} & \makebox[24pt][r]{\underline{96.0}} & \makebox[24pt][r]{\textbf{94.0}} & \makebox[24pt][r]{83.0} & \makebox[24pt][r]{\textbf{72.0}} & \makebox[24pt][r]{67.0} & \makebox[24pt][r]{\textbf{84.8}} & \makebox[24pt][r]{\textbf{91.2}} \\
    \midrule
    \multicolumn{17}{@{}l}{\textbf{(c) Pick Place}} \\
    DP & \makebox[24pt][r]{93.0} & \makebox[24pt][r]{43.0} & \makebox[24pt][r]{14.0} & \makebox[24pt][r]{1.0} & \makebox[24pt][r]{0.0} & \makebox[24pt][r]{1.0} & \makebox[24pt][r]{11.8} & \makebox[24pt][r]{93.0} & \makebox[24pt][r]{13.0} & \makebox[24pt][r]{4.0} & \makebox[24pt][r]{3.0} & \makebox[24pt][r]{3.0} & \makebox[24pt][r]{0.0} & \makebox[24pt][r]{0.0} & \makebox[24pt][r]{3.8} & \makebox[24pt][r]{7.8} \\
    DP\({}^{\dagger}\) & \makebox[24pt][r]{94.0} & \makebox[24pt][r]{82.0} & \makebox[24pt][r]{53.0} & \makebox[24pt][r]{47.0} & \makebox[24pt][r]{32.0} & \makebox[24pt][r]{21.0} & \makebox[24pt][r]{47.0} & \makebox[24pt][r]{95.0} & \makebox[24pt][r]{80.0} & \makebox[24pt][r]{45.0} & \makebox[24pt][r]{19.0} & \makebox[24pt][r]{9.0} & \makebox[24pt][r]{17.0} & \makebox[24pt][r]{11.0} & \makebox[24pt][r]{30.2} & \makebox[24pt][r]{38.6} \\
    DP3 & \makebox[24pt][r]{90.0} & \makebox[24pt][r]{88.0} & \makebox[24pt][r]{79.0} & \makebox[24pt][r]{74.0} & \makebox[24pt][r]{58.0} & \makebox[24pt][r]{24.0} & \makebox[24pt][r]{64.6} & \makebox[24pt][r]{92.0} & \makebox[24pt][r]{86.0} & \makebox[24pt][r]{78.0} & \makebox[24pt][r]{63.0} & \makebox[24pt][r]{58.0} & \makebox[24pt][r]{34.0} & \makebox[24pt][r]{31.0} & \makebox[24pt][r]{58.3} & \makebox[24pt][r]{61.5} \\
    3D-DA & \makebox[24pt][r]{97.0} & \makebox[24pt][r]{86.0} & \makebox[24pt][r]{83.0} & \makebox[24pt][r]{77.0} & \makebox[24pt][r]{59.0} & \makebox[24pt][r]{35.0} & \makebox[24pt][r]{68.0} & \makebox[24pt][r]{96.0} & \makebox[24pt][r]{90.0} & \makebox[24pt][r]{72.0} & \makebox[24pt][r]{40.0} & \makebox[24pt][r]{33.0} & \makebox[24pt][r]{36.0} & \makebox[24pt][r]{23.0} & \makebox[24pt][r]{49.0} & \makebox[24pt][r]{58.5} \\
    ReViWo & \makebox[24pt][r]{69.0} & \makebox[24pt][r]{69.0} & \makebox[24pt][r]{67.0} & \makebox[24pt][r]{49.0} & \makebox[24pt][r]{42.0} & \makebox[24pt][r]{23.0} & \makebox[24pt][r]{50.0} & \makebox[24pt][r]{63.0} & \makebox[24pt][r]{60.0} & \makebox[24pt][r]{59.0} & \makebox[24pt][r]{49.0} & \makebox[24pt][r]{48.0} & \makebox[24pt][r]{35.0} & \makebox[24pt][r]{29.0} & \makebox[24pt][r]{46.7} & \makebox[24pt][r]{48.3} \\
    CLASS & \makebox[24pt][r]{\textbf{99.0}} & \makebox[24pt][r]{\underline{94.0}} & \makebox[24pt][r]{\underline{88.0}} & \makebox[24pt][r]{\underline{80.0}} & \makebox[24pt][r]{\textbf{79.0}} & \makebox[24pt][r]{45.0} & \makebox[24pt][r]{\underline{77.2}} & \makebox[24pt][r]{\underline{98.0}} & \makebox[24pt][r]{\underline{97.0}} & \makebox[24pt][r]{\underline{88.0}} & \makebox[24pt][r]{\underline{80.0}} & \makebox[24pt][r]{\underline{82.0}} & \makebox[24pt][r]{\underline{76.0}} & \makebox[24pt][r]{\underline{73.0}} & \makebox[24pt][r]{\underline{82.7}} & \makebox[24pt][r]{\underline{79.9}} \\
    Cam-DP & \makebox[24pt][r]{\underline{98.0}} & \makebox[24pt][r]{86.0} & \makebox[24pt][r]{85.0} & \makebox[24pt][r]{78.0} & \makebox[24pt][r]{62.0} & \makebox[24pt][r]{37.0} & \makebox[24pt][r]{69.6} & \makebox[24pt][r]{\underline{98.0}} & \makebox[24pt][r]{83.0} & \makebox[24pt][r]{66.0} & \makebox[24pt][r]{55.0} & \makebox[24pt][r]{42.0} & \makebox[24pt][r]{20.0} & \makebox[24pt][r]{4.0} & \makebox[24pt][r]{45.0} & \makebox[24pt][r]{57.3} \\
    RAVEN & \makebox[24pt][r]{\underline{98.0}} & \makebox[24pt][r]{\textbf{95.0}} & \makebox[24pt][r]{66.0} & \makebox[24pt][r]{46.0} & \makebox[24pt][r]{45.0} & \makebox[24pt][r]{\textbf{79.0}} & \makebox[24pt][r]{66.2} & \makebox[24pt][r]{\textbf{100.0}} & \makebox[24pt][r]{\textbf{98.0}} & \makebox[24pt][r]{\textbf{98.0}} & \makebox[24pt][r]{\textbf{98.0}} & \makebox[24pt][r]{\textbf{96.0}} & \makebox[24pt][r]{\textbf{92.0}} & \makebox[24pt][r]{\textbf{87.0}} & \makebox[24pt][r]{\textbf{94.8}} & \makebox[24pt][r]{\textbf{80.5}} \\
    \cmidrule(lr){1-17}
    \textbf{ReGeoAnch} & \makebox[24pt][r]{94.0} & \makebox[24pt][r]{92.0} & \makebox[24pt][r]{\textbf{95.0}} & \makebox[24pt][r]{\textbf{87.0}} & \makebox[24pt][r]{\underline{76.0}} & \makebox[24pt][r]{\underline{57.0}} & \makebox[24pt][r]{\textbf{81.4}} & \makebox[24pt][r]{97.0} & \makebox[24pt][r]{95.0} & \makebox[24pt][r]{75.0} & \makebox[24pt][r]{66.0} & \makebox[24pt][r]{37.0} & \makebox[24pt][r]{31.0} & \makebox[24pt][r]{33.0} & \makebox[24pt][r]{56.2} & \makebox[24pt][r]{68.8} \\
    \midrule
    \multicolumn{17}{@{}l}{\textbf{(d) Shelf Place}} \\
    DP & \makebox[24pt][r]{92.0} & \makebox[24pt][r]{20.0} & \makebox[24pt][r]{0.0} & \makebox[24pt][r]{0.0} & \makebox[24pt][r]{0.0} & \makebox[24pt][r]{0.0} & \makebox[24pt][r]{4.0} & \makebox[24pt][r]{92.0} & \makebox[24pt][r]{4.0} & \makebox[24pt][r]{0.0} & \makebox[24pt][r]{0.0} & \makebox[24pt][r]{0.0} & \makebox[24pt][r]{0.0} & \makebox[24pt][r]{0.0} & \makebox[24pt][r]{0.7} & \makebox[24pt][r]{2.3} \\
    DP\({}^{\dagger}\) & \makebox[24pt][r]{\underline{94.0}} & \makebox[24pt][r]{53.0} & \makebox[24pt][r]{28.0} & \makebox[24pt][r]{17.0} & \makebox[24pt][r]{11.0} & \makebox[24pt][r]{25.0} & \makebox[24pt][r]{26.8} & \makebox[24pt][r]{\underline{93.0}} & \makebox[24pt][r]{29.0} & \makebox[24pt][r]{37.0} & \makebox[24pt][r]{29.0} & \makebox[24pt][r]{15.0} & \makebox[24pt][r]{12.0} & \makebox[24pt][r]{5.0} & \makebox[24pt][r]{21.2} & \makebox[24pt][r]{24.0} \\
    DP3 & \makebox[24pt][r]{82.0} & \makebox[24pt][r]{59.0} & \makebox[24pt][r]{55.0} & \makebox[24pt][r]{47.0} & \makebox[24pt][r]{35.0} & \makebox[24pt][r]{23.0} & \makebox[24pt][r]{43.8} & \makebox[24pt][r]{87.0} & \makebox[24pt][r]{73.0} & \makebox[24pt][r]{57.0} & \makebox[24pt][r]{45.0} & \makebox[24pt][r]{23.0} & \makebox[24pt][r]{26.0} & \makebox[24pt][r]{18.0} & \makebox[24pt][r]{40.3} & \makebox[24pt][r]{42.1} \\
    3D-DA & \makebox[24pt][r]{90.0} & \makebox[24pt][r]{70.0} & \makebox[24pt][r]{49.0} & \makebox[24pt][r]{39.0} & \makebox[24pt][r]{25.0} & \makebox[24pt][r]{24.0} & \makebox[24pt][r]{41.4} & \makebox[24pt][r]{\textbf{95.0}} & \makebox[24pt][r]{74.0} & \makebox[24pt][r]{66.0} & \makebox[24pt][r]{\underline{62.0}} & \makebox[24pt][r]{32.0} & \makebox[24pt][r]{25.0} & \makebox[24pt][r]{17.0} & \makebox[24pt][r]{46.0} & \makebox[24pt][r]{43.7} \\
    ReViWo & \makebox[24pt][r]{71.0} & \makebox[24pt][r]{61.0} & \makebox[24pt][r]{47.0} & \makebox[24pt][r]{52.0} & \makebox[24pt][r]{43.0} & \makebox[24pt][r]{39.0} & \makebox[24pt][r]{48.4} & \makebox[24pt][r]{78.0} & \makebox[24pt][r]{69.0} & \makebox[24pt][r]{53.0} & \makebox[24pt][r]{48.0} & \makebox[24pt][r]{40.0} & \makebox[24pt][r]{33.0} & \makebox[24pt][r]{27.0} & \makebox[24pt][r]{45.0} & \makebox[24pt][r]{46.7} \\
    CLASS & \makebox[24pt][r]{90.0} & \makebox[24pt][r]{\underline{81.0}} & \makebox[24pt][r]{\underline{82.0}} & \makebox[24pt][r]{\underline{69.0}} & \makebox[24pt][r]{\underline{62.0}} & \makebox[24pt][r]{\underline{40.0}} & \makebox[24pt][r]{\underline{66.8}} & \makebox[24pt][r]{85.0} & \makebox[24pt][r]{78.0} & \makebox[24pt][r]{\textbf{74.0}} & \makebox[24pt][r]{60.0} & \makebox[24pt][r]{41.0} & \makebox[24pt][r]{12.0} & \makebox[24pt][r]{2.0} & \makebox[24pt][r]{44.5} & \makebox[24pt][r]{\underline{55.7}} \\
    Cam-DP & \makebox[24pt][r]{\textbf{95.0}} & \makebox[24pt][r]{58.0} & \makebox[24pt][r]{30.0} & \makebox[24pt][r]{20.0} & \makebox[24pt][r]{9.0} & \makebox[24pt][r]{19.0} & \makebox[24pt][r]{27.2} & \makebox[24pt][r]{\textbf{95.0}} & \makebox[24pt][r]{65.0} & \makebox[24pt][r]{\underline{68.0}} & \makebox[24pt][r]{60.0} & \makebox[24pt][r]{\underline{54.0}} & \makebox[24pt][r]{\underline{39.0}} & \makebox[24pt][r]{\underline{28.0}} & \makebox[24pt][r]{\underline{52.3}} & \makebox[24pt][r]{39.8} \\
    RAVEN & \makebox[24pt][r]{89.0} & \makebox[24pt][r]{79.0} & \makebox[24pt][r]{45.0} & \makebox[24pt][r]{31.0} & \makebox[24pt][r]{37.0} & \makebox[24pt][r]{22.0} & \makebox[24pt][r]{42.8} & \makebox[24pt][r]{91.0} & \makebox[24pt][r]{\textbf{88.0}} & \makebox[24pt][r]{65.0} & \makebox[24pt][r]{41.0} & \makebox[24pt][r]{11.0} & \makebox[24pt][r]{7.0} & \makebox[24pt][r]{9.0} & \makebox[24pt][r]{36.8} & \makebox[24pt][r]{39.8} \\
    \cmidrule(lr){1-17}
    \textbf{ReGeoAnch} & \makebox[24pt][r]{88.0} & \makebox[24pt][r]{\textbf{86.0}} & \makebox[24pt][r]{\textbf{90.0}} & \makebox[24pt][r]{\textbf{85.0}} & \makebox[24pt][r]{\textbf{85.0}} & \makebox[24pt][r]{\textbf{75.0}} & \makebox[24pt][r]{\textbf{84.2}} & \makebox[24pt][r]{89.0} & \makebox[24pt][r]{\underline{80.0}} & \makebox[24pt][r]{\underline{68.0}} & \makebox[24pt][r]{\textbf{68.0}} & \makebox[24pt][r]{\textbf{59.0}} & \makebox[24pt][r]{\textbf{51.0}} & \makebox[24pt][r]{\textbf{35.0}} & \makebox[24pt][r]{\textbf{60.2}} & \makebox[24pt][r]{\textbf{72.2}} \\
    \bottomrule
    \end{tabular*}
    \endgroup
\end{table*}

\begin{table*}[p]
    \centering
    \caption{Per-view success rates (\%) on MetaWorld: tasks 5--8. Formatting and aggregation follow Table~\ref{tab:supp-mw-angles-1}.}
    \label{tab:supp-mw-angles-2}
    \begingroup
    \fontsize{9}{10.5}\selectfont
    \setlength{\tabcolsep}{0.5pt}
    \setlength{\aboverulesep}{1.5pt}
    \setlength{\belowrulesep}{1.5pt}
    \renewcommand{\arraystretch}{1.0}
    \begin{tabular*}{0.98\textwidth}{@{\extracolsep{\fill}}p{56pt}*{16}{r}@{}}
    \toprule
    & \multicolumn{7}{c}{Azimuth} & \multicolumn{8}{c}{Elevation} & \makebox[24pt][c]{Task} \\
    \cmidrule(lr){2-8}\cmidrule(lr){9-16}
    \textbf{Method} & \makebox[24pt][c]{$0^\circ$} & \makebox[24pt][c]{$-5^\circ$} & \makebox[24pt][c]{$-10^\circ$} & \makebox[24pt][c]{$-15^\circ$} & \makebox[24pt][c]{$-20^\circ$} & \makebox[24pt][c]{$-30^\circ$} & \makebox[24pt][c]{\textbf{Avg.}} & \makebox[24pt][c]{$0^\circ$} & \makebox[24pt][c]{$+5^\circ$} & \makebox[24pt][c]{$+10^\circ$} & \makebox[24pt][c]{$+15^\circ$} & \makebox[24pt][c]{$+20^\circ$} & \makebox[24pt][c]{$+25^\circ$} & \makebox[24pt][c]{$+30^\circ$} & \makebox[24pt][c]{\textbf{Avg.}} & \makebox[24pt][c]{\textbf{Avg.}} \\
    \midrule
    \multicolumn{17}{@{}l}{\textbf{(a) Coffee Button}} \\
    DP & \makebox[24pt][r]{\textbf{100.0}} & \makebox[24pt][r]{\textbf{100.0}} & \makebox[24pt][r]{\underline{99.0}} & \makebox[24pt][r]{\underline{88.0}} & \makebox[24pt][r]{73.0} & \makebox[24pt][r]{69.0} & \makebox[24pt][r]{85.8} & \makebox[24pt][r]{\textbf{100.0}} & \makebox[24pt][r]{\textbf{100.0}} & \makebox[24pt][r]{\underline{96.0}} & \makebox[24pt][r]{84.0} & \makebox[24pt][r]{73.0} & \makebox[24pt][r]{84.0} & \makebox[24pt][r]{65.0} & \makebox[24pt][r]{83.7} & \makebox[24pt][r]{84.7} \\
    DP\({}^{\dagger}\) & \makebox[24pt][r]{\textbf{100.0}} & \makebox[24pt][r]{\textbf{100.0}} & \makebox[24pt][r]{\textbf{100.0}} & \makebox[24pt][r]{\textbf{100.0}} & \makebox[24pt][r]{\underline{92.0}} & \makebox[24pt][r]{86.0} & \makebox[24pt][r]{95.6} & \makebox[24pt][r]{\textbf{100.0}} & \makebox[24pt][r]{\textbf{100.0}} & \makebox[24pt][r]{\textbf{100.0}} & \makebox[24pt][r]{\underline{97.0}} & \makebox[24pt][r]{\underline{90.0}} & \makebox[24pt][r]{\underline{98.0}} & \makebox[24pt][r]{70.0} & \makebox[24pt][r]{92.5} & \makebox[24pt][r]{94.1} \\
    DP3 & \makebox[24pt][r]{\textbf{100.0}} & \makebox[24pt][r]{\textbf{100.0}} & \makebox[24pt][r]{\textbf{100.0}} & \makebox[24pt][r]{\textbf{100.0}} & \makebox[24pt][r]{\textbf{100.0}} & \makebox[24pt][r]{\textbf{100.0}} & \makebox[24pt][r]{\textbf{100.0}} & \makebox[24pt][r]{\textbf{100.0}} & \makebox[24pt][r]{\textbf{100.0}} & \makebox[24pt][r]{\textbf{100.0}} & \makebox[24pt][r]{\textbf{100.0}} & \makebox[24pt][r]{\textbf{100.0}} & \makebox[24pt][r]{\textbf{100.0}} & \makebox[24pt][r]{95.0} & \makebox[24pt][r]{99.2} & \makebox[24pt][r]{99.6} \\
    3D-DA & \makebox[24pt][r]{\textbf{100.0}} & \makebox[24pt][r]{\textbf{100.0}} & \makebox[24pt][r]{\textbf{100.0}} & \makebox[24pt][r]{\textbf{100.0}} & \makebox[24pt][r]{\textbf{100.0}} & \makebox[24pt][r]{\underline{99.0}} & \makebox[24pt][r]{\underline{99.8}} & \makebox[24pt][r]{\textbf{100.0}} & \makebox[24pt][r]{\textbf{100.0}} & \makebox[24pt][r]{\textbf{100.0}} & \makebox[24pt][r]{\textbf{100.0}} & \makebox[24pt][r]{\textbf{100.0}} & \makebox[24pt][r]{\textbf{100.0}} & \makebox[24pt][r]{\textbf{100.0}} & \makebox[24pt][r]{\textbf{100.0}} & \makebox[24pt][r]{\underline{99.9}} \\
    ReViWo & \makebox[24pt][r]{\textbf{100.0}} & \makebox[24pt][r]{\textbf{100.0}} & \makebox[24pt][r]{\textbf{100.0}} & \makebox[24pt][r]{\textbf{100.0}} & \makebox[24pt][r]{\textbf{100.0}} & \makebox[24pt][r]{\textbf{100.0}} & \makebox[24pt][r]{\textbf{100.0}} & \makebox[24pt][r]{\textbf{100.0}} & \makebox[24pt][r]{\textbf{100.0}} & \makebox[24pt][r]{\textbf{100.0}} & \makebox[24pt][r]{\textbf{100.0}} & \makebox[24pt][r]{\textbf{100.0}} & \makebox[24pt][r]{\textbf{100.0}} & \makebox[24pt][r]{\textbf{100.0}} & \makebox[24pt][r]{\textbf{100.0}} & \makebox[24pt][r]{\textbf{100.0}} \\
    CLASS & \makebox[24pt][r]{\textbf{100.0}} & \makebox[24pt][r]{\textbf{100.0}} & \makebox[24pt][r]{\textbf{100.0}} & \makebox[24pt][r]{\textbf{100.0}} & \makebox[24pt][r]{\textbf{100.0}} & \makebox[24pt][r]{\textbf{100.0}} & \makebox[24pt][r]{\textbf{100.0}} & \makebox[24pt][r]{\textbf{100.0}} & \makebox[24pt][r]{\textbf{100.0}} & \makebox[24pt][r]{\textbf{100.0}} & \makebox[24pt][r]{\textbf{100.0}} & \makebox[24pt][r]{\textbf{100.0}} & \makebox[24pt][r]{\textbf{100.0}} & \makebox[24pt][r]{\textbf{100.0}} & \makebox[24pt][r]{\textbf{100.0}} & \makebox[24pt][r]{\textbf{100.0}} \\
    Cam-DP & \makebox[24pt][r]{\textbf{100.0}} & \makebox[24pt][r]{\textbf{100.0}} & \makebox[24pt][r]{\textbf{100.0}} & \makebox[24pt][r]{\textbf{100.0}} & \makebox[24pt][r]{\textbf{100.0}} & \makebox[24pt][r]{\underline{99.0}} & \makebox[24pt][r]{\underline{99.8}} & \makebox[24pt][r]{\textbf{100.0}} & \makebox[24pt][r]{\textbf{100.0}} & \makebox[24pt][r]{\textbf{100.0}} & \makebox[24pt][r]{\textbf{100.0}} & \makebox[24pt][r]{\textbf{100.0}} & \makebox[24pt][r]{\textbf{100.0}} & \makebox[24pt][r]{\underline{97.0}} & \makebox[24pt][r]{\underline{99.5}} & \makebox[24pt][r]{99.7} \\
    RAVEN & \makebox[24pt][r]{\textbf{100.0}} & \makebox[24pt][r]{\textbf{100.0}} & \makebox[24pt][r]{\textbf{100.0}} & \makebox[24pt][r]{\textbf{100.0}} & \makebox[24pt][r]{\textbf{100.0}} & \makebox[24pt][r]{\textbf{100.0}} & \makebox[24pt][r]{\textbf{100.0}} & \makebox[24pt][r]{\textbf{100.0}} & \makebox[24pt][r]{\textbf{100.0}} & \makebox[24pt][r]{\textbf{100.0}} & \makebox[24pt][r]{\textbf{100.0}} & \makebox[24pt][r]{\textbf{100.0}} & \makebox[24pt][r]{\textbf{100.0}} & \makebox[24pt][r]{\textbf{100.0}} & \makebox[24pt][r]{\textbf{100.0}} & \makebox[24pt][r]{\textbf{100.0}} \\
    \cmidrule(lr){1-17}
    \textbf{ReGeoAnch} & \makebox[24pt][r]{\textbf{100.0}} & \makebox[24pt][r]{\textbf{100.0}} & \makebox[24pt][r]{\textbf{100.0}} & \makebox[24pt][r]{\textbf{100.0}} & \makebox[24pt][r]{\textbf{100.0}} & \makebox[24pt][r]{\textbf{100.0}} & \makebox[24pt][r]{\textbf{100.0}} & \makebox[24pt][r]{\textbf{100.0}} & \makebox[24pt][r]{\textbf{100.0}} & \makebox[24pt][r]{\textbf{100.0}} & \makebox[24pt][r]{\textbf{100.0}} & \makebox[24pt][r]{\textbf{100.0}} & \makebox[24pt][r]{\textbf{100.0}} & \makebox[24pt][r]{\textbf{100.0}} & \makebox[24pt][r]{\textbf{100.0}} & \makebox[24pt][r]{\textbf{100.0}} \\
    \midrule
    \multicolumn{17}{@{}l}{\textbf{(b) Door Open}} \\
    DP & \makebox[24pt][r]{98.0} & \makebox[24pt][r]{92.0} & \makebox[24pt][r]{87.0} & \makebox[24pt][r]{35.0} & \makebox[24pt][r]{15.0} & \makebox[24pt][r]{31.0} & \makebox[24pt][r]{52.0} & \makebox[24pt][r]{97.0} & \makebox[24pt][r]{97.0} & \makebox[24pt][r]{93.0} & \makebox[24pt][r]{29.0} & \makebox[24pt][r]{4.0} & \makebox[24pt][r]{0.0} & \makebox[24pt][r]{0.0} & \makebox[24pt][r]{37.2} & \makebox[24pt][r]{44.6} \\
    DP\({}^{\dagger}\) & \makebox[24pt][r]{98.0} & \makebox[24pt][r]{\textbf{100.0}} & \makebox[24pt][r]{97.0} & \makebox[24pt][r]{98.0} & \makebox[24pt][r]{97.0} & \makebox[24pt][r]{\textbf{100.0}} & \makebox[24pt][r]{98.4} & \makebox[24pt][r]{97.0} & \makebox[24pt][r]{91.0} & \makebox[24pt][r]{\textbf{100.0}} & \makebox[24pt][r]{97.0} & \makebox[24pt][r]{96.0} & \makebox[24pt][r]{90.0} & \makebox[24pt][r]{83.0} & \makebox[24pt][r]{92.8} & \makebox[24pt][r]{95.6} \\
    DP3 & \makebox[24pt][r]{\underline{99.0}} & \makebox[24pt][r]{\underline{98.0}} & \makebox[24pt][r]{\underline{98.0}} & \makebox[24pt][r]{\underline{99.0}} & \makebox[24pt][r]{\textbf{100.0}} & \makebox[24pt][r]{\textbf{100.0}} & \makebox[24pt][r]{99.0} & \makebox[24pt][r]{\textbf{99.0}} & \makebox[24pt][r]{98.0} & \makebox[24pt][r]{\underline{99.0}} & \makebox[24pt][r]{\textbf{100.0}} & \makebox[24pt][r]{\textbf{100.0}} & \makebox[24pt][r]{\textbf{100.0}} & \makebox[24pt][r]{\textbf{100.0}} & \makebox[24pt][r]{\underline{99.5}} & \makebox[24pt][r]{\underline{99.3}} \\
    3D-DA & \makebox[24pt][r]{98.0} & \makebox[24pt][r]{\textbf{100.0}} & \makebox[24pt][r]{94.0} & \makebox[24pt][r]{\underline{99.0}} & \makebox[24pt][r]{\textbf{100.0}} & \makebox[24pt][r]{\textbf{100.0}} & \makebox[24pt][r]{98.6} & \makebox[24pt][r]{\underline{98.0}} & \makebox[24pt][r]{90.0} & \makebox[24pt][r]{96.0} & \makebox[24pt][r]{98.0} & \makebox[24pt][r]{95.0} & \makebox[24pt][r]{\underline{99.0}} & \makebox[24pt][r]{\underline{99.0}} & \makebox[24pt][r]{96.2} & \makebox[24pt][r]{97.4} \\
    ReViWo & \makebox[24pt][r]{\textbf{100.0}} & \makebox[24pt][r]{95.0} & \makebox[24pt][r]{97.0} & \makebox[24pt][r]{\textbf{100.0}} & \makebox[24pt][r]{\textbf{100.0}} & \makebox[24pt][r]{\textbf{100.0}} & \makebox[24pt][r]{98.4} & \makebox[24pt][r]{\underline{98.0}} & \makebox[24pt][r]{\underline{99.0}} & \makebox[24pt][r]{\underline{99.0}} & \makebox[24pt][r]{\underline{99.0}} & \makebox[24pt][r]{\underline{99.0}} & \makebox[24pt][r]{\textbf{100.0}} & \makebox[24pt][r]{\underline{99.0}} & \makebox[24pt][r]{99.2} & \makebox[24pt][r]{98.8} \\
    CLASS & \makebox[24pt][r]{98.0} & \makebox[24pt][r]{\textbf{100.0}} & \makebox[24pt][r]{95.0} & \makebox[24pt][r]{97.0} & \makebox[24pt][r]{\underline{98.0}} & \makebox[24pt][r]{88.0} & \makebox[24pt][r]{95.6} & \makebox[24pt][r]{\underline{98.0}} & \makebox[24pt][r]{95.0} & \makebox[24pt][r]{88.0} & \makebox[24pt][r]{80.0} & \makebox[24pt][r]{84.0} & \makebox[24pt][r]{85.0} & \makebox[24pt][r]{86.0} & \makebox[24pt][r]{86.3} & \makebox[24pt][r]{91.0} \\
    Cam-DP & \makebox[24pt][r]{97.0} & \makebox[24pt][r]{\textbf{100.0}} & \makebox[24pt][r]{\textbf{100.0}} & \makebox[24pt][r]{\textbf{100.0}} & \makebox[24pt][r]{\textbf{100.0}} & \makebox[24pt][r]{\underline{99.0}} & \makebox[24pt][r]{\underline{99.8}} & \makebox[24pt][r]{\underline{98.0}} & \makebox[24pt][r]{86.0} & \makebox[24pt][r]{94.0} & \makebox[24pt][r]{\underline{99.0}} & \makebox[24pt][r]{\textbf{100.0}} & \makebox[24pt][r]{98.0} & \makebox[24pt][r]{\textbf{100.0}} & \makebox[24pt][r]{96.2} & \makebox[24pt][r]{98.0} \\
    RAVEN & \makebox[24pt][r]{\textbf{100.0}} & \makebox[24pt][r]{\textbf{100.0}} & \makebox[24pt][r]{94.0} & \makebox[24pt][r]{82.0} & \makebox[24pt][r]{72.0} & \makebox[24pt][r]{66.0} & \makebox[24pt][r]{82.8} & \makebox[24pt][r]{\textbf{99.0}} & \makebox[24pt][r]{\textbf{100.0}} & \makebox[24pt][r]{98.0} & \makebox[24pt][r]{97.0} & \makebox[24pt][r]{98.0} & \makebox[24pt][r]{95.0} & \makebox[24pt][r]{90.0} & \makebox[24pt][r]{96.3} & \makebox[24pt][r]{89.6} \\
    \cmidrule(lr){1-17}
    \textbf{ReGeoAnch} & \makebox[24pt][r]{\underline{99.0}} & \makebox[24pt][r]{\textbf{100.0}} & \makebox[24pt][r]{\textbf{100.0}} & \makebox[24pt][r]{\textbf{100.0}} & \makebox[24pt][r]{\textbf{100.0}} & \makebox[24pt][r]{\textbf{100.0}} & \makebox[24pt][r]{\textbf{100.0}} & \makebox[24pt][r]{\underline{98.0}} & \makebox[24pt][r]{\textbf{100.0}} & \makebox[24pt][r]{\textbf{100.0}} & \makebox[24pt][r]{\textbf{100.0}} & \makebox[24pt][r]{\textbf{100.0}} & \makebox[24pt][r]{\textbf{100.0}} & \makebox[24pt][r]{\textbf{100.0}} & \makebox[24pt][r]{\textbf{100.0}} & \makebox[24pt][r]{\textbf{100.0}} \\
    \midrule
    \multicolumn{17}{@{}l}{\textbf{(c) Window Close}} \\
    DP & \makebox[24pt][r]{\textbf{100.0}} & \makebox[24pt][r]{\textbf{100.0}} & \makebox[24pt][r]{\textbf{100.0}} & \makebox[24pt][r]{\textbf{100.0}} & \makebox[24pt][r]{\underline{98.0}} & \makebox[24pt][r]{52.0} & \makebox[24pt][r]{90.0} & \makebox[24pt][r]{\textbf{100.0}} & \makebox[24pt][r]{\textbf{100.0}} & \makebox[24pt][r]{\textbf{100.0}} & \makebox[24pt][r]{\underline{95.0}} & \makebox[24pt][r]{\underline{91.0}} & \makebox[24pt][r]{82.0} & \makebox[24pt][r]{79.0} & \makebox[24pt][r]{91.2} & \makebox[24pt][r]{90.6} \\
    DP\({}^{\dagger}\) & \makebox[24pt][r]{\textbf{100.0}} & \makebox[24pt][r]{\textbf{100.0}} & \makebox[24pt][r]{\textbf{100.0}} & \makebox[24pt][r]{\textbf{100.0}} & \makebox[24pt][r]{\textbf{100.0}} & \makebox[24pt][r]{\underline{88.0}} & \makebox[24pt][r]{\underline{97.6}} & \makebox[24pt][r]{\textbf{100.0}} & \makebox[24pt][r]{\textbf{100.0}} & \makebox[24pt][r]{\textbf{100.0}} & \makebox[24pt][r]{\textbf{100.0}} & \makebox[24pt][r]{\textbf{100.0}} & \makebox[24pt][r]{\underline{99.0}} & \makebox[24pt][r]{\underline{98.0}} & \makebox[24pt][r]{\underline{99.5}} & \makebox[24pt][r]{\underline{98.6}} \\
    DP3 & \makebox[24pt][r]{\textbf{100.0}} & \makebox[24pt][r]{\textbf{100.0}} & \makebox[24pt][r]{\textbf{100.0}} & \makebox[24pt][r]{\textbf{100.0}} & \makebox[24pt][r]{\textbf{100.0}} & \makebox[24pt][r]{\textbf{100.0}} & \makebox[24pt][r]{\textbf{100.0}} & \makebox[24pt][r]{\textbf{100.0}} & \makebox[24pt][r]{\textbf{100.0}} & \makebox[24pt][r]{\textbf{100.0}} & \makebox[24pt][r]{\textbf{100.0}} & \makebox[24pt][r]{\textbf{100.0}} & \makebox[24pt][r]{\textbf{100.0}} & \makebox[24pt][r]{\textbf{100.0}} & \makebox[24pt][r]{\textbf{100.0}} & \makebox[24pt][r]{\textbf{100.0}} \\
    3D-DA & \makebox[24pt][r]{\textbf{100.0}} & \makebox[24pt][r]{\textbf{100.0}} & \makebox[24pt][r]{\textbf{100.0}} & \makebox[24pt][r]{\textbf{100.0}} & \makebox[24pt][r]{\textbf{100.0}} & \makebox[24pt][r]{\textbf{100.0}} & \makebox[24pt][r]{\textbf{100.0}} & \makebox[24pt][r]{\textbf{100.0}} & \makebox[24pt][r]{\textbf{100.0}} & \makebox[24pt][r]{\textbf{100.0}} & \makebox[24pt][r]{\textbf{100.0}} & \makebox[24pt][r]{\textbf{100.0}} & \makebox[24pt][r]{\textbf{100.0}} & \makebox[24pt][r]{\textbf{100.0}} & \makebox[24pt][r]{\textbf{100.0}} & \makebox[24pt][r]{\textbf{100.0}} \\
    ReViWo & \makebox[24pt][r]{\textbf{100.0}} & \makebox[24pt][r]{\textbf{100.0}} & \makebox[24pt][r]{\textbf{100.0}} & \makebox[24pt][r]{\textbf{100.0}} & \makebox[24pt][r]{\textbf{100.0}} & \makebox[24pt][r]{\textbf{100.0}} & \makebox[24pt][r]{\textbf{100.0}} & \makebox[24pt][r]{\textbf{100.0}} & \makebox[24pt][r]{\textbf{100.0}} & \makebox[24pt][r]{\textbf{100.0}} & \makebox[24pt][r]{\textbf{100.0}} & \makebox[24pt][r]{\textbf{100.0}} & \makebox[24pt][r]{\textbf{100.0}} & \makebox[24pt][r]{\textbf{100.0}} & \makebox[24pt][r]{\textbf{100.0}} & \makebox[24pt][r]{\textbf{100.0}} \\
    CLASS & \makebox[24pt][r]{\textbf{100.0}} & \makebox[24pt][r]{\textbf{100.0}} & \makebox[24pt][r]{\textbf{100.0}} & \makebox[24pt][r]{\textbf{100.0}} & \makebox[24pt][r]{\textbf{100.0}} & \makebox[24pt][r]{\textbf{100.0}} & \makebox[24pt][r]{\textbf{100.0}} & \makebox[24pt][r]{\textbf{100.0}} & \makebox[24pt][r]{\textbf{100.0}} & \makebox[24pt][r]{\textbf{100.0}} & \makebox[24pt][r]{\textbf{100.0}} & \makebox[24pt][r]{\textbf{100.0}} & \makebox[24pt][r]{\textbf{100.0}} & \makebox[24pt][r]{\textbf{100.0}} & \makebox[24pt][r]{\textbf{100.0}} & \makebox[24pt][r]{\textbf{100.0}} \\
    Cam-DP & \makebox[24pt][r]{\textbf{100.0}} & \makebox[24pt][r]{\textbf{100.0}} & \makebox[24pt][r]{\textbf{100.0}} & \makebox[24pt][r]{\textbf{100.0}} & \makebox[24pt][r]{\textbf{100.0}} & \makebox[24pt][r]{\textbf{100.0}} & \makebox[24pt][r]{\textbf{100.0}} & \makebox[24pt][r]{\textbf{100.0}} & \makebox[24pt][r]{\textbf{100.0}} & \makebox[24pt][r]{\textbf{100.0}} & \makebox[24pt][r]{\textbf{100.0}} & \makebox[24pt][r]{\textbf{100.0}} & \makebox[24pt][r]{\textbf{100.0}} & \makebox[24pt][r]{\textbf{100.0}} & \makebox[24pt][r]{\textbf{100.0}} & \makebox[24pt][r]{\textbf{100.0}} \\
    RAVEN & \makebox[24pt][r]{\textbf{100.0}} & \makebox[24pt][r]{\textbf{100.0}} & \makebox[24pt][r]{\textbf{100.0}} & \makebox[24pt][r]{\textbf{100.0}} & \makebox[24pt][r]{\textbf{100.0}} & \makebox[24pt][r]{75.0} & \makebox[24pt][r]{95.0} & \makebox[24pt][r]{\textbf{100.0}} & \makebox[24pt][r]{\textbf{100.0}} & \makebox[24pt][r]{\textbf{100.0}} & \makebox[24pt][r]{\textbf{100.0}} & \makebox[24pt][r]{\textbf{100.0}} & \makebox[24pt][r]{\textbf{100.0}} & \makebox[24pt][r]{\textbf{100.0}} & \makebox[24pt][r]{\textbf{100.0}} & \makebox[24pt][r]{97.5} \\
    \cmidrule(lr){1-17}
    \textbf{ReGeoAnch} & \makebox[24pt][r]{\textbf{100.0}} & \makebox[24pt][r]{\textbf{100.0}} & \makebox[24pt][r]{\textbf{100.0}} & \makebox[24pt][r]{\textbf{100.0}} & \makebox[24pt][r]{\textbf{100.0}} & \makebox[24pt][r]{\textbf{100.0}} & \makebox[24pt][r]{\textbf{100.0}} & \makebox[24pt][r]{\textbf{100.0}} & \makebox[24pt][r]{\textbf{100.0}} & \makebox[24pt][r]{\textbf{100.0}} & \makebox[24pt][r]{\textbf{100.0}} & \makebox[24pt][r]{\textbf{100.0}} & \makebox[24pt][r]{\textbf{100.0}} & \makebox[24pt][r]{\textbf{100.0}} & \makebox[24pt][r]{\textbf{100.0}} & \makebox[24pt][r]{\textbf{100.0}} \\
    \midrule
    \multicolumn{17}{@{}l}{\textbf{(d) Assembly}} \\
    DP & \makebox[24pt][r]{\underline{99.0}} & \makebox[24pt][r]{21.0} & \makebox[24pt][r]{0.0} & \makebox[24pt][r]{0.0} & \makebox[24pt][r]{0.0} & \makebox[24pt][r]{0.0} & \makebox[24pt][r]{4.2} & \makebox[24pt][r]{97.0} & \makebox[24pt][r]{0.0} & \makebox[24pt][r]{0.0} & \makebox[24pt][r]{0.0} & \makebox[24pt][r]{0.0} & \makebox[24pt][r]{0.0} & \makebox[24pt][r]{0.0} & \makebox[24pt][r]{0.0} & \makebox[24pt][r]{2.1} \\
    DP\({}^{\dagger}\) & \makebox[24pt][r]{\underline{99.0}} & \makebox[24pt][r]{92.0} & \makebox[24pt][r]{77.0} & \makebox[24pt][r]{73.0} & \makebox[24pt][r]{56.0} & \makebox[24pt][r]{13.0} & \makebox[24pt][r]{62.2} & \makebox[24pt][r]{\textbf{99.0}} & \makebox[24pt][r]{97.0} & \makebox[24pt][r]{75.0} & \makebox[24pt][r]{62.0} & \makebox[24pt][r]{57.0} & \makebox[24pt][r]{50.0} & \makebox[24pt][r]{38.0} & \makebox[24pt][r]{63.2} & \makebox[24pt][r]{62.7} \\
    DP3 & \makebox[24pt][r]{97.0} & \makebox[24pt][r]{96.0} & \makebox[24pt][r]{93.0} & \makebox[24pt][r]{75.0} & \makebox[24pt][r]{58.0} & \makebox[24pt][r]{32.0} & \makebox[24pt][r]{70.8} & \makebox[24pt][r]{80.0} & \makebox[24pt][r]{75.0} & \makebox[24pt][r]{62.0} & \makebox[24pt][r]{66.0} & \makebox[24pt][r]{50.0} & \makebox[24pt][r]{52.0} & \makebox[24pt][r]{35.0} & \makebox[24pt][r]{56.7} & \makebox[24pt][r]{63.7} \\
    3D-DA & \makebox[24pt][r]{\underline{99.0}} & \makebox[24pt][r]{94.0} & \makebox[24pt][r]{92.0} & \makebox[24pt][r]{90.0} & \makebox[24pt][r]{73.0} & \makebox[24pt][r]{38.0} & \makebox[24pt][r]{77.4} & \makebox[24pt][r]{97.0} & \makebox[24pt][r]{\underline{99.0}} & \makebox[24pt][r]{\textbf{100.0}} & \makebox[24pt][r]{92.0} & \makebox[24pt][r]{37.0} & \makebox[24pt][r]{\underline{96.0}} & \makebox[24pt][r]{61.0} & \makebox[24pt][r]{80.8} & \makebox[24pt][r]{79.1} \\
    ReViWo & \makebox[24pt][r]{92.0} & \makebox[24pt][r]{91.0} & \makebox[24pt][r]{85.0} & \makebox[24pt][r]{49.0} & \makebox[24pt][r]{65.0} & \makebox[24pt][r]{66.0} & \makebox[24pt][r]{71.2} & \makebox[24pt][r]{91.0} & \makebox[24pt][r]{84.0} & \makebox[24pt][r]{77.0} & \makebox[24pt][r]{80.0} & \makebox[24pt][r]{82.0} & \makebox[24pt][r]{79.0} & \makebox[24pt][r]{58.0} & \makebox[24pt][r]{76.7} & \makebox[24pt][r]{73.9} \\
    CLASS & \makebox[24pt][r]{96.0} & \makebox[24pt][r]{\textbf{100.0}} & \makebox[24pt][r]{\textbf{96.0}} & \makebox[24pt][r]{\textbf{93.0}} & \makebox[24pt][r]{\textbf{93.0}} & \makebox[24pt][r]{\textbf{98.0}} & \makebox[24pt][r]{\textbf{96.0}} & \makebox[24pt][r]{95.0} & \makebox[24pt][r]{98.0} & \makebox[24pt][r]{\underline{99.0}} & \makebox[24pt][r]{\textbf{100.0}} & \makebox[24pt][r]{\textbf{99.0}} & \makebox[24pt][r]{\textbf{98.0}} & \makebox[24pt][r]{\textbf{98.0}} & \makebox[24pt][r]{\textbf{98.7}} & \makebox[24pt][r]{\textbf{97.3}} \\
    Cam-DP & \makebox[24pt][r]{95.0} & \makebox[24pt][r]{\underline{97.0}} & \makebox[24pt][r]{\textbf{96.0}} & \makebox[24pt][r]{\underline{92.0}} & \makebox[24pt][r]{\underline{92.0}} & \makebox[24pt][r]{52.0} & \makebox[24pt][r]{85.8} & \makebox[24pt][r]{\underline{98.0}} & \makebox[24pt][r]{\textbf{100.0}} & \makebox[24pt][r]{\underline{99.0}} & \makebox[24pt][r]{\underline{97.0}} & \makebox[24pt][r]{\underline{87.0}} & \makebox[24pt][r]{72.0} & \makebox[24pt][r]{54.0} & \makebox[24pt][r]{84.8} & \makebox[24pt][r]{85.3} \\
    RAVEN & \makebox[24pt][r]{\textbf{100.0}} & \makebox[24pt][r]{66.0} & \makebox[24pt][r]{43.0} & \makebox[24pt][r]{32.0} & \makebox[24pt][r]{44.0} & \makebox[24pt][r]{43.0} & \makebox[24pt][r]{45.6} & \makebox[24pt][r]{96.0} & \makebox[24pt][r]{91.0} & \makebox[24pt][r]{81.0} & \makebox[24pt][r]{64.0} & \makebox[24pt][r]{54.0} & \makebox[24pt][r]{52.0} & \makebox[24pt][r]{74.0} & \makebox[24pt][r]{69.3} & \makebox[24pt][r]{57.5} \\
    \cmidrule(lr){1-17}
    \textbf{ReGeoAnch} & \makebox[24pt][r]{98.0} & \makebox[24pt][r]{96.0} & \makebox[24pt][r]{\underline{95.0}} & \makebox[24pt][r]{\textbf{93.0}} & \makebox[24pt][r]{91.0} & \makebox[24pt][r]{\underline{95.0}} & \makebox[24pt][r]{\underline{94.0}} & \makebox[24pt][r]{96.0} & \makebox[24pt][r]{94.0} & \makebox[24pt][r]{90.0} & \makebox[24pt][r]{91.0} & \makebox[24pt][r]{\underline{87.0}} & \makebox[24pt][r]{88.0} & \makebox[24pt][r]{\underline{81.0}} & \makebox[24pt][r]{\underline{88.5}} & \makebox[24pt][r]{\underline{91.3}} \\
    \bottomrule
    \end{tabular*}
    \endgroup
\end{table*}

\begin{table*}[p]
    \centering
    \caption{Per-view success rates (\%) on ManiSkill3. Task Avg. matches the task scores in Table~\ref{tab:main-results}, with equal weighting of azimuth, elevation and distance-scale means. Other conventions follow Table~\ref{tab:supp-mw-angles-1}.}
    \label{tab:supp-ms-angles}
    \begingroup
    \fontsize{9}{10.5}\selectfont
    \setlength{\tabcolsep}{0.5pt}
    \setlength{\aboverulesep}{1.5pt}
    \setlength{\belowrulesep}{1.5pt}
    \renewcommand{\arraystretch}{1.0}
    \begin{tabular*}{0.98\textwidth}{@{\extracolsep{\fill}}p{56pt}*{16}{r}@{}}
    \toprule
    & \multicolumn{7}{c}{Azimuth} & \multicolumn{8}{c}{Elevation} & \makebox[24pt][c]{Task} \\
    \cmidrule(lr){2-8}\cmidrule(lr){9-16}
    \textbf{Method} & \makebox[24pt][c]{$0^\circ$} & \makebox[24pt][c]{$-5^\circ$} & \makebox[24pt][c]{$-10^\circ$} & \makebox[24pt][c]{$-15^\circ$} & \makebox[24pt][c]{$-20^\circ$} & \makebox[24pt][c]{$-30^\circ$} & \makebox[24pt][c]{\textbf{Avg.}} & \makebox[24pt][c]{$-10^\circ$} & \makebox[24pt][c]{$-5^\circ$} & \makebox[24pt][c]{$0^\circ$} & \makebox[24pt][c]{$+5^\circ$} & \makebox[24pt][c]{$+10^\circ$} & \makebox[24pt][c]{$+15^\circ$} & \makebox[24pt][c]{$+20^\circ$} & \makebox[24pt][c]{\textbf{Avg.}} & \makebox[24pt][c]{\textbf{Avg.}} \\
    \midrule
    \multicolumn{17}{@{}l}{\textbf{(a) PickCube}} \\
    DP & \makebox[24pt][r]{89.0} & \makebox[24pt][r]{91.0} & \makebox[24pt][r]{75.0} & \makebox[24pt][r]{53.0} & \makebox[24pt][r]{31.0} & \makebox[24pt][r]{14.0} & \makebox[24pt][r]{52.8} & \makebox[24pt][r]{52.0} & \makebox[24pt][r]{85.0} & \makebox[24pt][r]{89.0} & \makebox[24pt][r]{72.0} & \makebox[24pt][r]{36.0} & \makebox[24pt][r]{8.0} & \makebox[24pt][r]{2.0} & \makebox[24pt][r]{42.5} & \makebox[24pt][r]{53.8} \\
    DP\({}^{\dagger}\) & \makebox[24pt][r]{94.0} & \makebox[24pt][r]{95.0} & \makebox[24pt][r]{92.0} & \makebox[24pt][r]{94.0} & \makebox[24pt][r]{81.0} & \makebox[24pt][r]{39.0} & \makebox[24pt][r]{80.2} & \makebox[24pt][r]{75.0} & \makebox[24pt][r]{88.0} & \makebox[24pt][r]{92.0} & \makebox[24pt][r]{93.0} & \makebox[24pt][r]{84.0} & \makebox[24pt][r]{63.0} & \makebox[24pt][r]{29.0} & \makebox[24pt][r]{72.0} & \makebox[24pt][r]{79.6} \\
    DP3 & \makebox[24pt][r]{91.0} & \makebox[24pt][r]{90.0} & \makebox[24pt][r]{88.0} & \makebox[24pt][r]{73.0} & \makebox[24pt][r]{58.0} & \makebox[24pt][r]{14.0} & \makebox[24pt][r]{64.6} & \makebox[24pt][r]{78.0} & \makebox[24pt][r]{89.0} & \makebox[24pt][r]{\underline{99.0}} & \makebox[24pt][r]{93.0} & \makebox[24pt][r]{88.0} & \makebox[24pt][r]{65.0} & \makebox[24pt][r]{27.0} & \makebox[24pt][r]{73.3} & \makebox[24pt][r]{60.0} \\
    3D-DA & \makebox[24pt][r]{93.0} & \makebox[24pt][r]{\underline{96.0}} & \makebox[24pt][r]{96.0} & \makebox[24pt][r]{92.0} & \makebox[24pt][r]{86.0} & \makebox[24pt][r]{\underline{52.0}} & \makebox[24pt][r]{\underline{84.4}} & \makebox[24pt][r]{81.0} & \makebox[24pt][r]{92.0} & \makebox[24pt][r]{92.0} & \makebox[24pt][r]{94.0} & \makebox[24pt][r]{90.0} & \makebox[24pt][r]{73.0} & \makebox[24pt][r]{30.0} & \makebox[24pt][r]{76.7} & \makebox[24pt][r]{84.8} \\
    ReViWo & \makebox[24pt][r]{62.0} & \makebox[24pt][r]{42.0} & \makebox[24pt][r]{18.0} & \makebox[24pt][r]{2.0} & \makebox[24pt][r]{0.0} & \makebox[24pt][r]{1.0} & \makebox[24pt][r]{12.6} & \makebox[24pt][r]{42.0} & \makebox[24pt][r]{60.0} & \makebox[24pt][r]{63.0} & \makebox[24pt][r]{45.0} & \makebox[24pt][r]{26.0} & \makebox[24pt][r]{11.0} & \makebox[24pt][r]{3.0} & \makebox[24pt][r]{31.2} & \makebox[24pt][r]{32.2} \\
    CLASS & \makebox[24pt][r]{90.0} & \makebox[24pt][r]{88.0} & \makebox[24pt][r]{84.0} & \makebox[24pt][r]{68.0} & \makebox[24pt][r]{40.0} & \makebox[24pt][r]{17.0} & \makebox[24pt][r]{59.4} & \makebox[24pt][r]{51.0} & \makebox[24pt][r]{74.0} & \makebox[24pt][r]{80.0} & \makebox[24pt][r]{78.0} & \makebox[24pt][r]{62.0} & \makebox[24pt][r]{40.0} & \makebox[24pt][r]{8.0} & \makebox[24pt][r]{52.2} & \makebox[24pt][r]{66.7} \\
    Cam-DP & \makebox[24pt][r]{95.0} & \makebox[24pt][r]{92.0} & \makebox[24pt][r]{94.0} & \makebox[24pt][r]{\underline{95.0}} & \makebox[24pt][r]{\underline{87.0}} & \makebox[24pt][r]{39.0} & \makebox[24pt][r]{81.4} & \makebox[24pt][r]{72.0} & \makebox[24pt][r]{\underline{93.0}} & \makebox[24pt][r]{97.0} & \makebox[24pt][r]{\underline{96.0}} & \makebox[24pt][r]{90.0} & \makebox[24pt][r]{70.0} & \makebox[24pt][r]{30.0} & \makebox[24pt][r]{75.2} & \makebox[24pt][r]{82.3} \\
    RAVEN & \makebox[24pt][r]{\textbf{100.0}} & \makebox[24pt][r]{\textbf{99.0}} & \makebox[24pt][r]{\underline{97.0}} & \makebox[24pt][r]{81.0} & \makebox[24pt][r]{71.0} & \makebox[24pt][r]{31.0} & \makebox[24pt][r]{75.8} & \makebox[24pt][r]{\textbf{95.0}} & \makebox[24pt][r]{\textbf{98.0}} & \makebox[24pt][r]{\textbf{100.0}} & \makebox[24pt][r]{\textbf{100.0}} & \makebox[24pt][r]{\textbf{100.0}} & \makebox[24pt][r]{\textbf{95.0}} & \makebox[24pt][r]{\textbf{89.0}} & \makebox[24pt][r]{\textbf{96.2}} & \makebox[24pt][r]{\underline{90.5}} \\
    \cmidrule(lr){1-17}
    \textbf{ReGeoAnch} & \makebox[24pt][r]{\underline{99.0}} & \makebox[24pt][r]{\textbf{99.0}} & \makebox[24pt][r]{\textbf{98.0}} & \makebox[24pt][r]{\textbf{98.0}} & \makebox[24pt][r]{\textbf{98.0}} & \makebox[24pt][r]{\textbf{94.0}} & \makebox[24pt][r]{\textbf{97.4}} & \makebox[24pt][r]{\underline{89.0}} & \makebox[24pt][r]{92.0} & \makebox[24pt][r]{94.0} & \makebox[24pt][r]{94.0} & \makebox[24pt][r]{\underline{96.0}} & \makebox[24pt][r]{\underline{90.0}} & \makebox[24pt][r]{\underline{86.0}} & \makebox[24pt][r]{\underline{91.2}} & \makebox[24pt][r]{\textbf{95.8}} \\
    \midrule
    \multicolumn{17}{@{}l}{\textbf{(b) PushCube}} \\
    DP & \makebox[24pt][r]{94.0} & \makebox[24pt][r]{91.0} & \makebox[24pt][r]{79.0} & \makebox[24pt][r]{49.0} & \makebox[24pt][r]{25.0} & \makebox[24pt][r]{15.0} & \makebox[24pt][r]{51.8} & \makebox[24pt][r]{52.0} & \makebox[24pt][r]{44.0} & \makebox[24pt][r]{94.0} & \makebox[24pt][r]{88.0} & \makebox[24pt][r]{90.0} & \makebox[24pt][r]{83.0} & \makebox[24pt][r]{71.0} & \makebox[24pt][r]{71.3} & \makebox[24pt][r]{64.7} \\
    DP\({}^{\dagger}\) & \makebox[24pt][r]{97.0} & \makebox[24pt][r]{96.0} & \makebox[24pt][r]{96.0} & \makebox[24pt][r]{\underline{95.0}} & \makebox[24pt][r]{85.0} & \makebox[24pt][r]{44.0} & \makebox[24pt][r]{83.2} & \makebox[24pt][r]{89.0} & \makebox[24pt][r]{82.0} & \makebox[24pt][r]{96.0} & \makebox[24pt][r]{95.0} & \makebox[24pt][r]{\underline{97.0}} & \makebox[24pt][r]{86.0} & \makebox[24pt][r]{81.0} & \makebox[24pt][r]{\underline{88.3}} & \makebox[24pt][r]{88.0} \\
    DP3 & \makebox[24pt][r]{96.0} & \makebox[24pt][r]{92.0} & \makebox[24pt][r]{89.0} & \makebox[24pt][r]{85.0} & \makebox[24pt][r]{83.0} & \makebox[24pt][r]{\underline{61.0}} & \makebox[24pt][r]{82.0} & \makebox[24pt][r]{80.0} & \makebox[24pt][r]{87.0} & \makebox[24pt][r]{94.0} & \makebox[24pt][r]{94.0} & \makebox[24pt][r]{88.0} & \makebox[24pt][r]{80.0} & \makebox[24pt][r]{62.0} & \makebox[24pt][r]{81.8} & \makebox[24pt][r]{76.9} \\
    3D-DA & \makebox[24pt][r]{95.0} & \makebox[24pt][r]{96.0} & \makebox[24pt][r]{95.0} & \makebox[24pt][r]{\underline{95.0}} & \makebox[24pt][r]{89.0} & \makebox[24pt][r]{13.0} & \makebox[24pt][r]{77.6} & \makebox[24pt][r]{79.0} & \makebox[24pt][r]{89.0} & \makebox[24pt][r]{96.0} & \makebox[24pt][r]{92.0} & \makebox[24pt][r]{87.0} & \makebox[24pt][r]{79.0} & \makebox[24pt][r]{63.0} & \makebox[24pt][r]{81.5} & \makebox[24pt][r]{81.6} \\
    ReViWo & \makebox[24pt][r]{79.0} & \makebox[24pt][r]{96.0} & \makebox[24pt][r]{73.0} & \makebox[24pt][r]{40.0} & \makebox[24pt][r]{8.0} & \makebox[24pt][r]{3.0} & \makebox[24pt][r]{44.0} & \makebox[24pt][r]{55.0} & \makebox[24pt][r]{86.0} & \makebox[24pt][r]{80.0} & \makebox[24pt][r]{28.0} & \makebox[24pt][r]{26.0} & \makebox[24pt][r]{11.0} & \makebox[24pt][r]{8.0} & \makebox[24pt][r]{35.7} & \makebox[24pt][r]{47.2} \\
    CLASS & \makebox[24pt][r]{\underline{99.0}} & \makebox[24pt][r]{\textbf{100.0}} & \makebox[24pt][r]{\textbf{100.0}} & \makebox[24pt][r]{\textbf{100.0}} & \makebox[24pt][r]{\textbf{100.0}} & \makebox[24pt][r]{42.0} & \makebox[24pt][r]{\underline{88.4}} & \makebox[24pt][r]{\underline{95.0}} & \makebox[24pt][r]{\underline{99.0}} & \makebox[24pt][r]{\underline{99.0}} & \makebox[24pt][r]{\underline{99.0}} & \makebox[24pt][r]{68.0} & \makebox[24pt][r]{24.0} & \makebox[24pt][r]{19.0} & \makebox[24pt][r]{67.3} & \makebox[24pt][r]{84.4} \\
    Cam-DP & \makebox[24pt][r]{98.0} & \makebox[24pt][r]{95.0} & \makebox[24pt][r]{\underline{97.0}} & \makebox[24pt][r]{94.0} & \makebox[24pt][r]{92.0} & \makebox[24pt][r]{40.0} & \makebox[24pt][r]{83.6} & \makebox[24pt][r]{78.0} & \makebox[24pt][r]{91.0} & \makebox[24pt][r]{93.0} & \makebox[24pt][r]{93.0} & \makebox[24pt][r]{91.0} & \makebox[24pt][r]{\underline{91.0}} & \makebox[24pt][r]{\underline{84.0}} & \makebox[24pt][r]{88.0} & \makebox[24pt][r]{86.5} \\
    RAVEN & \makebox[24pt][r]{\textbf{100.0}} & \makebox[24pt][r]{\textbf{100.0}} & \makebox[24pt][r]{\textbf{100.0}} & \makebox[24pt][r]{94.0} & \makebox[24pt][r]{75.0} & \makebox[24pt][r]{35.0} & \makebox[24pt][r]{80.8} & \makebox[24pt][r]{\textbf{100.0}} & \makebox[24pt][r]{\textbf{100.0}} & \makebox[24pt][r]{\textbf{100.0}} & \makebox[24pt][r]{\textbf{100.0}} & \makebox[24pt][r]{\textbf{100.0}} & \makebox[24pt][r]{87.0} & \makebox[24pt][r]{37.0} & \makebox[24pt][r]{87.3} & \makebox[24pt][r]{\underline{89.4}} \\
    \cmidrule(lr){1-17}
    \textbf{ReGeoAnch} & \makebox[24pt][r]{\textbf{100.0}} & \makebox[24pt][r]{\underline{98.0}} & \makebox[24pt][r]{\textbf{100.0}} & \makebox[24pt][r]{\textbf{100.0}} & \makebox[24pt][r]{\underline{99.0}} & \makebox[24pt][r]{\textbf{99.0}} & \makebox[24pt][r]{\textbf{99.2}} & \makebox[24pt][r]{93.0} & \makebox[24pt][r]{96.0} & \makebox[24pt][r]{97.0} & \makebox[24pt][r]{\underline{99.0}} & \makebox[24pt][r]{\underline{97.0}} & \makebox[24pt][r]{\textbf{96.0}} & \makebox[24pt][r]{\textbf{98.0}} & \makebox[24pt][r]{\textbf{96.5}} & \makebox[24pt][r]{\textbf{97.3}} \\
    \midrule
    \multicolumn{17}{@{}l}{\textbf{(c) PullCube}} \\
    DP & \makebox[24pt][r]{\textbf{100.0}} & \makebox[24pt][r]{88.0} & \makebox[24pt][r]{63.0} & \makebox[24pt][r]{34.0} & \makebox[24pt][r]{24.0} & \makebox[24pt][r]{8.0} & \makebox[24pt][r]{43.4} & \makebox[24pt][r]{78.0} & \makebox[24pt][r]{96.0} & \makebox[24pt][r]{\textbf{100.0}} & \makebox[24pt][r]{80.0} & \makebox[24pt][r]{58.0} & \makebox[24pt][r]{28.0} & \makebox[24pt][r]{4.0} & \makebox[24pt][r]{57.3} & \makebox[24pt][r]{58.3} \\
    DP\({}^{\dagger}\) & \makebox[24pt][r]{\textbf{100.0}} & \makebox[24pt][r]{\textbf{100.0}} & \makebox[24pt][r]{\textbf{100.0}} & \makebox[24pt][r]{\underline{99.0}} & \makebox[24pt][r]{88.0} & \makebox[24pt][r]{55.0} & \makebox[24pt][r]{88.4} & \makebox[24pt][r]{\underline{99.0}} & \makebox[24pt][r]{\textbf{100.0}} & \makebox[24pt][r]{\textbf{100.0}} & \makebox[24pt][r]{\textbf{100.0}} & \makebox[24pt][r]{\underline{99.0}} & \makebox[24pt][r]{96.0} & \makebox[24pt][r]{76.0} & \makebox[24pt][r]{95.0} & \makebox[24pt][r]{94.4} \\
    DP3 & \makebox[24pt][r]{\textbf{100.0}} & \makebox[24pt][r]{\textbf{100.0}} & \makebox[24pt][r]{\textbf{100.0}} & \makebox[24pt][r]{\underline{99.0}} & \makebox[24pt][r]{92.0} & \makebox[24pt][r]{53.0} & \makebox[24pt][r]{88.8} & \makebox[24pt][r]{96.0} & \makebox[24pt][r]{\textbf{100.0}} & \makebox[24pt][r]{\textbf{100.0}} & \makebox[24pt][r]{\textbf{100.0}} & \makebox[24pt][r]{\textbf{100.0}} & \makebox[24pt][r]{\underline{98.0}} & \makebox[24pt][r]{91.0} & \makebox[24pt][r]{97.5} & \makebox[24pt][r]{92.8} \\
    3D-DA & \makebox[24pt][r]{\textbf{100.0}} & \makebox[24pt][r]{\textbf{100.0}} & \makebox[24pt][r]{\underline{99.0}} & \makebox[24pt][r]{\underline{99.0}} & \makebox[24pt][r]{83.0} & \makebox[24pt][r]{49.0} & \makebox[24pt][r]{86.0} & \makebox[24pt][r]{\underline{99.0}} & \makebox[24pt][r]{\textbf{100.0}} & \makebox[24pt][r]{\underline{99.0}} & \makebox[24pt][r]{\underline{99.0}} & \makebox[24pt][r]{\textbf{100.0}} & \makebox[24pt][r]{\textbf{100.0}} & \makebox[24pt][r]{95.0} & \makebox[24pt][r]{98.8} & \makebox[24pt][r]{94.8} \\
    ReViWo & \makebox[24pt][r]{\underline{96.0}} & \makebox[24pt][r]{83.0} & \makebox[24pt][r]{46.0} & \makebox[24pt][r]{28.0} & \makebox[24pt][r]{8.0} & \makebox[24pt][r]{0.0} & \makebox[24pt][r]{33.0} & \makebox[24pt][r]{53.0} & \makebox[24pt][r]{82.0} & \makebox[24pt][r]{98.0} & \makebox[24pt][r]{97.0} & \makebox[24pt][r]{69.0} & \makebox[24pt][r]{5.0} & \makebox[24pt][r]{2.0} & \makebox[24pt][r]{51.3} & \makebox[24pt][r]{54.0} \\
    CLASS & \makebox[24pt][r]{\textbf{100.0}} & \makebox[24pt][r]{\textbf{100.0}} & \makebox[24pt][r]{\textbf{100.0}} & \makebox[24pt][r]{95.0} & \makebox[24pt][r]{91.0} & \makebox[24pt][r]{\underline{63.0}} & \makebox[24pt][r]{89.8} & \makebox[24pt][r]{\textbf{100.0}} & \makebox[24pt][r]{\textbf{100.0}} & \makebox[24pt][r]{\textbf{100.0}} & \makebox[24pt][r]{\textbf{100.0}} & \makebox[24pt][r]{\textbf{100.0}} & \makebox[24pt][r]{96.0} & \makebox[24pt][r]{66.0} & \makebox[24pt][r]{93.7} & \makebox[24pt][r]{94.1} \\
    Cam-DP & \makebox[24pt][r]{\textbf{100.0}} & \makebox[24pt][r]{\underline{99.0}} & \makebox[24pt][r]{\underline{99.0}} & \makebox[24pt][r]{97.0} & \makebox[24pt][r]{89.0} & \makebox[24pt][r]{31.0} & \makebox[24pt][r]{83.0} & \makebox[24pt][r]{\underline{99.0}} & \makebox[24pt][r]{\underline{99.0}} & \makebox[24pt][r]{\underline{99.0}} & \makebox[24pt][r]{\textbf{100.0}} & \makebox[24pt][r]{\textbf{100.0}} & \makebox[24pt][r]{\underline{98.0}} & \makebox[24pt][r]{90.0} & \makebox[24pt][r]{97.7} & \makebox[24pt][r]{93.5} \\
    RAVEN & \makebox[24pt][r]{\textbf{100.0}} & \makebox[24pt][r]{\textbf{100.0}} & \makebox[24pt][r]{\textbf{100.0}} & \makebox[24pt][r]{\underline{99.0}} & \makebox[24pt][r]{\underline{98.0}} & \makebox[24pt][r]{57.0} & \makebox[24pt][r]{\underline{90.8}} & \makebox[24pt][r]{\textbf{100.0}} & \makebox[24pt][r]{\textbf{100.0}} & \makebox[24pt][r]{\textbf{100.0}} & \makebox[24pt][r]{\textbf{100.0}} & \makebox[24pt][r]{\textbf{100.0}} & \makebox[24pt][r]{\textbf{100.0}} & \makebox[24pt][r]{\underline{98.0}} & \makebox[24pt][r]{\underline{99.7}} & \makebox[24pt][r]{\underline{96.8}} \\
    \cmidrule(lr){1-17}
    \textbf{ReGeoAnch} & \makebox[24pt][r]{\textbf{100.0}} & \makebox[24pt][r]{\textbf{100.0}} & \makebox[24pt][r]{\textbf{100.0}} & \makebox[24pt][r]{\textbf{100.0}} & \makebox[24pt][r]{\textbf{100.0}} & \makebox[24pt][r]{\textbf{99.0}} & \makebox[24pt][r]{\textbf{99.8}} & \makebox[24pt][r]{\textbf{100.0}} & \makebox[24pt][r]{\textbf{100.0}} & \makebox[24pt][r]{\textbf{100.0}} & \makebox[24pt][r]{\textbf{100.0}} & \makebox[24pt][r]{\textbf{100.0}} & \makebox[24pt][r]{\textbf{100.0}} & \makebox[24pt][r]{\textbf{100.0}} & \makebox[24pt][r]{\textbf{100.0}} & \makebox[24pt][r]{\textbf{99.9}} \\
    \midrule
    \multicolumn{17}{@{}l}{\textbf{(d) PokeCube}} \\
    DP & \makebox[24pt][r]{94.0} & \makebox[24pt][r]{70.0} & \makebox[24pt][r]{24.0} & \makebox[24pt][r]{3.0} & \makebox[24pt][r]{2.0} & \makebox[24pt][r]{1.0} & \makebox[24pt][r]{20.0} & \makebox[24pt][r]{40.0} & \makebox[24pt][r]{84.0} & \makebox[24pt][r]{94.0} & \makebox[24pt][r]{66.0} & \makebox[24pt][r]{13.0} & \makebox[24pt][r]{2.0} & \makebox[24pt][r]{0.0} & \makebox[24pt][r]{34.2} & \makebox[24pt][r]{33.8} \\
    DP\({}^{\dagger}\) & \makebox[24pt][r]{95.0} & \makebox[24pt][r]{91.0} & \makebox[24pt][r]{\underline{94.0}} & \makebox[24pt][r]{85.0} & \makebox[24pt][r]{74.0} & \makebox[24pt][r]{20.0} & \makebox[24pt][r]{72.8} & \makebox[24pt][r]{89.0} & \makebox[24pt][r]{\underline{93.0}} & \makebox[24pt][r]{96.0} & \makebox[24pt][r]{93.0} & \makebox[24pt][r]{\textbf{95.0}} & \makebox[24pt][r]{\underline{88.0}} & \makebox[24pt][r]{58.0} & \makebox[24pt][r]{86.0} & \makebox[24pt][r]{84.0} \\
    DP3 & \makebox[24pt][r]{86.0} & \makebox[24pt][r]{83.0} & \makebox[24pt][r]{80.0} & \makebox[24pt][r]{76.0} & \makebox[24pt][r]{61.0} & \makebox[24pt][r]{12.0} & \makebox[24pt][r]{62.4} & \makebox[24pt][r]{64.0} & \makebox[24pt][r]{81.0} & \makebox[24pt][r]{81.0} & \makebox[24pt][r]{69.0} & \makebox[24pt][r]{66.0} & \makebox[24pt][r]{70.0} & \makebox[24pt][r]{44.0} & \makebox[24pt][r]{65.7} & \makebox[24pt][r]{61.7} \\
    3D-DA & \makebox[24pt][r]{\textbf{99.0}} & \makebox[24pt][r]{92.0} & \makebox[24pt][r]{92.0} & \makebox[24pt][r]{\underline{88.0}} & \makebox[24pt][r]{\underline{84.0}} & \makebox[24pt][r]{12.0} & \makebox[24pt][r]{73.6} & \makebox[24pt][r]{86.0} & \makebox[24pt][r]{92.0} & \makebox[24pt][r]{94.0} & \makebox[24pt][r]{\underline{96.0}} & \makebox[24pt][r]{\underline{93.0}} & \makebox[24pt][r]{\textbf{89.0}} & \makebox[24pt][r]{\underline{71.0}} & \makebox[24pt][r]{\underline{87.8}} & \makebox[24pt][r]{85.9} \\
    ReViWo & \makebox[24pt][r]{86.0} & \makebox[24pt][r]{64.0} & \makebox[24pt][r]{43.0} & \makebox[24pt][r]{13.0} & \makebox[24pt][r]{0.0} & \makebox[24pt][r]{0.0} & \makebox[24pt][r]{24.0} & \makebox[24pt][r]{15.0} & \makebox[24pt][r]{53.0} & \makebox[24pt][r]{80.0} & \makebox[24pt][r]{74.0} & \makebox[24pt][r]{42.0} & \makebox[24pt][r]{10.0} & \makebox[24pt][r]{9.0} & \makebox[24pt][r]{33.8} & \makebox[24pt][r]{42.2} \\
    CLASS & \makebox[24pt][r]{94.0} & \makebox[24pt][r]{93.0} & \makebox[24pt][r]{78.0} & \makebox[24pt][r]{60.0} & \makebox[24pt][r]{14.0} & \makebox[24pt][r]{0.0} & \makebox[24pt][r]{49.0} & \makebox[24pt][r]{63.0} & \makebox[24pt][r]{77.0} & \makebox[24pt][r]{86.0} & \makebox[24pt][r]{71.0} & \makebox[24pt][r]{53.0} & \makebox[24pt][r]{34.0} & \makebox[24pt][r]{39.0} & \makebox[24pt][r]{56.2} & \makebox[24pt][r]{65.0} \\
    Cam-DP & \makebox[24pt][r]{\underline{97.0}} & \makebox[24pt][r]{88.0} & \makebox[24pt][r]{87.0} & \makebox[24pt][r]{81.0} & \makebox[24pt][r]{76.0} & \makebox[24pt][r]{2.0} & \makebox[24pt][r]{66.8} & \makebox[24pt][r]{92.0} & \makebox[24pt][r]{92.0} & \makebox[24pt][r]{\underline{98.0}} & \makebox[24pt][r]{89.0} & \makebox[24pt][r]{88.0} & \makebox[24pt][r]{\textbf{89.0}} & \makebox[24pt][r]{66.0} & \makebox[24pt][r]{86.0} & \makebox[24pt][r]{81.6} \\
    RAVEN & \makebox[24pt][r]{\textbf{99.0}} & \makebox[24pt][r]{\textbf{99.0}} & \makebox[24pt][r]{\textbf{99.0}} & \makebox[24pt][r]{\textbf{97.0}} & \makebox[24pt][r]{\textbf{87.0}} & \makebox[24pt][r]{\underline{52.0}} & \makebox[24pt][r]{\textbf{86.8}} & \makebox[24pt][r]{\textbf{96.0}} & \makebox[24pt][r]{\textbf{100.0}} & \makebox[24pt][r]{\textbf{99.0}} & \makebox[24pt][r]{\textbf{97.0}} & \makebox[24pt][r]{91.0} & \makebox[24pt][r]{78.0} & \makebox[24pt][r]{56.0} & \makebox[24pt][r]{86.3} & \makebox[24pt][r]{\textbf{91.0}} \\
    \cmidrule(lr){1-17}
    \textbf{ReGeoAnch} & \makebox[24pt][r]{96.0} & \makebox[24pt][r]{\underline{95.0}} & \makebox[24pt][r]{93.0} & \makebox[24pt][r]{\underline{88.0}} & \makebox[24pt][r]{75.0} & \makebox[24pt][r]{\textbf{56.0}} & \makebox[24pt][r]{\underline{81.4}} & \makebox[24pt][r]{\underline{93.0}} & \makebox[24pt][r]{92.0} & \makebox[24pt][r]{94.0} & \makebox[24pt][r]{89.0} & \makebox[24pt][r]{91.0} & \makebox[24pt][r]{87.0} & \makebox[24pt][r]{\textbf{87.0}} & \makebox[24pt][r]{\textbf{89.8}} & \makebox[24pt][r]{\underline{89.4}} \\
    \bottomrule
    \end{tabular*}
    \endgroup
\end{table*}

\begin{table*}[p]
    \centering
    \caption{Distance-scale success rates (\%) on ManiSkill3. Formatting follows Table~\ref{tab:supp-mw-angles-1}; Avg. excludes the source scale $1.0$.}
    \label{tab:supp-ms-distance}
    \begingroup
    \fontsize{9}{10.5}\selectfont
    \setlength{\tabcolsep}{0.5pt}
    \setlength{\aboverulesep}{1.5pt}
    \setlength{\belowrulesep}{1.5pt}
    \renewcommand{\arraystretch}{1.0}
    \begin{tabular*}{0.98\textwidth}{@{\extracolsep{\fill}}p{56pt}*{16}{r}@{}}
    \toprule
    & \multicolumn{8}{c}{\textbf{(a) PickCube}} & \multicolumn{8}{c}{\textbf{(b) PushCube}} \\
    \cmidrule(lr){2-9}\cmidrule(lr){10-17}
    \textbf{Method} & \makebox[24pt][c]{0.95} & \makebox[24pt][c]{0.975} & \makebox[24pt][c]{1.0} & \makebox[24pt][c]{1.05} & \makebox[24pt][c]{1.1} & \makebox[24pt][c]{1.15} & \makebox[24pt][c]{1.2} & \makebox[24pt][c]{\textbf{Avg.}} & \makebox[24pt][c]{0.95} & \makebox[24pt][c]{0.975} & \makebox[24pt][c]{1.0} & \makebox[24pt][c]{1.05} & \makebox[24pt][c]{1.1} & \makebox[24pt][c]{1.15} & \makebox[24pt][c]{1.2} & \makebox[24pt][c]{\textbf{Avg.}} \\
    \midrule
    DP & \makebox[24pt][r]{88.0} & \makebox[24pt][r]{92.0} & \makebox[24pt][r]{89.0} & \makebox[24pt][r]{79.0} & \makebox[24pt][r]{53.0} & \makebox[24pt][r]{42.0} & \makebox[24pt][r]{43.0} & \makebox[24pt][r]{66.2} & \makebox[24pt][r]{67.0} & \makebox[24pt][r]{79.0} & \makebox[24pt][r]{94.0} & \makebox[24pt][r]{91.0} & \makebox[24pt][r]{85.0} & \makebox[24pt][r]{60.0} & \makebox[24pt][r]{43.0} & \makebox[24pt][r]{70.8} \\
    DP\({}^{\dagger}\) & \makebox[24pt][r]{94.0} & \makebox[24pt][r]{95.0} & \makebox[24pt][r]{97.0} & \makebox[24pt][r]{91.0} & \makebox[24pt][r]{86.0} & \makebox[24pt][r]{82.0} & \makebox[24pt][r]{72.0} & \makebox[24pt][r]{86.7} & \makebox[24pt][r]{94.0} & \makebox[24pt][r]{96.0} & \makebox[24pt][r]{97.0} & \makebox[24pt][r]{92.0} & \makebox[24pt][r]{93.0} & \makebox[24pt][r]{92.0} & \makebox[24pt][r]{87.0} & \makebox[24pt][r]{92.3} \\
    DP3 & \makebox[24pt][r]{63.0} & \makebox[24pt][r]{84.0} & \makebox[24pt][r]{90.0} & \makebox[24pt][r]{62.0} & \makebox[24pt][r]{25.0} & \makebox[24pt][r]{11.0} & \makebox[24pt][r]{7.0} & \makebox[24pt][r]{42.0} & \makebox[24pt][r]{81.0} & \makebox[24pt][r]{91.0} & \makebox[24pt][r]{96.0} & \makebox[24pt][r]{81.0} & \makebox[24pt][r]{67.0} & \makebox[24pt][r]{53.0} & \makebox[24pt][r]{29.0} & \makebox[24pt][r]{67.0} \\
    3D-DA & \makebox[24pt][r]{95.0} & \makebox[24pt][r]{\underline{96.0}} & \makebox[24pt][r]{96.0} & \makebox[24pt][r]{\underline{95.0}} & \makebox[24pt][r]{93.0} & \makebox[24pt][r]{91.0} & \makebox[24pt][r]{\underline{90.0}} & \makebox[24pt][r]{93.3} & \makebox[24pt][r]{\underline{99.0}} & \makebox[24pt][r]{97.0} & \makebox[24pt][r]{96.0} & \makebox[24pt][r]{92.0} & \makebox[24pt][r]{91.0} & \makebox[24pt][r]{84.0} & \makebox[24pt][r]{52.0} & \makebox[24pt][r]{85.8} \\
    ReViWo & \makebox[24pt][r]{66.0} & \makebox[24pt][r]{71.0} & \makebox[24pt][r]{66.0} & \makebox[24pt][r]{64.0} & \makebox[24pt][r]{53.0} & \makebox[24pt][r]{40.0} & \makebox[24pt][r]{23.0} & \makebox[24pt][r]{52.8} & \makebox[24pt][r]{82.0} & \makebox[24pt][r]{80.0} & \makebox[24pt][r]{87.0} & \makebox[24pt][r]{74.0} & \makebox[24pt][r]{56.0} & \makebox[24pt][r]{44.0} & \makebox[24pt][r]{36.0} & \makebox[24pt][r]{62.0} \\
    CLASS & \makebox[24pt][r]{89.0} & \makebox[24pt][r]{90.0} & \makebox[24pt][r]{94.0} & \makebox[24pt][r]{92.0} & \makebox[24pt][r]{95.0} & \makebox[24pt][r]{86.0} & \makebox[24pt][r]{80.0} & \makebox[24pt][r]{88.7} & \makebox[24pt][r]{94.0} & \makebox[24pt][r]{\underline{98.0}} & \makebox[24pt][r]{\underline{99.0}} & \makebox[24pt][r]{\underline{99.0}} & \makebox[24pt][r]{\underline{98.0}} & \makebox[24pt][r]{\underline{98.0}} & \makebox[24pt][r]{\underline{98.0}} & \makebox[24pt][r]{\underline{97.5}} \\
    Cam-DP & \makebox[24pt][r]{97.0} & \makebox[24pt][r]{\underline{96.0}} & \makebox[24pt][r]{94.0} & \makebox[24pt][r]{93.0} & \makebox[24pt][r]{89.0} & \makebox[24pt][r]{86.0} & \makebox[24pt][r]{81.0} & \makebox[24pt][r]{90.3} & \makebox[24pt][r]{98.0} & \makebox[24pt][r]{96.0} & \makebox[24pt][r]{94.0} & \makebox[24pt][r]{92.0} & \makebox[24pt][r]{91.0} & \makebox[24pt][r]{79.0} & \makebox[24pt][r]{71.0} & \makebox[24pt][r]{87.8} \\
    RAVEN & \makebox[24pt][r]{\textbf{100.0}} & \makebox[24pt][r]{\textbf{100.0}} & \makebox[24pt][r]{\textbf{100.0}} & \makebox[24pt][r]{\textbf{100.0}} & \makebox[24pt][r]{\textbf{100.0}} & \makebox[24pt][r]{\textbf{99.0}} & \makebox[24pt][r]{\textbf{98.0}} & \makebox[24pt][r]{\textbf{99.5}} & \makebox[24pt][r]{\textbf{100.0}} & \makebox[24pt][r]{\textbf{100.0}} & \makebox[24pt][r]{\textbf{100.0}} & \makebox[24pt][r]{\textbf{100.0}} & \makebox[24pt][r]{\textbf{100.0}} & \makebox[24pt][r]{\textbf{100.0}} & \makebox[24pt][r]{\textbf{100.0}} & \makebox[24pt][r]{\textbf{100.0}} \\
    \cmidrule(lr){1-17}
    \textbf{ReGeoAnch} & \makebox[24pt][r]{\underline{99.0}} & \makebox[24pt][r]{\textbf{100.0}} & \makebox[24pt][r]{\underline{99.0}} & \makebox[24pt][r]{\textbf{100.0}} & \makebox[24pt][r]{\underline{99.0}} & \makebox[24pt][r]{\underline{97.0}} & \makebox[24pt][r]{\textbf{98.0}} & \makebox[24pt][r]{\underline{98.8}} & \makebox[24pt][r]{96.0} & \makebox[24pt][r]{95.0} & \makebox[24pt][r]{97.0} & \makebox[24pt][r]{97.0} & \makebox[24pt][r]{97.0} & \makebox[24pt][r]{97.0} & \makebox[24pt][r]{95.0} & \makebox[24pt][r]{96.2} \\
    \midrule
    & \multicolumn{8}{c}{\textbf{(c) PullCube}} & \multicolumn{8}{c}{\textbf{(d) PokeCube}} \\
    \cmidrule(lr){2-9}\cmidrule(lr){10-17}
    \textbf{Method} & \makebox[24pt][c]{0.95} & \makebox[24pt][c]{0.975} & \makebox[24pt][c]{1.0} & \makebox[24pt][c]{1.05} & \makebox[24pt][c]{1.1} & \makebox[24pt][c]{1.15} & \makebox[24pt][c]{1.2} & \makebox[24pt][c]{\textbf{Avg.}} & \makebox[24pt][c]{0.95} & \makebox[24pt][c]{0.975} & \makebox[24pt][c]{1.0} & \makebox[24pt][c]{1.05} & \makebox[24pt][c]{1.1} & \makebox[24pt][c]{1.15} & \makebox[24pt][c]{1.2} & \makebox[24pt][c]{\textbf{Avg.}} \\
    \midrule
    DP & \makebox[24pt][r]{\textbf{100.0}} & \makebox[24pt][r]{\textbf{100.0}} & \makebox[24pt][r]{\textbf{100.0}} & \makebox[24pt][r]{\underline{99.0}} & \makebox[24pt][r]{85.0} & \makebox[24pt][r]{44.0} & \makebox[24pt][r]{17.0} & \makebox[24pt][r]{74.2} & \makebox[24pt][r]{87.0} & \makebox[24pt][r]{92.0} & \makebox[24pt][r]{94.0} & \makebox[24pt][r]{85.0} & \makebox[24pt][r]{18.0} & \makebox[24pt][r]{2.0} & \makebox[24pt][r]{0.0} & \makebox[24pt][r]{47.3} \\
    DP\({}^{\dagger}\) & \makebox[24pt][r]{\textbf{100.0}} & \makebox[24pt][r]{\textbf{100.0}} & \makebox[24pt][r]{\textbf{100.0}} & \makebox[24pt][r]{\textbf{100.0}} & \makebox[24pt][r]{\textbf{100.0}} & \makebox[24pt][r]{\textbf{100.0}} & \makebox[24pt][r]{98.0} & \makebox[24pt][r]{99.7} & \makebox[24pt][r]{95.0} & \makebox[24pt][r]{97.0} & \makebox[24pt][r]{97.0} & \makebox[24pt][r]{95.0} & \makebox[24pt][r]{93.0} & \makebox[24pt][r]{91.0} & \makebox[24pt][r]{89.0} & \makebox[24pt][r]{93.3} \\
    DP3 & \makebox[24pt][r]{\textbf{100.0}} & \makebox[24pt][r]{\textbf{100.0}} & \makebox[24pt][r]{\textbf{100.0}} & \makebox[24pt][r]{\textbf{100.0}} & \makebox[24pt][r]{\underline{96.0}} & \makebox[24pt][r]{88.0} & \makebox[24pt][r]{68.0} & \makebox[24pt][r]{92.0} & \makebox[24pt][r]{76.0} & \makebox[24pt][r]{83.0} & \makebox[24pt][r]{80.0} & \makebox[24pt][r]{70.0} & \makebox[24pt][r]{56.0} & \makebox[24pt][r]{36.0} & \makebox[24pt][r]{22.0} & \makebox[24pt][r]{57.2} \\
    3D-DA & \makebox[24pt][r]{\textbf{100.0}} & \makebox[24pt][r]{\textbf{100.0}} & \makebox[24pt][r]{\textbf{100.0}} & \makebox[24pt][r]{\textbf{100.0}} & \makebox[24pt][r]{\textbf{100.0}} & \makebox[24pt][r]{\textbf{100.0}} & \makebox[24pt][r]{98.0} & \makebox[24pt][r]{99.7} & \makebox[24pt][r]{\underline{96.0}} & \makebox[24pt][r]{96.0} & \makebox[24pt][r]{95.0} & \makebox[24pt][r]{\underline{98.0}} & \makebox[24pt][r]{\underline{98.0}} & \makebox[24pt][r]{95.0} & \makebox[24pt][r]{94.0} & \makebox[24pt][r]{96.2} \\
    ReViWo & \makebox[24pt][r]{\underline{98.0}} & \makebox[24pt][r]{\underline{98.0}} & \makebox[24pt][r]{\underline{99.0}} & \makebox[24pt][r]{96.0} & \makebox[24pt][r]{87.0} & \makebox[24pt][r]{60.0} & \makebox[24pt][r]{27.0} & \makebox[24pt][r]{77.7} & \makebox[24pt][r]{78.0} & \makebox[24pt][r]{83.0} & \makebox[24pt][r]{86.0} & \makebox[24pt][r]{77.0} & \makebox[24pt][r]{66.0} & \makebox[24pt][r]{61.0} & \makebox[24pt][r]{47.0} & \makebox[24pt][r]{68.7} \\
    CLASS & \makebox[24pt][r]{\textbf{100.0}} & \makebox[24pt][r]{\textbf{100.0}} & \makebox[24pt][r]{\textbf{100.0}} & \makebox[24pt][r]{\textbf{100.0}} & \makebox[24pt][r]{\textbf{100.0}} & \makebox[24pt][r]{\underline{99.0}} & \makebox[24pt][r]{94.0} & \makebox[24pt][r]{98.8} & \makebox[24pt][r]{91.0} & \makebox[24pt][r]{93.0} & \makebox[24pt][r]{92.0} & \makebox[24pt][r]{91.0} & \makebox[24pt][r]{88.0} & \makebox[24pt][r]{90.0} & \makebox[24pt][r]{86.0} & \makebox[24pt][r]{89.8} \\
    Cam-DP & \makebox[24pt][r]{\textbf{100.0}} & \makebox[24pt][r]{\textbf{100.0}} & \makebox[24pt][r]{\textbf{100.0}} & \makebox[24pt][r]{\textbf{100.0}} & \makebox[24pt][r]{\textbf{100.0}} & \makebox[24pt][r]{\textbf{100.0}} & \makebox[24pt][r]{\underline{99.0}} & \makebox[24pt][r]{\underline{99.8}} & \makebox[24pt][r]{\underline{96.0}} & \makebox[24pt][r]{95.0} & \makebox[24pt][r]{\underline{98.0}} & \makebox[24pt][r]{95.0} & \makebox[24pt][r]{91.0} & \makebox[24pt][r]{93.0} & \makebox[24pt][r]{82.0} & \makebox[24pt][r]{92.0} \\
    RAVEN & \makebox[24pt][r]{\textbf{100.0}} & \makebox[24pt][r]{\textbf{100.0}} & \makebox[24pt][r]{\textbf{100.0}} & \makebox[24pt][r]{\textbf{100.0}} & \makebox[24pt][r]{\textbf{100.0}} & \makebox[24pt][r]{\textbf{100.0}} & \makebox[24pt][r]{\textbf{100.0}} & \makebox[24pt][r]{\textbf{100.0}} & \makebox[24pt][r]{\textbf{100.0}} & \makebox[24pt][r]{\textbf{100.0}} & \makebox[24pt][r]{\textbf{100.0}} & \makebox[24pt][r]{\textbf{100.0}} & \makebox[24pt][r]{\textbf{100.0}} & \makebox[24pt][r]{\textbf{100.0}} & \makebox[24pt][r]{\textbf{99.0}} & \makebox[24pt][r]{\textbf{99.8}} \\
    \cmidrule(lr){1-17}
    \textbf{ReGeoAnch} & \makebox[24pt][r]{\textbf{100.0}} & \makebox[24pt][r]{\textbf{100.0}} & \makebox[24pt][r]{\textbf{100.0}} & \makebox[24pt][r]{\textbf{100.0}} & \makebox[24pt][r]{\textbf{100.0}} & \makebox[24pt][r]{\textbf{100.0}} & \makebox[24pt][r]{\textbf{100.0}} & \makebox[24pt][r]{\textbf{100.0}} & \makebox[24pt][r]{\underline{96.0}} & \makebox[24pt][r]{\underline{99.0}} & \makebox[24pt][r]{97.0} & \makebox[24pt][r]{97.0} & \makebox[24pt][r]{94.0} & \makebox[24pt][r]{\underline{97.0}} & \makebox[24pt][r]{\underline{98.0}} & \makebox[24pt][r]{\underline{96.8}} \\
    \bottomrule
    \end{tabular*}
    \endgroup
\end{table*}

\clearpage

\end{document}